\pdfoutput=1
\documentclass{article}

\PassOptionsToPackage{numbers,sort&compress}{natbib}
\usepackage[preprint]{neurips_2026}         %

\usepackage[utf8]{inputenc}
\usepackage[T1]{fontenc}
\usepackage{hyperref}
\usepackage{url}
\usepackage{booktabs}
\usepackage{tabularx}
\usepackage{amsmath,amsfonts,amssymb}
\usepackage{nicefrac}
\usepackage{microtype}
\usepackage{xcolor}
\usepackage{graphicx}

\newcommand{\modelname}{TWICE}                    %
\newcommand{\pf}{PaO$_2$/FiO$_2$}
\newcommand{\sfr}{SpO$_2$/FiO$_2$}
\newcommand{\figpage}[3][0in]{\IfFileExists{figures/figures.pdf}%
  {\includegraphics[page=#2,width=#3,trim=0 #1 0 0,clip]{figures/figures.pdf}}%
  {\fbox{\parbox{\dimexpr#3-2\fboxsep-2\fboxrule\relax}{\centering\vspace{2em}Figure page #2: export figures/figures.pptx to 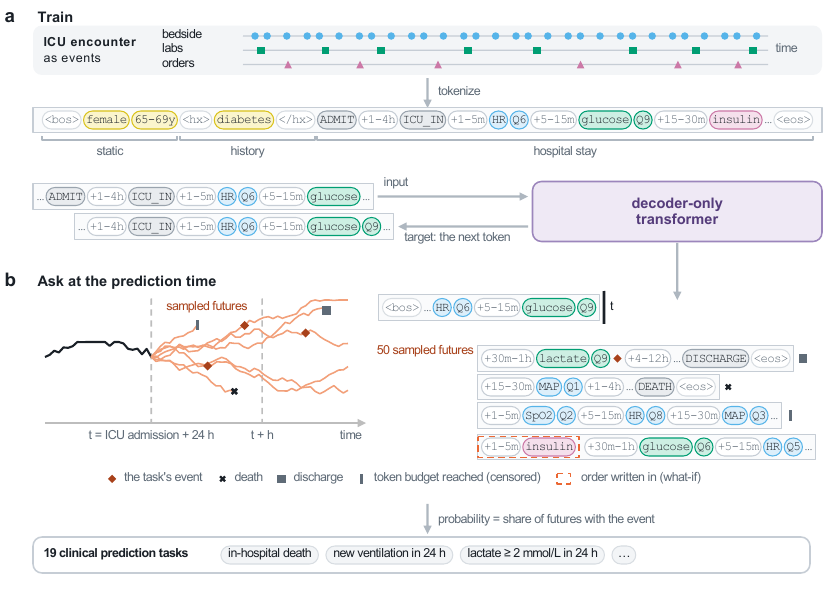\vspace{2em}}}}}

\title{\modelname{}: Toward a digital twin of ICU patients with a generative event model for forecasting trajectories and simulating treatment response}  %
\author{%
  Zhuoyang Lyu$^{1,2,*}$ \quad Yiming Li$^{2,3}$ \quad Peter C. Hou$^{4,5}$ \quad Pengyu Hong$^{6}$ \quad Li Zhou$^{1,2,3,*}$ \\[0.5em]
  \parbox{0.95\textwidth}{\centering\small
    $^{1}$Department of Biomedical Informatics, Harvard Medical School, Boston, MA, USA \\
    $^{2}$Division of General Internal Medicine and Primary Care, Department of Medicine, Brigham and Women's Hospital, Boston, MA, USA \\
    $^{3}$Department of Medicine, Harvard Medical School, Boston, MA, USA \\
    $^{4}$Division of Emergency Critical Care Medicine, Department of Emergency Medicine, Brigham and Women's Hospital, Boston, MA, USA \\
    $^{5}$Department of Emergency Medicine, Harvard Medical School, Boston, MA, USA \\
    $^{6}$Department of Computer Science, Brandeis University, Waltham, MA, USA \\[0.4em]
    $^{*}$Corresponding authors: \texttt{zhuoyang\_lyu@hms.harvard.edu}, \texttt{lzhou@bwh.harvard.edu}}
}

\begin{document}
\maketitle

\begin{abstract}
A digital twin of a patient in intensive care would forecast the patient's course and what a treatment would change. Generative event models, trained to continue a patient's electronic health record (EHR), have been proposed as a path toward such twins. We trained \modelname{}, a 79M-parameter decoder-only transformer, on the ICU records of one academic health system (127,332 training patients and 1.57 billion tokens). We evaluated it on clinical prediction tasks, generation fidelity, individual trajectories and counterfactual validation. Reading 19 clinical outcomes off 50 simulated futures, with no task-specific training, \modelname{} matched the best of 144 supervised configurations on held-out patients (mean AUROC 0.836 vs.\ 0.834). It led on events that change care within a day and trailed on death. Transferred without retraining to two public ICU databases, it matched the supervised model on MIMIC-IV and outperformed it on eICU. With the training data fixed, more and longer simulated futures were worth as much as or more than a 6.8-fold larger model. Its generated records matched real ones closely and were as specific to each patient as its own variability allowed. Its predicted risk rose as events approached. Pre-specified controls showed that it reproduced the side effects of drugs and was barely moved by an inert drug. On the values that treatments are given to correct, however, an opposite treatment moved its forecasts like the right one. \modelname{} forecasts ICU events about as well as task-specific models, and we see it as a step toward a digital twin of the ICU patient.
\end{abstract}

\section{Introduction}\label{sec:intro}

A digital twin is a virtual model, kept up to date with data from a real system, that predicts the system's behaviour and informs decisions about it~\citep{nasem2024twins}. In medicine, the system is a patient and the decisions concern their care. In the intensive care unit (ICU), the electronic health record (EHR) documents this care as a time-stamped sequence of observations and interventions, such as vital signs, laboratory results, drug orders and ventilator settings. Generative event models trained directly on such sequences learn to predict each event from those before it, and their predictions for a patient can be updated with every new entry in the EHR~\citep{renc2024ethos,waxler2025comet,shmatko2025delphi}. Prior work has used such models to sample many possible futures for a patient and estimate the risk of an outcome from how often it occurs among them~\citep{renc2024ethos,renc2025ares,waxler2025comet}. These zero-shot estimates were competitive with supervised models~\citep{renc2025ares,waxler2025comet}. Recent work has written a treatment into the record before sampling to simulate what it would change~\citep{pickard2026ehrmpc,lutsker2026healthformer,sainsbury2026flatascend}. Generative models of patient trajectories have been proposed as a path toward digital twins~\citep{makarov2025dtgpt,pickard2026ehrmpc,lutsker2026healthformer}.

Two gaps remain between these models and a twin at the ICU bedside. First, care in the ICU responds to events that unfold over hours, such as a fall in blood pressure, a rise in oxygen need, or the start or end of ventilation. Such events have been predicted by supervised models~\citep{yeche2021hirid,vandewater2024yaib}, by probes of generative event models~\citep{lee2026values,burkhart2026federated} and by ICU forecasters~\citep{makarov2025dtgpt,burger2025icarefm}. Whether simulated futures forecast them as well as supervised models do has been less studied. Second, whether they respond correctly when a treatment is written into the record has rarely been tested against controls. Without controls, a model can appear to respond correctly because it has learned which patients receive a treatment rather than what the treatment does.

We trained \modelname{}, a 79M-parameter decoder, on the ICU records of one academic health system (127,332 training patients and 1.57 billion training tokens), and evaluated it in four ways. The first two come from prior work on generative event models. Zero-shot clinical prediction~\citep{renc2024ethos,pang2025cehrgpt,waxler2025comet} is scored on 19 ICU tasks at 24\,h into a stay. Most of these tasks predict an event within the next day. Generation fidelity~\citep{waxler2025comet,pang2025cehrgpt,shmatko2025delphi} compares generated records with real ones. The other two reflect how a twin would be used. Individual trajectories test whether the model follows one patient through a stay~\citep{renc2025ares,shmatko2025delphi}. Counterfactual validation uses pre-specified controls to test whether the model responds correctly when a treatment is written into the record~\citep{pickard2026ehrmpc,lutsker2026healthformer,sainsbury2026flatascend}. Fig.~\ref{fig:overview} shows how \modelname{} is trained and how it makes predictions. We make four contributions.
\begin{itemize}
  \item \emph{Zero-shot prediction of ICU events.} With no task-specific training, \modelname{} matches the strongest of 144 supervised configurations on 19 ICU tasks on held-out patients (mean area under the receiver operating characteristic curve [AUROC] 0.836 for \modelname{} and 0.834 for the best supervised configuration). It scores higher on events that change care within a day and lower on death and oxygen weaning. Transferred without retraining to MIMIC-IV and eICU, it matches the same baseline on the first and outperforms it on the second (§\ref{sec:l1}).
  \item \emph{The readout as a lever.} In matched ablations with the training data fixed, four to five times more readout compute (more and longer simulated futures) improved AUROC as much as or more than a 6.8-fold larger model. Futures that stop at the token budget are censored informatively, so the token budget is part of the estimator (§\ref{sec:levers}).
  \item \emph{Generation fidelity and individual trajectories.} Compared at the next token and over the next 24\,h, generated records match real ones closely. A patient's real future overlaps a generated one about as much as two generated futures overlap each other. Patients whose event comes sooner receive a higher predicted risk (§\ref{sec:l2}, §\ref{sec:l3}).
  \item \emph{Counterfactual validation with controls.} Pre-specified positive, negative and opposite-treatment controls show that \modelname{} reproduces drug side effects. On a treatment's target, however, the opposite treatment moved the forecast like the right one, as expected if an order is read as a sign of illness (§\ref{sec:l4}). We propose these controls as a general test of counterfactual responses in generative event models.
\end{itemize}

\section{Related work}\label{sec:related}

\paragraph{Generative event models.} Foresight extracted medical concepts such as disorders and procedures from clinical notes and placed them on a timeline. It was trained to predict each patient's next concept~\citep{kraljevic2024foresight}. ETHOS turned each MIMIC-IV patient's record into a sequence of tokens. From simulated futures it estimated in-hospital and ICU death, ICU readmission and ICU length of stay~\citep{renc2024ethos}. ARES used ETHOS to estimate the risks of hospital admission, ICU admission and a prolonged stay, and to update them as the record grows~\citep{renc2025ares}. Delphi-2M was trained on the records of 0.4 million UK Biobank participants. From a person's past diagnoses it predicts which of more than 1,000 diseases comes next and when~\citep{shmatko2025delphi}. CEHR-XGPT uses one model in three ways. It provides patient features for downstream classifiers, predicts outcomes zero-shot and generates synthetic patient records~\citep{pang2025cehrgpt}. CoMET (Curiosity) was trained on 118 million patients from Epic Cosmos, which pools records from 310 health systems. Its authors fitted scaling laws for compute, data and model size. More simulated futures improved its estimates, and it generally matched or beat supervised models on 78 tasks~\citep{waxler2025comet}. A simulator trained on more than 200 million records from one Japanese hospital generated each patient's next seven days. The event rates, laboratory values and timing in these futures closely matched those in the patients' real futures~\citep{akagi2026simulator}. DT-Transformer adapted Delphi-2M's architecture and trained it on 1.7 million patients at Mass General Brigham, the health system our data come from. It predicts the next diagnosis across 896 disease categories~\citep{zhu2026dttransformer}. These models are judged mainly by zero-shot prediction of clinical outcomes~\citep{renc2024ethos,renc2025ares,pang2025cehrgpt,waxler2025comet,akagi2026simulator}, by next-event prediction~\citep{shmatko2025delphi,zhu2026dttransformer}, and by how closely generated records match real ones. CoMET and the simulator compare how often generated events occur, alone and together, with the patients' real futures~\citep{waxler2025comet,akagi2026simulator}. CEHR-XGPT and Delphi-2M train models on synthetic records and test them on real ones~\citep{pang2025cehrgpt,shmatko2025delphi}. Representation models such as CLMBR, MOTOR and APOLLO, the last trained on Mass General Brigham records, are probed or fine-tuned rather than simulated~\citep{steinberg2021clmbr,steinberg2024motor,zhang2026apollo}. In the ICU, DT-GPT forecasts laboratory and vital-sign values with a language model~\citep{makarov2025dtgpt}, and ICareFM predicts when physiological thresholds will be crossed~\citep{burger2025icarefm}.

\paragraph{ICU prediction and baselines.} Supervised models have predicted ICU mortality, decompensation, length of stay and organ failure on public ICU databases~\citep{harutyunyan2019multitask,yeche2021hirid}. YAIB ran shared tasks across five such databases and found that the choice of dataset, cohort and preprocessing often affected performance more than the choice of model~\citep{vandewater2024yaib}. We therefore compare every model on the same cohort, splits and labels. \citet{lee2026values} and \citet{burkhart2026federated} trained generative event models on hospitalisations and fitted probes on the model's state at 24\,h to predict events later in the stay. \citet{burkhart2026federated} predicted 12 such events, including invasive ventilation and vasopressor use, and trained one model across three health systems by federated learning. MEDS-Tab summarises MEDS event data over time windows into tables and trains XGBoost on them as a baseline for any MEDS dataset~\citep{oufattole2024medstab}. Our strongest supervised baseline is of this kind (§\ref{sec:eval}).

\paragraph{Interventions in generative models.} Writing a treatment into the record and sampling forward from a generative model is a form of Monte Carlo g-computation~\citep{robins1986}. Several generative event models have done so. EHR-MPC planned sepsis treatment by model predictive control~\citep{pickard2026ehrmpc}. HealthFormer compared simulated treatment effects outside the ICU with 41 published trial results~\citep{lutsker2026healthformer}. FlatASCEND inserted drugs into MIMIC-IV timelines and found 6 of 10 effects in the expected direction, though a null control failed~\citep{sainsbury2026flatascend}. Estimates of this kind assume that the record captures why each treatment was given. When it does not, a treatment can read as a sign of the illness it was given for. A vasopressor, for example, is started when blood pressure is low and not recovering. A model trained on such records may then predict lower blood pressure after a vasopressor is written in. This bias is known as confounding by indication~\citep{kyriacou2016indication}. Negative controls can detect it~\citep{lipsitch2010negative}. Target trial emulation designs an observational analysis to mirror a randomised trial~\citep{hernan2016target}, and when that trial has been run its result gives a second check. We apply both checks to an ICU model with pre-specified controls. They include side effects, an inert drug, opposite treatments and two trial emulations, one of a null trial.

\section{Data, prediction contract and tasks}\label{sec:data}

\begin{figure}[t]
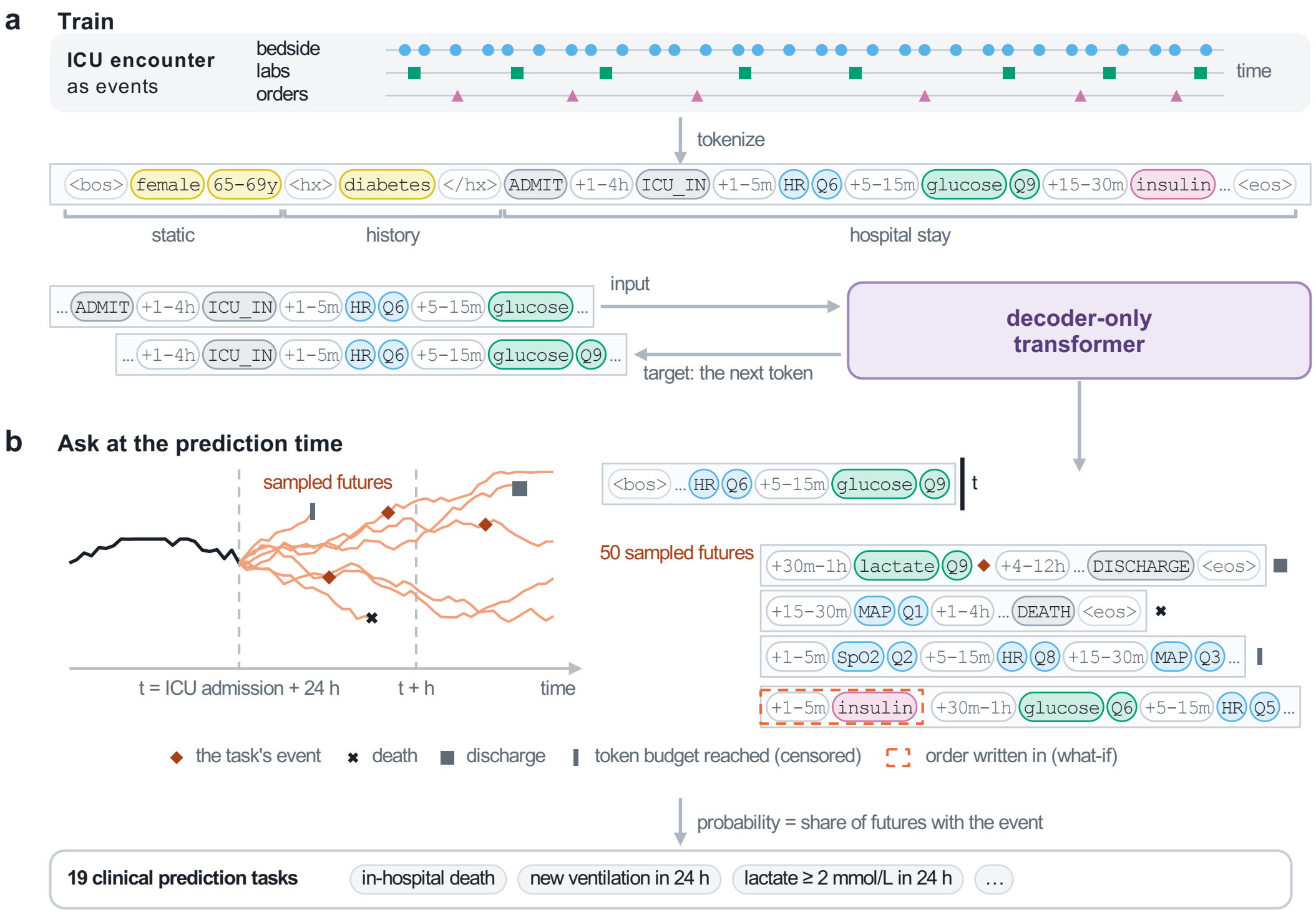

  \centering
  \figpage[0.14in]{1}{\textwidth}
  \caption{Overview of \modelname{}. \textbf{(a)} Training. The events of an ICU encounter are represented by one token sequence (§\ref{sec:tokens}). The transformer is trained to predict each next token (§\ref{sec:model}). \textbf{(b)} Prediction. At the prediction time $t$, \modelname{} samples 50 futures from the encounter up to $t$ (§\ref{sec:readout}). The predicted probability of a task is the fraction of decided futures in which its event occurs. For a what-if, an order is written into the encounter before sampling (§\ref{sec:intervene}). Token names are shortened for display.}
  \label{fig:overview}
\end{figure}

\paragraph{Cohort.} We used the Epic electronic health record of Mass General Brigham (MGB), an academic health system in Massachusetts. The cohort is every ICU stay of an adult (aged 18 or older, rounded to the year, at the start of the stay) that began between 2017 and 2025. It comprises 141,556 patients, 185,244 hospital encounters and 236,906 ICU stays (Table~\ref{tab:cohort}). For each encounter we took bedside flowsheet measurements (vital signs, ventilator settings, organ-support markers), laboratory results, medication orders, procedure, imaging and echocardiography orders, and note types (no text), from 24\,h before the first ICU admission to 24\,h after the last ICU discharge, together with the times of hospital admission and discharge, ICU admission and discharge, and death. Encounter diagnoses (ICD-10) are placed at the discharge time, so they cannot be read before the encounter ends (App.~\ref{app:integrity}). Problem-list diagnoses documented before admission form a history block. Converted to the Medical Event Data Standard (MEDS)~\citep{mcdermott2026meds}, the record holds 958,956,444 time-stamped events. Patients are split at random 90/5/5 into training, validation and test sets, so all of a patient's encounters fall in one set.

\paragraph{Prediction contract.} All methods are scored on the same patients at the same prediction time, 24\,h after ICU admission. Of the 236,906 ICU stays, 178,506 (75.3\%) are still in the ICU at that time. Stays that end earlier are not scored, so the results describe patients who remain in the ICU for a day. Inputs hold only events recorded at or before the prediction time. Validation has 8,947 scored stays from 5,836 patients and test 8,995 from 5,794 patients. A stay is not scored for a task when the patient is not eligible (for example, already ventilated, for new ventilation) or when the measurement the label reads was not taken. If the patient leaves the setting where the event would be recorded before the window closes (the ICU for bedside measurements, the hospital for orders), the label is negative (App.~\ref{app:tasks}).

\paragraph{Tasks.} We report 19 binary tasks in five groups (Table~\ref{tab:tasks}). Each task is a rule over structured events. Seven tasks are respiratory. They are severe hypoxaemia by \pf{} and by \sfr{}, moderate hypoxaemia, FiO$_2$ escalation and weaning, new invasive ventilation and ventilator liberation. The one haemodynamic task is hypotension. Six tasks are metabolic and renal. They are two levels each of acidaemia and hyperlactataemia, acute kidney injury and new continuous renal replacement therapy (CRRT). Four tasks are outcomes. They are ICU and in-hospital mortality, ICU discharge and death among patients on vasopressors. The fifth group is the care process. Its one task is an imaging or echocardiography order. This makes the care process itself a target~\citep{agniel2018biases}. Thresholds follow published definitions where they exist~\citep{ranieri2012berlin,matthay2024global,boles2007weaning,evans2021ssc,ardsnet2000arma,singer2016sepsis3,kdigo2012aki,casserly2015lactate,jung2011acidosis}. Most horizons are 24\,h. Table~\ref{tab:tasks} gives each task's threshold, horizon and eligibility rule. Two rules chose these tasks from a larger registry before the test set was scored. A task needs at least 50 positive rows in validation. Its label must also be computable from structured data for every stay (App.~\ref{app:tasks}). The same 19 tasks are used on validation and test. Predicting which discharge diagnosis groups (HCUP CCSR~\citep{hcup2026ccsr}) a stay will be recorded with is a separate family (§\ref{sec:l1}, App.~\ref{app:dx}). Each group collects related ICD-10 codes into one clinical category, such as heart failure or pneumonia.

\paragraph{External databases.} We also evaluated \modelname{} on two public ICU databases. MIMIC-IV (version 3.1) holds the records of patients admitted to an ICU or the emergency department of Beth Israel Deaconess Medical Center in Boston between 2008 and 2022~\citep{johnson2024mimiciv31,johnson2023mimic}. The eICU Collaborative Research Database (eICU, version 2.0) holds more than 200,000 ICU admissions at 208 hospitals across the United States in 2014 and 2015~\citep{pollard2019eicu20,pollard2018eicu}. Both are distributed by PhysioNet~\citep{pollard2026physionet}. We converted each database to MEDS with its published converter~\citep{mcdermott2026mimicivmeds,mcdermott2026eicumeds}. We then mapped its bedside flowsheet measurements, laboratory results and medication orders to the codes used at MGB. Procedure, imaging and echocardiography orders, note types and diagnoses were not mapped, nor was the charting of CRRT. From each database we drew 5,000 ICU stays of adults, sampling patients at random. The MIMIC-IV stays began in 2017 or later, as at MGB. The eICU stays come from the 133 hospitals that charted heart rate for at least half of their stays and named medication orders for at least half (App.~\ref{app:ext}). Prediction times, inputs and labels follow the prediction contract above. The imaging-order and new-CRRT tasks have no positive stays in either database, so the other 17 tasks are scored at all three sites.

\begin{table}[t]
  \caption{Cohort by split. App.~\ref{app:data} adds race, ethnicity and the outcomes of the scored stays. $^a$One ICU unit admission in the record, so a transfer between ICUs starts a new stay. $^b$Median [interquartile range]. $^c$Read from ventilator-setting markers (App.~\ref{app:tasks}).}
  \label{tab:cohort}
  \centering
  \small
  \setlength{\tabcolsep}{3.5pt}
  \begin{tabular}{@{}lrrrr@{}}
  \toprule
  & All & Train & Validation & Test \\
  \midrule
  Patients & 141,556 & 127,332 & 7,173 & 7,051 \\
  Hospital encounters & 185,244 & 166,745 & 9,195 & 9,304 \\
  ICU stays$^a$ & 236,906 & 213,265 & 11,789 & 11,852 \\
  Events (millions) & 959.0 & 863.5 & 48.0 & 47.5 \\
  Scored stays (in ICU at 24\,h) & 178,506 & 160,564 & 8,947 & 8,995 \\
  Patients with a scored stay & 115,295 & 103,665 & 5,836 & 5,794 \\
  Age at ICU admission, years$^b$ & 65.7 [53.8, 75.5] & 65.7 [53.8, 75.4] & 65.6 [53.3, 75.7] & 65.7 [53.3, 75.8] \\
  Female, \% & 43.7 & 43.7 & 43.7 & 43.8 \\
  ICU length of stay, days$^b$ & 2.0 [1.0, 4.2] & 2.0 [1.0, 4.2] & 2.0 [1.0, 4.3] & 2.1 [1.0, 4.2] \\
  Hospital length of stay, days$^b$ & 6.6 [3.4, 12.6] & 6.7 [3.4, 12.6] & 6.7 [3.4, 12.7] & 6.4 [3.3, 12.1] \\
  Invasive ventilation, \% of stays$^c$ & 43.4 & 43.5 & 43.5 & 42.3 \\
  In-hospital death, \% of encounters & 10.8 & 10.8 & 10.9 & 11.1 \\
  \bottomrule
\end{tabular}

\end{table}

\section{Method}\label{sec:method}

\subsection{Tokenisation}\label{sec:tokens}
We represent each hospital encounter as one token sequence (Fig.~\ref{fig:overview}a and App.~\ref{app:tokens}). The sequence begins with a static block (sex, race, ethnicity, age in 5-year bands and admission year). It continues with the diagnoses on the patient's problem list from before the admission, and then with the events of the encounter in time order. The sequence ends at discharge.
\begin{itemize}
  \item \emph{Codes.} Bedside, laboratory, medication, procedure and diagnosis events are each represented by a token for their code. We map a code recorded for fewer than 0.5\% of training patients to the token of its broader parent code, which is the LOINC analyte for a laboratory test and the three-character ICD-10 category for a diagnosis. A rare code without a parent, or whose parent is also rare, is mapped to a single ``other'' token for its family, which pools all such codes. Bedside physiology, ventilator, organ-support, admission and discharge codes are exempt from this rule. Diagnoses and medications are represented hierarchically so that related codes share tokens~\citep{waxler2025comet}. An ICD-10 code is split into up to three tokens (characters 1--3, 4--5 and 6--7). A medication is split in the same way into up to three tokens from its Anatomical Therapeutic Chemical (ATC) code~\citep{whoatc}.
  \item \emph{Values.} Each numeric result is represented by a token for its code followed by a token for the decile of its value~\citep{renc2024ethos,waxler2025comet}. We fit the deciles for each code and unit on the training split only. All codes share the same ten decile tokens. The ``other'' token records presence only and has no decile token.
  \item \emph{Time.} We insert time-interval tokens~\citep{renc2024ethos} between consecutive events that are at least one minute apart. Each token is selected from seven intervals (1--5, 5--15, 15--30 and 30--60\,min, and 1--4, 4--12 and $\geq$12\,h). A longer gap is represented by up to eight repeated $\geq$12\,h tokens. The intervals are spaced for the ICU, where most gaps are minutes long.
  \item \emph{Order within one minute.} Events stamped with the same minute are placed in a fixed clinical order. Admission and transfer events, which record where the patient is, come first. Observations follow. These are bedside measurements, such as heart rate or a ventilator setting, and laboratory results. Values computed from the observations, such as the oxygenation bands below, come after them. Treatments (medication orders and procedures) come next, then documentation (imaging, echocardiography and clinical notes) and diagnoses. Leaving events (ICU discharge, death and hospital discharge) come last. A computed value never precedes its evidence, and a treatment follows the physiology charted with it.
  \item \emph{Oxygenation bands.} \pf{} and \sfr{} are ratios of two measurements taken at different times, so no single value token can represent them. Wherever a ratio can be computed, we add a token for its band. The \pf{} bands are the Berlin bands ($\geq$300, 200--300, 100--200 and $<$100)~\citep{ranieri2012berlin}. \sfr{} has two bands, above and below 148~\citep{matthay2024global}.
\end{itemize}
The vocabulary contains 4,986 tokens. The training split yields 166,745 encounter sequences with 1.65 billion tokens. To select the checkpoint, we hold out 5\% of the training encounters at random (8,337 sequences and 82.8 million tokens). This hold-out is separate from the validation set. The remaining 158,408 sequences and 1.57 billion tokens are used for training.

\subsection{Model and training}\label{sec:model}
\modelname{} is a decoder-only transformer with RMSNorm pre-normalisation~\citep{zhang2019rmsnorm}, SwiGLU feed-forward layers~\citep{shazeer2020glu}, rotary position embeddings~\citep{su2024roformer} and grouped-query attention~\citep{ainslie2023gqa}. The input and output embeddings are tied. The position embeddings encode only where a token sits in the sequence, not when the event happened. The model learns how much time has passed from the time-interval tokens alone. The model has 12 layers of width 768 and 79.3 million parameters, and it uses a context window of 2,048 tokens. It is trained by next-token cross-entropy on 2,048-token windows that start at a uniformly random token of the concatenated training sequences, so every training token is equally likely to be seen~\citep{renc2024ethos}. Table~\ref{tab:train} lists the optimiser and the other training settings. Training stopped when the held-out cross-entropy had not improved for 20 evaluations, at step 435,000 (57.0 billion tokens, or 36.4 passes over the training tokens). We evaluate the best checkpoint, at step 415,000, which reaches 0.630 nats per token compared with 2.98 for a bigram model. Training took about 32\,h on one NVIDIA H200 GPU. The model size, the context length and the readout settings below were chosen on validation (§\ref{sec:levers}). No validation or test patient is seen in training.

\subsection{Reading clinical predictions off simulated futures}\label{sec:readout}
To make a prediction for a patient at the prediction time $t$, we prompt \modelname{} with the last 2,048 tokens of the encounter up to $t$. We then sample $N=50$ futures at temperature 1 from the full next-token distribution. Each generated time-interval token advances a future's clock by the geometric midpoint of its interval~\citep{waxler2025comet}. The open-ended $\geq$12\,h interval advances it by $\sqrt{12\cdot 24}\approx 17$\,h. A future stops at hospital discharge or death, after 8,192 generated tokens (the budget), or when its clock passes 30 days~\citep{renc2024ethos}. For a task with horizon $h$, we call a future \emph{decided} when its outcome within $h$ is known. It is positive if the event occurs within $h$. It is negative if the future reaches $h$ or the end of the stay without the event. A future that hits the budget or the 30-day limit before either outcome is \emph{undecided} for that task (right-censored). A future can decide a 24\,h task long before it stops. It decides in-hospital mortality only when it stops at discharge or death. The predicted probability is the fraction of decided futures in which the event occurs~\citep[eq.~2]{waxler2025comet},
\begin{equation}
  \hat p = \frac{\#\{\text{futures with the event within } h\}}{\#\{\text{decided futures}\}}.
  \label{eq:readout}
\end{equation}
If no future is decided, Eq.~\eqref{eq:readout} has nothing to count, and the futures give no evidence on whether the event occurs. We therefore predict the fraction of training stays in which the event occurred, which uses no information about the patient. Every scored stay thus receives a prediction. Events are identified by exact token matches. Ventilation is identified by its ventilator-setting tokens, and an oxygenation threshold by the band tokens. A threshold on one analyte (for example lactate $\geq$2\,mmol/L) is identified by the analyte's code token followed by a decile token. The label for such a task exists only where the analyte was measured, so the readout likewise counts only the futures in which the analyte is measured within the window. When the threshold falls inside a decile, the token for that decile does not show whether the value is above or below the threshold. We therefore compute $\hat p$ twice. The first count uses only the deciles wholly beyond the threshold, and the second also includes the decile that contains it. We report the midpoint of the two. A rise or fall is judged against the patient's own value before $t$. The same 50 futures are used for every task. During sampling, we cache the keys and values of only the last 2,048 tokens. Generation therefore uses sliding-window attention, whereas the model was trained with full attention within each window. App.~\ref{app:readout} measures what this changes.

\subsection{Writing an intervention into the record}\label{sec:intervene}
To ask ``what if'', we compare treatment arms with a control arm at a decision time $t$, as in a clinical trial. Treatments are compared within a clinical scenario, such as a high potassium. In six of our seven scenarios the treatments are drugs. In one they are ventilator settings (§\ref{sec:eval}). The arms differ only in what is ordered. When the treatments are drugs, the control arm forbids all of them for the whole future. Whenever the model samples one of them, the event is discarded and redrawn from the same step's distribution, renormalised over the allowed events. For a ventilator change, the control arm is usual care. Nothing is inserted or forbidden, and the model may change the ventilator settings as it predicts. The order is placed at $t+d$, where $d$ is the median delay from the decision time to the order in the training data. Until then, every arm is sampled as the control arm. At $t+d$ we insert a time-interval token and the order's own token sequence and continue sampling. All arms share the prompt and the random seeds, so the tokens before the order are identical across arms and every contrast is paired. After the order, nothing is forbidden and care continues as the model predicts, so an arm estimates ``order $a$ now, then usual care''. This is Monte Carlo g-computation~\citep{robins1986}. Its contrasts are causal only if the history holds what drove the order (no unmeasured confounding) and similar patients received each arm (positivity)~\citep{hernan2020whatif}. §\ref{sec:l4} tests these assumptions with controls rather than granting them.

\section{Evaluation design}\label{sec:eval}

\paragraph{Protocol and statistics.} A single generative model, selected on validation, was evaluated once on the test split, as were the supervised baselines. We then scored \modelname{} and the strongest supervised baseline once on the two external databases (§\ref{sec:data}), without retraining or recalibration. The clinical prediction tasks are scored on the test split. Generation fidelity, individual trajectories and counterfactual validation are scored on validation. Ablations are reported on validation and labelled as such. For the clinical prediction tasks, we compare \modelname{} with the strongest supervised baseline. This is the best of 144 supervised configurations of tabular and sequence models trained on the task labels, selected by mean validation AUROC over the 19 tasks. It is LightGBM~\citep{ke2017lightgbm} on summaries of all events charted in the previous 24\,h (App.~\ref{app:baselines}). We report 95\% percentile confidence intervals from 1,000 bootstrap replicates. Each replicate resamples patients rather than stays, to account for patients with several stays~\citep{field2007clustered}. A mean over tasks uses the same resampled patients for every task. Two-sided $p$-values are computed from the bootstrap distribution of the paired difference and adjusted across the 19 tasks with Holm's method~\citep{holm1979}. The expected calibration error (ECE)~\citep{guo2017calibration} is computed over 10 equal-width bins of predicted probability.

\paragraph{Counterfactual validation.} We test the interventions of §\ref{sec:intervene} in seven clinical scenarios on validation. Five begin at a charted event after which clinicians commonly choose a treatment. These events are a high potassium, a low glucose, a low blood pressure, a low oxygen saturation on the ventilator and the start of mechanical ventilation. The other two copy the entry criteria of the ROSE trial~\citep{moss2019rose} and of the invasively ventilated patients in the RECOVERY trial~\citep{recovery2021dexa}. The effect of a treatment on an outcome is the change in the outcome's predicted probability against the control arm (§\ref{sec:intervene}), averaged over ICU stays. We pre-specified the outcomes and the expected direction of each effect. App.~\ref{app:cf} gives the full design and a worked example.

\section{Results}\label{sec:results}

\subsection{Clinical prediction tasks}\label{sec:l1}

\begin{table}[t]
  \caption{Clinical prediction tasks, as means over the 19 test tasks. Differences are \modelname{} minus the strongest supervised baseline, paired on identical rows. The CI comes from 1,000 bootstrap replicates that resample patients (§\ref{sec:eval}). Validation AUROC is given for reference. Per-task values are in App.~\ref{app:full}.}
  \label{tab:l1}
  \centering
  \footnotesize
  \setlength{\tabcolsep}{4pt}
  \begin{tabular}{@{}lccccc@{}}
  \toprule
  & \multicolumn{4}{c}{Test} & Validation \\
  \cmidrule(lr){2-5}\cmidrule(l){6-6}
  & AUROC $\uparrow$ & AUPRC $\uparrow$ & Brier $\downarrow$ & ECE $\downarrow$ & AUROC $\uparrow$ \\
  \midrule
  Strongest supervised baseline & 0.8345 & 0.4954 & 0.0889 & 0.0122 & 0.8329 \\
  \modelname{} & 0.8362 & 0.5226 & 0.0892 & 0.0297 & 0.8357 \\
  \midrule
  Difference & $+0.0017$ & $+0.0272$ & & & $+0.0028$ \\
  \quad 95\% CI & [$-0.0053$, $+0.0090$] & [$+0.0129$, $+0.0405$] & & & [$-0.0047$, $+0.0104$] \\
  \bottomrule
\end{tabular}

\end{table}

\begin{figure}[t]
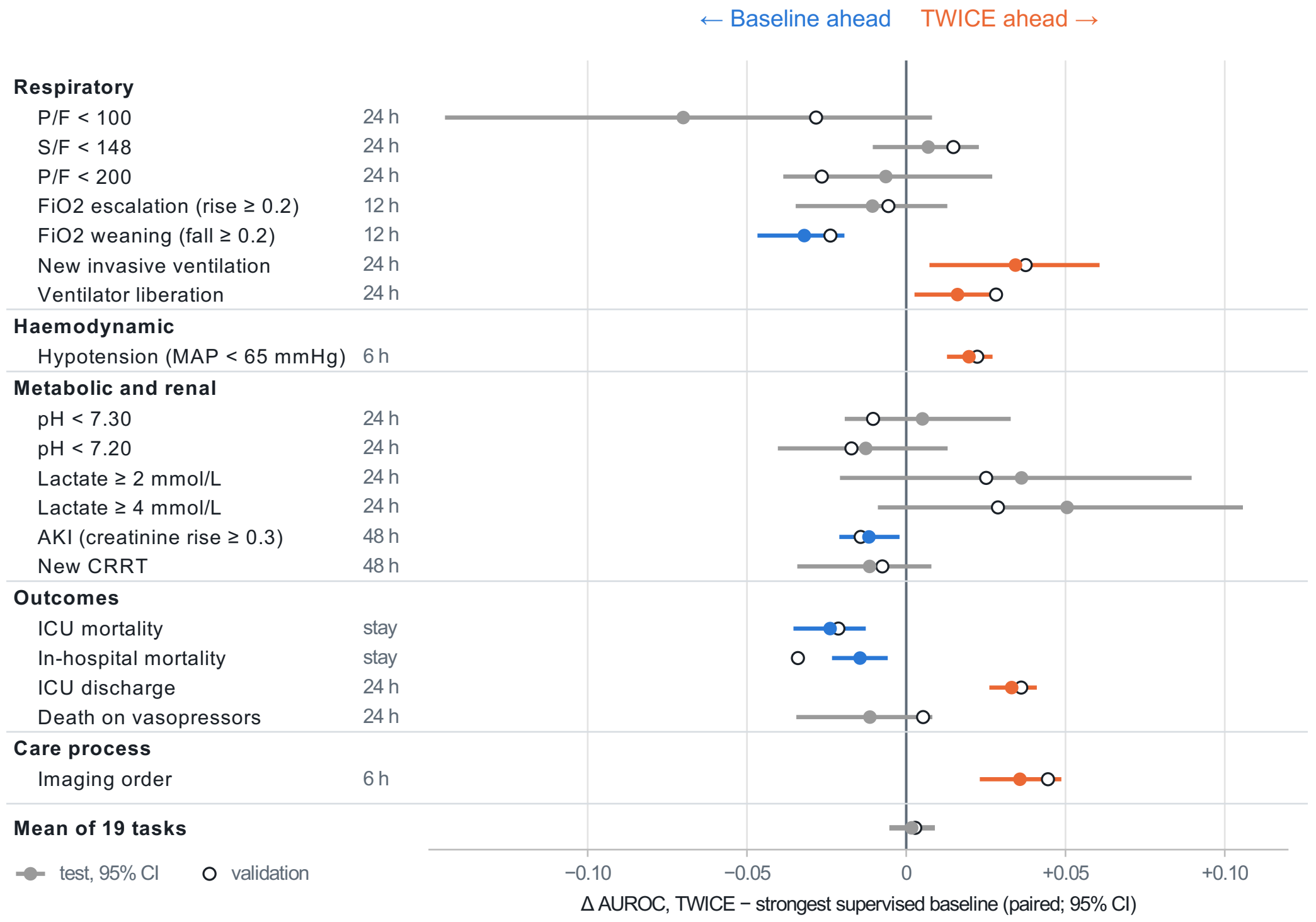

  \centering
  \figpage{2}{\textwidth}
  \caption{Clinical prediction tasks on test. Dots are the AUROC difference per task, \modelname{} minus the strongest supervised baseline, with 95\% CIs from 1,000 patient-clustered bootstrap replicates. Open circles are the same difference on validation. Orange marks a test CI above 0, blue a CI below 0 and grey a CI that includes 0. Holm's correction across the 19 tasks keeps seven of the nine coloured tasks, all but ventilator liberation and acute kidney injury. Each method's test AUROC is in Table~\ref{tab:full-auroc}. P/F is \pf{} and S/F is \sfr{}. Definitions are in Table~\ref{tab:tasks}.}
  \label{fig:l1}
\end{figure}

On the 19 test tasks, \modelname{}'s mean AUROC is 0.8362 and that of the strongest supervised baseline is 0.8345, a paired difference of $+0.0017$ (95\% CI $-0.0053$ to $+0.0090$; $p=0.64$; Table~\ref{tab:l1}). The two tie, although \modelname{} was trained on none of these labels and the strongest supervised baseline was trained on each, in the configuration that scored best of 144 on validation. Both scored slightly higher on test than on validation (0.8357 and 0.8329). Mean area under the precision--recall curve (AUPRC) is higher for \modelname{} (0.523 vs.\ 0.495), a paired difference of $+0.027$ (95\% CI $+0.013$ to $+0.041$). The Brier scores are similar (0.0892 and 0.0889 for the strongest supervised baseline), but \modelname{}'s expected calibration error is higher (0.030 vs.\ 0.012). \modelname{}'s probabilities are frequencies over 50 futures, not recalibrated against any label, while the strongest supervised baseline is fitted to each one. Reliability curves are in App.~\ref{app:full}.

The tie on the mean is not a tie on every task (Fig.~\ref{fig:l1} and App.~\ref{app:full}). \modelname{} leads on five tasks whose CIs exclude zero. They are imaging orders ($+0.036$), new invasive ventilation ($+0.034$), ICU discharge ($+0.033$), hypotension ($+0.020$) and ventilator liberation ($+0.016$). The strongest supervised baseline leads on four. They are FiO$_2$ weaning ($-0.032$), ICU mortality ($-0.024$), in-hospital mortality ($-0.015$) and acute kidney injury ($-0.012$). Seven of the nine survive Holm's correction across the 19 tasks. Ventilator liberation and acute kidney injury do not. The other ten tasks tie. \modelname{} leads on events that change the state of care within hours or a day (an imaging order, the start or end of ventilation, ICU discharge, a fall in blood pressure), and it trails on death, on oxygen weaning and on a rise in creatinine. The same five and four tasks clear zero, in the same directions, on validation, whose patients are disjoint from those in test.

\paragraph{Discharge diagnoses, from the same futures.} A future that reaches discharge writes the encounter's diagnoses, so the same 50 futures also predict which diagnosis groups (HCUP CCSR~\citep{hcup2026ccsr}) the stay will be recorded with (App.~\ref{app:dx}). We trained no supervised model for these groups. As a reference, the \emph{documented} rule uses only whether a group is already documented at the prediction time, on the problem list or from an earlier admission. It predicts the group's training frequency among patients with the same documentation status. Over the 116 groups common enough to score on test, \modelname{} reaches a macro AUROC of 0.717 on all of a stay's diagnoses and the documented rule reaches 0.742. \modelname{} has the higher macro AUPRC (0.174 vs.\ 0.141). On diagnoses not yet documented at the prediction time, where the rule has no information, \modelname{} reaches 0.693.

App.~\ref{app:baselines} gives the other baselines. They are 143 more supervised configurations, of which the best sequence model on validation reaches 0.782 on test, and 40 bedside-score configurations, of which the best on validation reaches 0.601 on test.

\begin{table}[t]
  \caption{External validation, as means over the 17 tasks scored at all three sites. Both models were trained on MGB and applied without retraining. Differences are \modelname{} minus the strongest supervised baseline, paired on identical rows, with 95\% CIs from 1,000 bootstrap replicates that resample patients (§\ref{sec:eval}). The MGB column is the test split restricted to the same 17 tasks. Per-task values are in App.~\ref{app:ext}.}
  \label{tab:ext}
  \centering
  \footnotesize
  \setlength{\tabcolsep}{4pt}
  \begin{tabular}{@{}lccc@{}}
  \toprule
  & MGB (test) & MIMIC-IV & eICU \\
  \midrule
  Scored stays & 8,995 & 5,245 & 5,000 \\
  Patients & 5,794 & 3,884 & 3,863 \\
  \midrule
  \multicolumn{4}{@{}l}{\emph{AUROC $\uparrow$}} \\
  \quad Strongest supervised baseline & 0.8291 & 0.7410 & 0.7336 \\
  \quad \modelname{} & 0.8296 & 0.7368 & 0.7627 \\
  \quad Difference & $+0.0005$ & $-0.0043$ & $+0.0291$ \\
  \quad\quad 95\% CI & [$-0.0074$, $+0.0088$] & [$-0.0175$, $+0.0082$] & [$+0.0113$, $+0.0476$] \\
  \midrule
  \multicolumn{4}{@{}l}{\emph{AUPRC $\uparrow$}} \\
  \quad Strongest supervised baseline & 0.5056 & 0.3797 & 0.3297 \\
  \quad \modelname{} & 0.5234 & 0.3812 & 0.3686 \\
  \quad Difference & $+0.0178$ & $+0.0015$ & $+0.0389$ \\
  \quad\quad 95\% CI & [$+0.0034$, $+0.0314$] & [$-0.0137$, $+0.0179$] & [$+0.0163$, $+0.0617$] \\
  \bottomrule
\end{tabular}

\end{table}

\paragraph{External validation.} We scored both models on 5,245 stays in MIMIC-IV (the 5,000 drawn stays, split at transfers between ICUs) and 5,000 in eICU (Table~\ref{tab:ext}). Over the 17 tasks scored at all three sites, \modelname{}'s mean AUROC is 0.830 on the MGB test split, 0.737 in MIMIC-IV and 0.763 in eICU. That of the strongest supervised baseline is 0.829, 0.741 and 0.734. In MIMIC-IV the two predict about as well, with a paired difference of $-0.0043$ (95\% CI $-0.0175$ to $+0.0082$). In eICU \modelname{} scores higher, by $+0.0291$ ($+0.0113$ to $+0.0476$) in AUROC and by $+0.0389$ ($+0.0163$ to $+0.0617$) in AUPRC. The strongest supervised baseline loses about as much AUROC as \modelname{} in MIMIC-IV and more in eICU. This suggests that the fall comes mostly from differences between the databases, which affect both models. After Holm's correction across the 17 tasks, \modelname{} trails on FiO$_2$ escalation in MIMIC-IV and leads on ICU discharge and in-hospital mortality in eICU. A pre-specified analysis without the two acidaemia tasks gives differences of $-0.0001$ and $+0.0257$ (App.~\ref{app:ext}). The tie at MGB thus holds in MIMIC-IV and becomes a lead in eICU.

\subsection{Readout compute, model size and context}\label{sec:levers}

\begin{figure}[t]
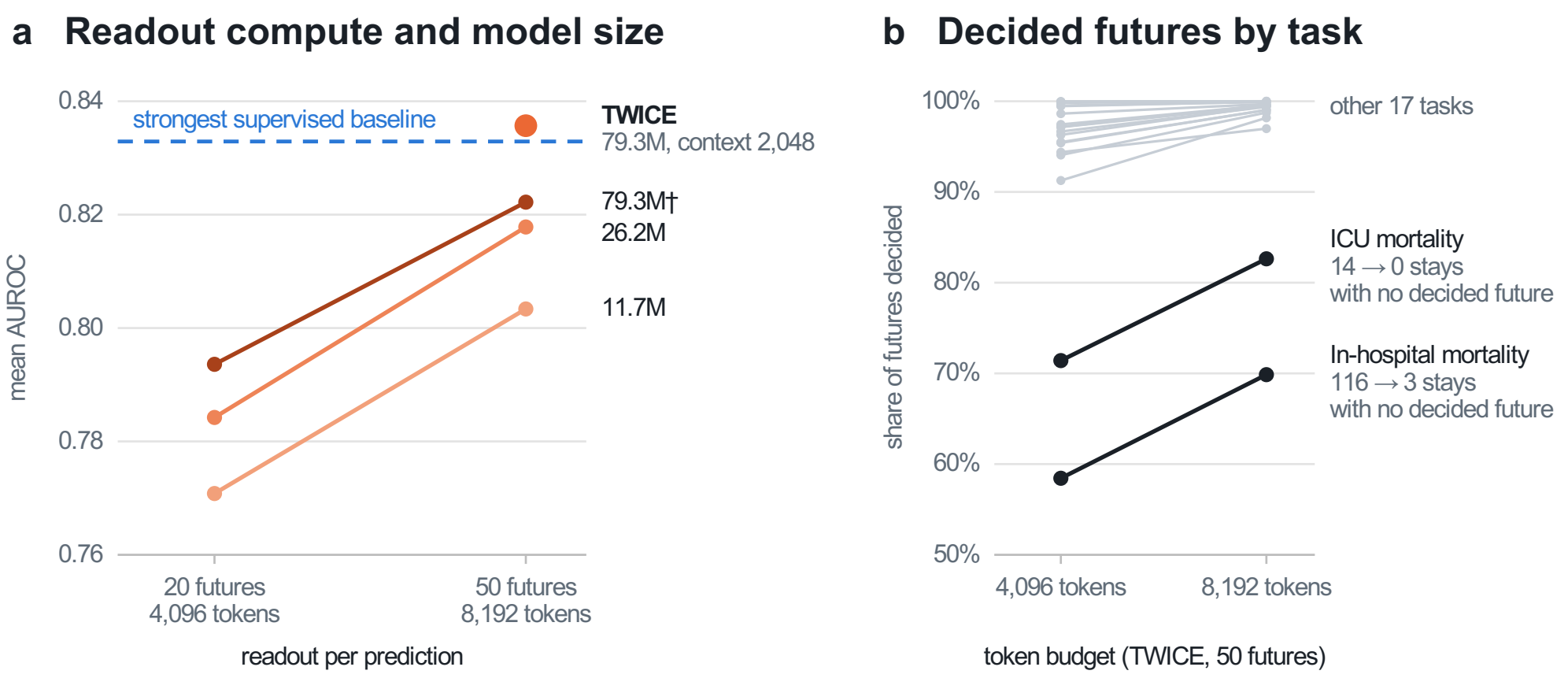

  \centering
  \figpage[1.7in]{3}{\textwidth}
  \caption{Readout compute, model size and context on validation. \textbf{(a)} Mean AUROC over the 19 tasks for each checkpoint, read with 20 futures of at most 4,096 tokens and with 50 futures of at most 8,192. The three lines are models trained at context 1,024. \modelname{} is the 79.3M-parameter model trained at context 2,048. $\dagger$ Read with 40 futures, as this model was not read with 50 futures of 8,192 tokens. \textbf{(b)} Share of \modelname{}'s 50 futures that decide each task, as a mean over the scored stays, at token budgets of 4,096 and 8,192. Grey lines are the other 17 tasks. The number beside each mortality task counts the scored stays with no decided future.}
  \label{fig:levers}
\end{figure}

All results in this section are on validation. Each comparison is paired on identical rows (App.~\ref{app:ablations}, Table~\ref{tab:ablations}).

\paragraph{Readout compute.} The same checkpoint scores higher when it samples more futures and lets each run longer (Fig.~\ref{fig:levers}a). CoMET found the first of these at scale, with gains flattening near 64 simulations~\citep{waxler2025comet}. Here we price both against model size, with the training data fixed. From 20 futures of at most 4,096 tokens to 50 of at most 8,192, the mean AUROC of an 11.7M-parameter model rose from 0.771 to 0.803, and that of a 26.2M model from 0.784 to 0.818. A 79.3M model with the same 1,024-token context rose from 0.794 to 0.822 at 40 futures of 8,192 tokens. Four to five times more readout compute was thus worth $+0.029$ to $+0.034$ to every checkpoint, against the $+0.023$ gained from 11.7M to 79.3M parameters at the cheapest readout. The smallest model read at the largest setting scored 0.803, and a model 6.8 times its size read at the smallest scored 0.794. To test where the gain from more futures flattens, we also read the 79.3M model at context 1,024 with 40 and 50 futures of 4,096 tokens. Going from 20 to 40 futures added $+0.022$. Going from 40 to 50 added only $+0.003$ (unpaired, Table~\ref{tab:ablations}), so the gain had nearly flattened by 50. Doubling the token budget alone, from 4,096 to 8,192 tokens at 50 futures, added $+0.009$ (95\% CI $+0.002$ to $+0.015$) to \modelname{}.

\paragraph{Why the budget matters.} A future that stops at the token budget before it decides a task leaves that task undecided (Eq.~\eqref{eq:readout}). At a budget of 4,096 tokens, 40.6\% of \modelname{}'s futures stopped at the budget. At 8,192, 28.5\% did. The 17 tasks with horizons of 2 days or less were decided in 91\% to 100\% of futures at either budget (Fig.~\ref{fig:levers}b). The tasks whose horizon is the whole stay were decided less often. With the longer budget, the share of futures that decided in-hospital mortality rose from 58\% to 70\%. The number of scored stays with no decided future, which receive the training rate, fell from 116 to 3. The longer budget helped these two tasks most. AUROC rose by $+0.032$ (95\% CI $+0.019$ to $+0.045$) for in-hospital and by $+0.035$ ($+0.019$ to $+0.051$) for ICU mortality, against $+0.006$ on average over the other 17 tasks. Censoring by the budget is informative. The futures decided within it are those of patients who die or leave soon. For \modelname{} this did not shift the mean predicted risk of in-hospital death, which was 1.12 times the training rate at both budgets. For the 79.3M model at context 1,024 it did. Its ratio fell from 1.08 to 1.00 as its budget doubled, so its short-budget futures had over-stated death. The budget is therefore part of the estimator, not only its cost, and it must be long enough for nearly every future to decide the question asked.

\paragraph{Model size.} Larger models scored higher. At 50 futures of 8,192 tokens, the mean AUROC was 0.803, 0.818 and 0.831 at 11.7M, 26.2M and 79.3M parameters. Each step added $+0.014$, and both 95\% CIs exclude zero (Table~\ref{tab:ablations}). Only the first step changes size alone. The 79.3M model was also trained at a longer context, with stronger regularisation and a longer early-stopping patience.

\paragraph{Context length.} Training at a longer context also helped, as it tended to in other EHR models~\citep{wornow2025context}. In matched pairs, a context of 2,048 tokens instead of 1,024 added $+0.012$ at 26.2M parameters and $+0.009$ at 79.3M (95\% CIs $+0.003$ to $+0.020$ and $+0.001$ to $+0.017$). A warm start that extended a converged 1,024-token model to 2,048 tokens added $+0.001$. The gain therefore came from training at that length. \modelname{} is the 79.3M model trained at context 2,048 without the stronger regularisation (0.836).

\paragraph{What did not help.} Ten other changes left the mean within 0.005 of their matched controls, except one (App.~\ref{app:ablations}, Table~\ref{tab:ablations}). They touched the input (a demographic prefix on every training window, so that a window starting mid-stay still sees the patient's age and sex, and training windows drawn per stay rather than per token), the value tokens (tail bins added to the deciles, log-spaced bins, 30 quantile bins, and an auxiliary head regressing the value, at two sizes), the training objective (same-minute events shuffled, and auxiliary heads predicting which codes occur in the next 24\,h, or when), and the regularisation above. The exception, log-spaced bins, scored $0.008$ lower, with a 95\% CI that includes zero ($-0.017$ to $+0.001$). With four bins across the bulk of a distribution, a threshold such as a creatinine rise of 0.3\,mg/dL falls inside a wide bin.

\subsection{Generation fidelity}\label{sec:l2}

\begin{figure}[t]
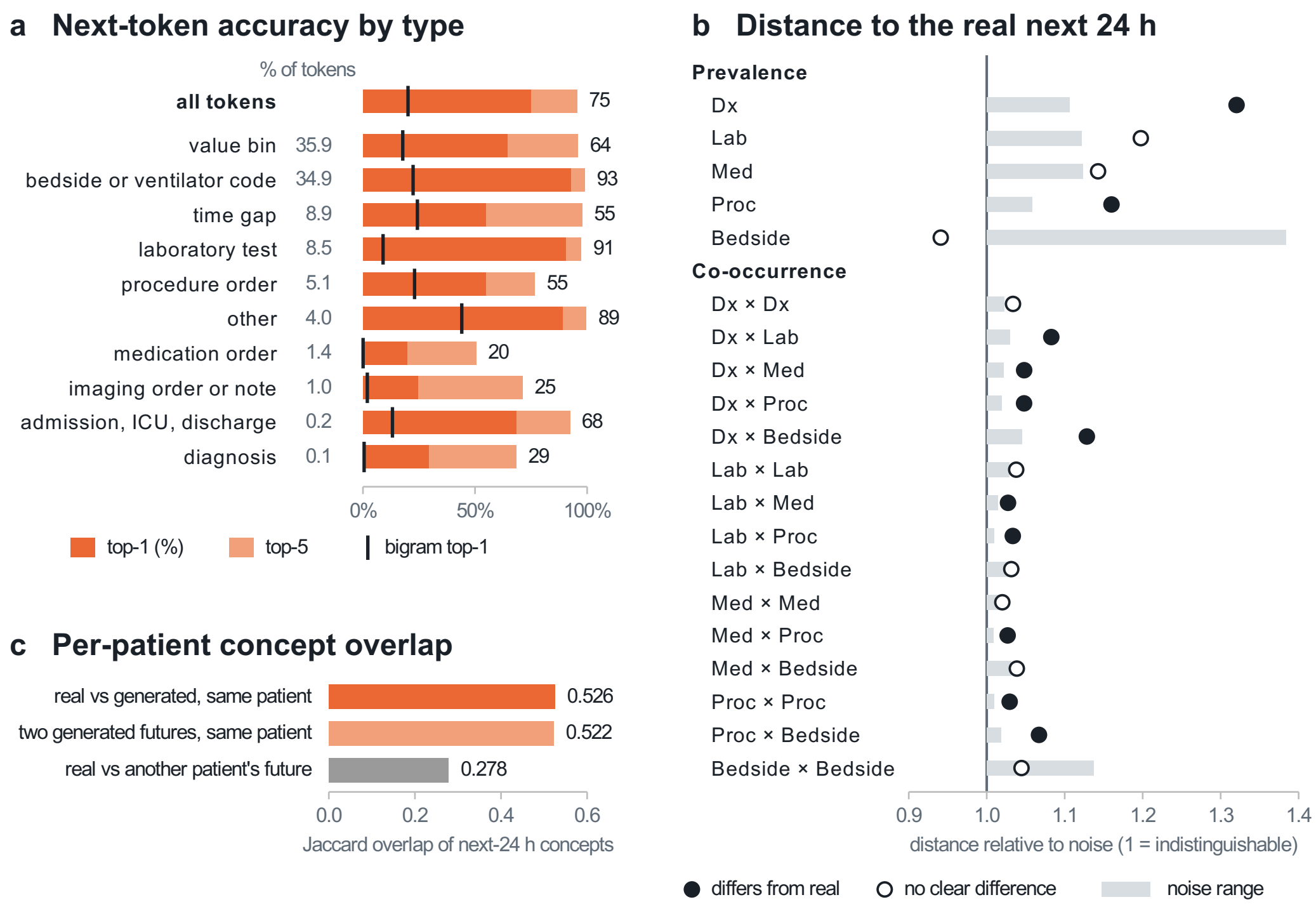

  \centering
  \figpage{4}{\textwidth}
  \caption{Generation fidelity on validation. \textbf{(a)} Next-token accuracy by the type of the true next token, at 571,002 random positions in 8,938 encounters. Bars show top-1 (with its value) and top-5, and the tick shows the bigram model's top-1. \textbf{(b)} The real next 24\,h against one generated future, per panel of concepts. Each dot is CoMET's RMSLE~\citep{waxler2025comet} divided by the median of 999 RMSLEs recomputed after swapping each patient's real and generated windows with probability one half, so that 1 is the noise level. The grey bar (noise range) runs from 1 to the 95th percentile of the swapped RMSLEs. Filled dots differ after Holm's correction across the 20 panels. Panel labels abbreviate bedside and ventilator codes (Bedside), diagnoses (Dx), laboratory tests (Lab), medications (Med) and procedures (Proc). \textbf{(c)} Per-patient Jaccard overlap of the concepts in the next 24\,h.}
  \label{fig:fidelity}
\end{figure}

We evaluated generation fidelity on the next token~\citep{shmatko2025delphi,zhu2026dttransformer}, on the orders placed after an abnormal value and on the next 24\,h of events~\citep{waxler2025comet} (Fig.~\ref{fig:fidelity}).

\paragraph{Next token.} At 571,002 positions drawn at random from 8,938 validation encounters (up to 64 per encounter), \modelname{} ranks the true next token first 75.0\% of the time (95\% CI 74.7 to 75.2) and among its first five 95.5\% of the time, against 20.1\% top-1 for a bigram model and 5.1\% for token frequencies. Accuracy follows how predictable each kind of event is from what came before. Bedside codes (92.6\%) and laboratory tests (90.6\%), which recur on schedules, are almost always placed first. Value bins (64.3\%) and time gaps (54.6\%) are placed first more than half the time and among the first five 96\% and 98\% of the time. Medication orders (19.8\%) and diagnoses (29.4\%), which turn on decisions not yet in the record, are among the hardest. Of the distinct event codes among the next five events, 54.2\% are among the model's ten most likely next tokens, against 29.6\% for the bigram model (set recall SR@(10,\,5)~\citep{pickard2026ehrmpc}).

\paragraph{Orders after an abnormal value.} Because a medication order is hard to place at the exact next position, we also asked whether the futures contain the order that clinicians placed after an abnormal value. When a high potassium, a low glucose or a low blood pressure was first charted (§\ref{sec:eval}), the futures sampled under usual care (§\ref{sec:intervene}) predict six orders about as well as a gradient-boosted model on summaries of the previous 24\,h, with overlapping 95\% CIs for all six (App.~\ref{app:fidelity}). For example, a vasopressor within 1\,h of a low blood pressure is predicted with an AUROC of 0.935 (0.910 to 0.954), against 0.900 (0.864 to 0.930). Four of the six orders appear in the futures at 0.4 to 0.5 times the rate at which clinicians placed them, and vasopressors at the same rate (0.14).

\paragraph{Generated versus real next 24\,h.} We compare the concepts in each patient's real next 24\,h with those in one generated window (App.~\ref{app:fidelity}). The statistic is the root mean squared logarithmic error (RMSLE) between real and generated rates of single concepts (prevalence) and of concept pairs (co-occurrence)~\citep{waxler2025comet}. We compare the observed RMSLE with 999 RMSLEs recomputed after swapping each patient's real and generated windows with probability one half. Each window is a single draw, so the RMSLE is above zero even when generated and real windows follow the same distribution. The median of the 999 swapped RMSLEs estimates this noise level, and a ratio of 1 means no difference beyond noise. After Holm's correction, 11 of 20 panels differ, by at most 1.32 times the noise level (Fig.~\ref{fig:fidelity}b). For co-occurrence the ratio is 1.02 to 1.13. Prevalence differs most for diagnoses, which enter the window only when the stay ends within it. Of 2,697 single concepts, 16 occur at a different rate in the generated 24\,h (12 laboratory tests, 3 procedures and 1 medication class). No diagnosis or bedside code differs.

\paragraph{Patient specificity.} A model could match these rates without using the patient's history. We compare each patient's real next 24\,h with one generated window by the Jaccard overlap of their concept sets (Fig.~\ref{fig:fidelity}c). The mean overlap is 0.526, against 0.278 when the real window is compared with a window generated for another patient~\citep{shmatko2025delphi} (permutation test $p=0.001$). Two windows generated for the same patient have a mean overlap of 0.522. The real window thus overlaps a generated one about as much as two generated windows overlap each other (sign-flip test $p=0.063$). The two overlaps are equal in expectation when the futures are draws from the right distribution.

\subsection{Individual trajectories}\label{sec:l3}

\begin{figure}[t]
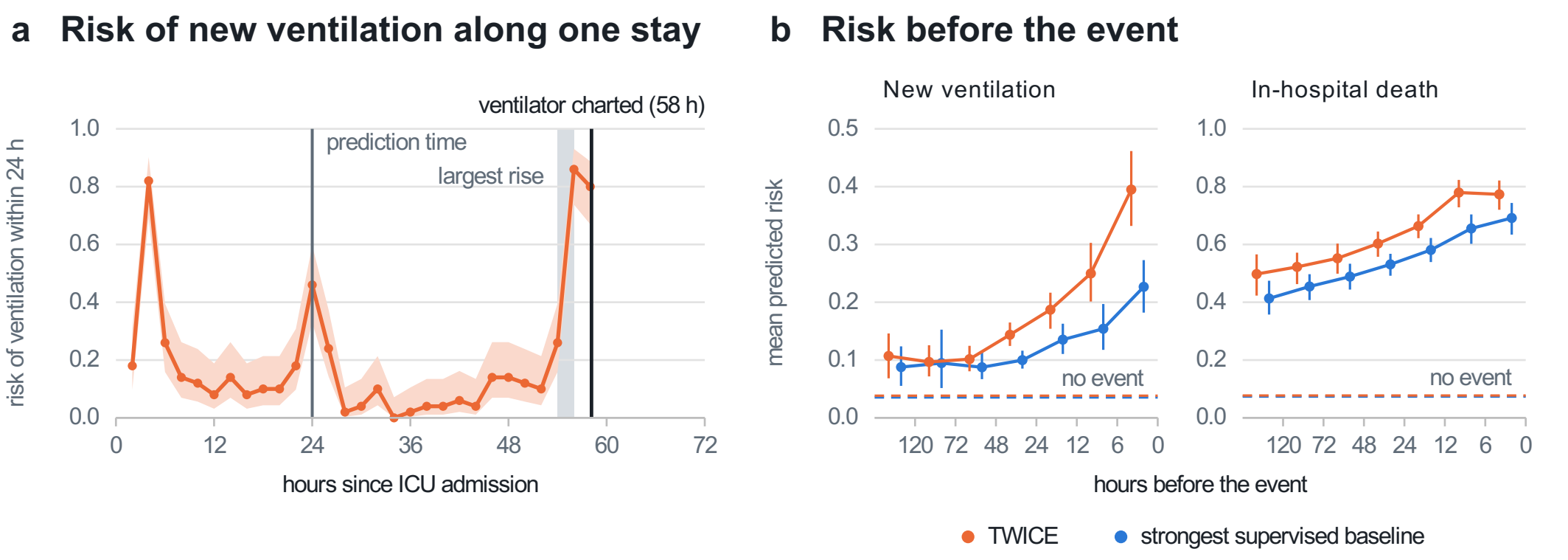

  \centering
  \figpage[1.98in]{5}{\textwidth}
  \caption{Individual trajectories on validation. \textbf{(a)} Risk of new ventilation within 24\,h along the first stay that progressed to ventilation, from 50 futures sampled every 2\,h, with a binomial 95\% CI. The grey interval is the largest rise after the prediction time. \textbf{(b)} Mean predicted risk by hours before the event, 150 stays per outcome, with patient-clustered 95\% CIs. Dashed lines are stays with neither event.}
  \label{fig:trajectories}
\end{figure}

We evaluated whether \modelname{} follows one patient through a stay~\citep{renc2025ares,shmatko2025delphi} (Fig.~\ref{fig:trajectories}). The two outcomes, new invasive ventilation and in-hospital death, were fixed before any output was read. \modelname{} leads the strongest supervised baseline on the first and trails it on the second (§\ref{sec:l1}).

\paragraph{Case studies.} We followed eight validation stays through their ICU course. Every 2\,h along each stay, \modelname{} was given the record up to that hour and sampled 50 futures, from which the risk at that hour is read (§\ref{sec:readout}). Stays were grouped by what happened after the prediction time. We drew at random three stays that progressed to ventilation, three that ended in death and two with neither event, before any output was read (App.~\ref{app:full}). In the first stay that progressed to ventilation, the risk of ventilation within 24\,h was 0.46 at the prediction time. It then stayed at or below 0.26 from 26\,h until 54\,h after ICU admission. It rose to 0.86 at 56\,h, about 2\,h before ventilation was charted (Fig.~\ref{fig:trajectories}a). Leaving out every event charted between 54 and 56\,h lowers the risk at 56\,h by 0.56 (95\% CI 0.40 to 0.72, App.~\ref{app:full}). Leaving out any one of the interval's 39 codes alone changes it by at most 0.06. In the other two stays that progressed to ventilation, leaving out the interval of the largest rise lowers the risk by 0.18 and 0.70, each with a CI above zero. In all three stays, the largest rise in risk can thus be traced to the events of one 2\,h interval. In the first stay, the rise came 2\,h before ventilation and may reflect care that had already begun. To protect the patient's privacy, Fig.~\ref{fig:trajectories}a shows no dates or measured values.

\paragraph{Risk before the event.} We also sampled futures every 12\,h along 150 validation stays that progressed to ventilation, 150 that ended in death and 148 with neither event (Fig.~\ref{fig:trajectories}b). Within 6\,h of ventilator charting, \modelname{}'s mean risk of new ventilation is 0.39 (95\% CI 0.33 to 0.46), against 0.14 (0.12 to 0.16) one to two days before and 0.038 in stays with neither event. The strongest supervised baseline gives 0.23, 0.10 and 0.035 on the same rows. Within 6\,h of death, \modelname{}'s risk of in-hospital death is 0.77 (0.72 to 0.82), against 0.50 (0.42 to 0.57) five days or more before and 0.077 in stays with neither event (strongest supervised baseline 0.69, 0.41 and 0.074). Both models give more risk as the event nears. \modelname{} gives more risk to imminent ventilation at a similar risk in stays without it. The same ordering holds at the prediction time on every scored stay (App.~\ref{app:full}).

\subsection{Counterfactual validation}\label{sec:l4}

\begin{figure}[t]
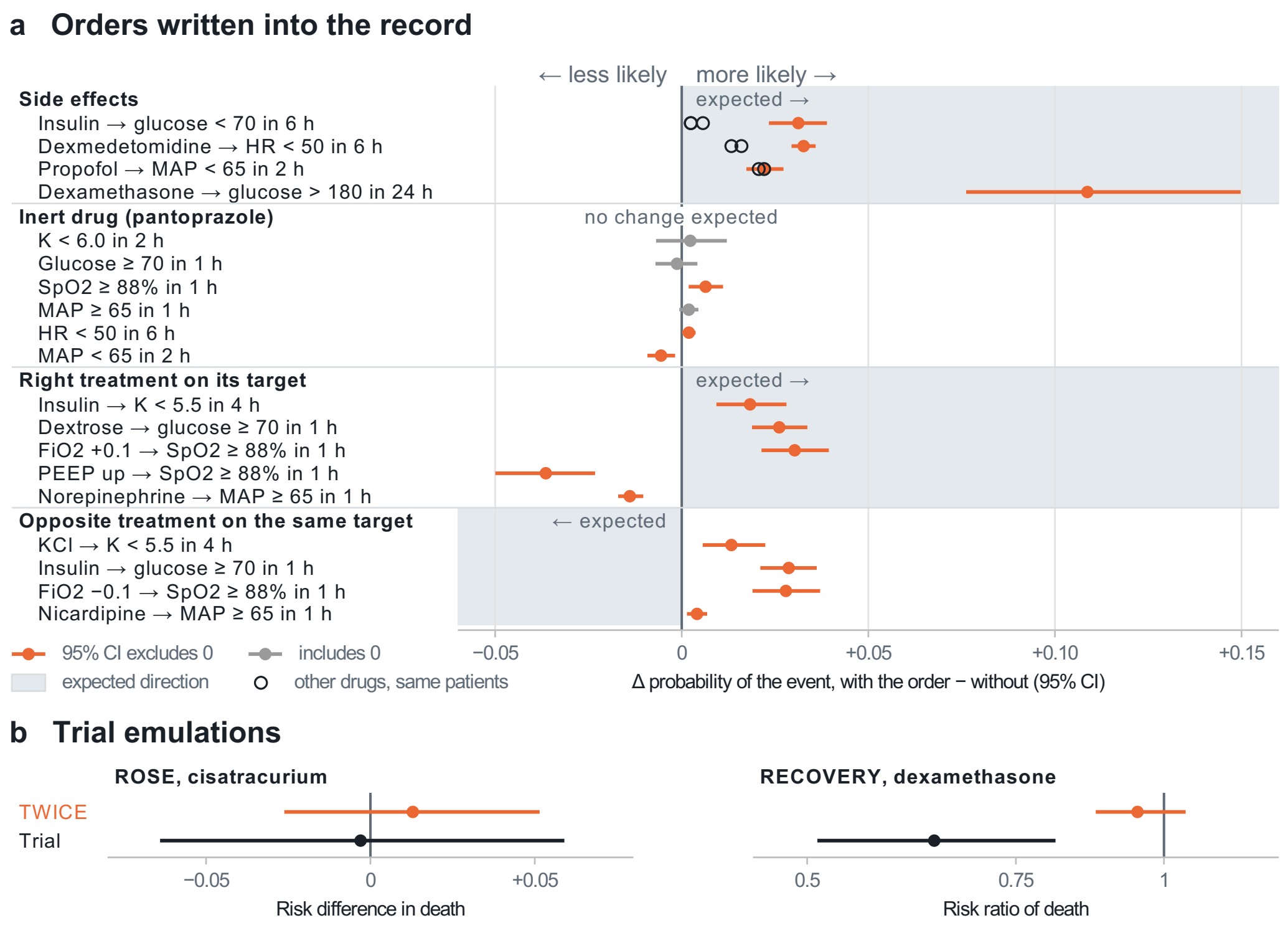

  \centering
  \figpage{6}{\textwidth}
  \caption{Counterfactual validation on validation. \textbf{(a)} Effect of each order on the probability of each event, against the control arm (§\ref{sec:intervene}), with patient-clustered 95\% CIs. The shaded side of zero is the pre-specified direction. Pantoprazole is expected to change nothing. Open circles show the other drugs tested in the same scenario. \textbf{(b)} In-hospital death in two trial emulations at the longest token budget (App.~\ref{app:cf}), \modelname{} against the published estimate~\citep{moss2019rose,recovery2021dexa}.}
  \label{fig:interventions}
\end{figure}

We evaluated whether \modelname{} responds correctly when a treatment is written into the EHR~\citep{pickard2026ehrmpc,lutsker2026healthformer,sainsbury2026flatascend}, in the seven clinical scenarios of §\ref{sec:eval}.

\paragraph{Side effects.} Clinicians rarely order a drug for its side effects, so a side effect is less likely than an intended effect to reflect why the drug was ordered~\citep{miettinen1983}. \modelname{} predicted all four pre-specified side effects, each with a 95\% CI above zero. Two of them were specific to the drug (Fig.~\ref{fig:interventions}a). At a high potassium, insulin raised the probability of glucose $<$70\,mg/dL within 6\,h by 0.031 (0.023 to 0.039). The other drugs tested there raised it by at most 0.006. At the start of ventilation, where the tested drugs are sedatives, dexmedetomidine raised the probability of a heart rate $<$50 within 6\,h by 0.033 (0.029 to 0.036), about twice as much as propofol or midazolam. Propofol raised the probability of a mean arterial pressure $<$65\,mmHg within 2\,h by 0.022 (0.017 to 0.027). Dexmedetomidine and midazolam raised it by as much. In the RECOVERY emulation, dexamethasone raised the probability of glucose $>$180\,mg/dL within 24\,h by 0.109 (0.076 to 0.150). These effects suggest that \modelname{} learned what these drugs do, not only when clinicians order them.

\paragraph{Inert drugs.} Pantoprazole is ordered to prevent stress ulcers in patients sick enough to be at risk~\citep{cook2024revise}. It is not expected to change potassium, glucose, blood pressure, oxygen saturation or heart rate within hours. A model that reads any order as a sign of illness could still move these values after it~\citep{lipsitch2010negative}. We therefore wrote it in as an extra treatment in every scenario. As expected, it barely moved the forecasts. It moved no primary outcome at a high potassium, a low glucose or a low blood pressure. Three of its effects had CIs that exclude zero. They were on oxygen saturation on the ventilator and on low blood pressure and low heart rate at the start of ventilation. Each was at most 0.006, against 0.022 to 0.033 for the side effects outside the trials. At a high potassium, IV calcium moved neither potassium outcome either. It protects the heart but is not expected to lower potassium~\citep{ukka2020hyperk}. Its effect on the first potassium below 5.5\,mmol/L within 4\,h was $+0.007$ ($-0.001$ to $+0.016$). \modelname{} thus did not read every order as a sign of illness.

\paragraph{Effects on the target.} The target of a treatment is the value it is given to correct, such as a high potassium. Four scenarios compare the right treatment, which guidelines recommend, with an opposite treatment, which is expected to push the target the other way. At a high potassium, for example, insulin is the right treatment and potassium chloride the opposite one. Insulin, dextrose and a higher FiO$_2$ improved their targets (Fig.~\ref{fig:interventions}a). In all four scenarios, however, the opposite treatment failed to make the target worse. At a high potassium, insulin raised the probability that the first potassium within 4\,h is below 5.5\,mmol/L by 0.018 (0.009 to 0.028). Potassium chloride, which adds potassium, raised it by 0.013 (0.006 to 0.022). Both effects are smaller on the pre-specified primary outcome (App.~\ref{app:cf}). At a low glucose and a low oxygen saturation on the ventilator, the opposite treatment also improved the target about as much as the right one. At a low blood pressure, norepinephrine lowered the probability that the first mean arterial pressure within 1\,h is at least 65\,mmHg ($-0.014$, $-0.017$ to $-0.010$). A PEEP increase at a low oxygen saturation also moved the target against the expected direction, although the oxygenation response to PEEP varies between patients~\citep{goligher2014peep}. These patterns are expected when an order is read as a sign of the patient's state, as in the vasopressor example of §\ref{sec:related}.

\paragraph{Measurement of the target.} We checked whether the gains after the opposite treatments reflect how outcomes on a measured value are counted. Such an outcome counts only the futures that measure the value within the window, as in the prediction tasks (§\ref{sec:readout}). An order can therefore change an outcome by changing which futures are counted, without changing the value itself~\citep{agniel2018biases}. At a low oxygen saturation on the ventilator, nearly every future measured SpO$_2$ within the hour (0.98 in every arm). Measurement therefore cannot explain why lowering FiO$_2$ raised SpO$_2$ as much as raising it did. At a low glucose, writing in dextrose or insulin made a glucose measurement within the hour more likely (0.48 and 0.41 of futures, against 0.35 in the control arm and with pantoprazole). Part of the glucose result may therefore reflect which futures are measured rather than the glucose value itself.

\paragraph{Observed outcomes.} Some patients in the five scenarios that start at a charted event received one of the tested treatments within 1 to 2\,h. For these patients we asked whether writing in the treatment they actually received predicts what followed better than futures with nothing written in or forbidden. Each pre-specified outcome is scored by the Brier score against what the EHR recorded (App.~\ref{app:cf}). Knowing the treatment did not improve the forecast. Across the 26 pairs of a treatment and an outcome, the gain in Brier score ranged from $-0.041$ to $+0.019$. Only one CI excluded zero, and it favoured the futures with nothing written in (midazolam on a heart rate $<$50, $-0.006$, $-0.014$ to $-0.001$). A written-in treatment moved the predicted probabilities by only a few hundredths (Fig.~\ref{fig:interventions}a), which limits how much it can change the Brier score.

\paragraph{Tabular estimators.} We also estimated the effects with two standard methods for observational data, fitted on the same scenarios in the training split (App.~\ref{app:cf}). The S-learner is a gradient-boosted model of the outcome given the history and the treatment, read with the treatment switched on and off~\citep{kunzel2019metalearners}. The other is inverse-probability weighting~\citep{hernan2020whatif}. The S-learner reversed both effects. After norepinephrine, it lowered the probability that the first mean arterial pressure within 1\,h is at least 65\,mmHg ($-0.147$, $-0.150$ to $-0.144$), against $-0.014$ for \modelname{}. After insulin, it lowered the probability that the first potassium within 4\,h is below 5.5\,mmol/L ($-0.110$, $-0.126$ to $-0.093$), where \modelname{} raised it by 0.018. Inverse-probability weighting pointed the same way in the 113 stays where both choices were plausible ($-0.227$, $-0.471$ to $+0.020$). Its CI includes zero. Neither method did better than \modelname{} on the targets.

\paragraph{Trial emulations.} \modelname{} agreed with the null trial, ROSE, and disagreed with the effective one, RECOVERY (Fig.~\ref{fig:interventions}b, App.~\ref{app:cf}). In ROSE, early neuromuscular blockade in moderate to severe acute respiratory distress syndrome (ARDS) did not change mortality at 90 days~\citep{moss2019rose}. In the emulation, cisatracurium changed the probability of in-hospital death by $+0.013$ ($-0.026$ to $+0.051$), within the trial's 95\% CI of $-0.064$ to $+0.059$~\citep{moss2019rose}. This agreement cannot be told apart from insensitivity, however. Every effect on death in the two emulations, pantoprazole included, lies within $\pm$0.025. A single order therefore barely moves \modelname{}'s predicted death. In the invasively ventilated stratum of RECOVERY, dexamethasone reduced 28-day mortality, with a rate ratio of 0.64 (0.51 to 0.81)~\citep{recovery2021dexa}. In the emulation, the risk ratio for in-hospital death was 0.95 (0.88 to 1.04). The two intervals do not overlap. The emulation reads COVID-19 from a diagnosis code recorded at discharge, and \modelname{} does not see SARS-CoV-2 test results. Both emulations also differ from the trials in endpoint, era and control arm. For example, 17\% of ROSE's usual-care patients received a neuromuscular blocker in the first 48\,h, which the control arm here forbids (App.~\ref{app:cf}).

\section{Discussion}\label{sec:discussion}

We asked how far a generative event model trained on the ICU records of one health system goes toward a digital twin of the ICU patient~\citep{nasem2024twins}. Without task-specific training, \modelname{} matched the strongest of 144 supervised configurations on 19 clinical prediction tasks on held-out patients. Its mean AUPRC was also higher. It led on the events that change care within hours or a day, such as the start of ventilation, a fall in blood pressure, ICU discharge and an imaging order. These leads appeared on validation and again on the separate test patients. Its generated records were close to real ones at the next token and over the next 24\,h (§\ref{sec:l2}). A patient's real future overlapped a generated one about as much as two generated futures overlapped each other. Its futures predicted the orders that clinicians placed after an abnormal value about as well as a gradient-boosted model. Its predicted risk rose as an event approached. Within 6\,h of ventilation, it gave more risk than the strongest supervised baseline. Because a twin is meant to inform decisions about care, we also tested how \modelname{} responds when a treatment is written into the EHR. It reproduced all four pre-specified side effects, two of them specific to the drug, and an inert drug barely moved its forecasts. Insulin, dextrose and a higher FiO$_2$ moved their targets in the expected direction. Two standard methods for observational data did no better on the targets. In all four scenarios that compare a right and an opposite treatment, however, the opposite treatment failed to make the target worse.

\paragraph{Target effects and confounding by indication.} An order in the EHR follows the clinician's reading of the patient, and much of that reading is not recorded. A potassium result judged falsely high from a haemolysed sample is one example (App.~\ref{app:cf}), and the patient's appearance at the bedside is another. A model trained to continue the EHR may therefore learn an order as a sign of the state that prompted it, which is confounding by indication (§\ref{sec:related}). A gradient-boosted S-learner, which sees the same history through tabular features, reversed the effects of both norepinephrine and insulin on their targets (§\ref{sec:l4}). This suggests that the failure arises from what observational records contain rather than from the architecture alone. We suggest that generative event models be evaluated with controls that can detect this failure, such as the side effects, inert drugs and opposite treatments of §\ref{sec:l4}. EHR-MPC and FlatASCEND insert treatments into generative models of sepsis care and of MIMIC-IV~\citep{pickard2026ehrmpc,sainsbury2026flatascend}. Our set of controls is a template on which such models can be compared.

\paragraph{Readout compute and its cost.} With the training data fixed, four to five times more readout compute raised the mean AUROC of every checkpoint as much as or more than a 6.8-fold larger model did (§\ref{sec:levers}). More simulated futures improve the estimates of generative event models~\citep{waxler2025comet,mccann2026continuous}, as test-time compute improves language models~\citep{snell2025testtime}. Here we compared this gain with the gain from model size. For \modelname{}, the readout took 29.8\,h for 8,947 scored validation stays on one NVIDIA H200, about 12\,s per stay for 50 futures. At this rate, the forecasts for an ICU could be refreshed every few hours. Estimators that read probabilities from the next-token distribution may reduce the cost further~\citep{solo2026scope}. A hospital that cannot pool its data with others can still increase its readout compute. At 79.3M parameters, \modelname{} had about 20 unique training tokens per parameter, close to Chinchilla's compute-optimal ratio of about 20~\citep{hoffmann2022chinchilla}, against 134 at 11.7M. It reached its best held-out loss after about 35 passes over the training tokens, well beyond the four or so passes for which repeated data were nearly as useful as new data~\citep{muennighoff2023data}. The token budget is also part of the estimator. CoMET right-censors futures that stop before the horizon and leaves futures that always reach it to future work~\citep{waxler2025comet}. We found that this censoring is informative (§\ref{sec:levers}). The budget should therefore be set for the question, and long-horizon questions need long budgets.

\paragraph{Shared futures across tasks.} The same 50 futures per patient answered the 19 tasks, the diagnosis groups and the fidelity tests. Futures sampled the same way, with an order written in, answered the counterfactual scenarios. Each future is one course of the stay, so the answers to different questions about a patient come from the same courses. Risks read from shared futures therefore respect the dependencies between events~\citep{renc2025ares}. A patient who dies in a future, for example, cannot also leave the ICU alive in it. Separately trained models for the 19 tasks carry no such constraint. The supervised baseline needed a labelled training set and a model for each task. \modelname{} needed neither, and it can be asked a new question without training.

\paragraph{Limitations.} This study has several limitations. First, \modelname{} was trained on the records of one health system. It was tested without retraining on samples of 5,000 stays from two public databases, whose records lack several of its inputs, such as note types and imaging orders. Both models lost accuracy there (§\ref{sec:l1}). Second, every modelling and readout choice was made on validation, which makes validation results optimistic. Only the results of §\ref{sec:l1} are on the test split, which was read once. Third, the evaluation has a limited scope. Tasks were excluded by stated rules. The single prediction time covers only patients still in the ICU after a day. Fourth, some events are recorded imprecisely. Ventilation is read from ventilator charting, which lags intubation. An ICU stay is one unit's admission, so a death after a transfer between ICUs is not an ICU death for the first stay (App.~\ref{app:tasks}). Medications are orders, not administrations. Fifth, values are represented by deciles, so a threshold inside a bin is bracketed rather than resolved. This limits how finely \modelname{} represents laboratory targets (App.~\ref{app:cf}). Sixth, probabilities are not recalibrated, and the expected calibration error is higher than that of the strongest supervised baseline. The readout also needs more compute than a supervised model. Finally, the simulated interventions estimate the effect of ``order now, then usual care'' under no unmeasured confounding and positivity. Counterfactual validation suggests that the first of these assumptions does not hold for target effects. Performance by subgroup is reported in App.~\ref{app:subgroups}.

\paragraph{Next steps.} We see three directions for further work. The first is to add notes and images. \modelname{} reads only structured events. It sees that a note was written or an image was ordered but not what the note says or what the image shows. Notes record part of the clinician's reading of the patient, and chest radiographs show the bilateral opacities that the Berlin definition of ARDS requires~\citep{ranieri2012berlin}. Adding them may reduce the confounding by indication described above. Autoregressive EHR models have been conditioned on notes, chest radiographs and ECGs, although these inputs did not always improve the prediction of ICU death~\citep{liu2026multimodal}. APOLLO places notes and images on the same timeline as structured events~\citep{zhang2026apollo}. The second is to change the training target. \modelname{} is trained to predict each next token, so it must also predict how care is documented, such as when a value is charted. Joint-embedding predictive architectures predict the representation of a hidden part of an image rather than its pixels, which may let a model leave out details that cannot be predicted~\citep{assran2023ijepa}. An ICU stay can also be viewed as a hidden Markov model. The hidden state is the patient's condition. The EHR and other data are observations of that state, and orders are actions that change it. Deep state-space models have represented EHR data in this way, with treatments as inputs to the change of state~\citep{krishnan2017sin}. A recent model of ICU patients predicts future states in a learned latent space~\citep{yang2026clinjepa}. Any model trained only on observational records may still face confounding by indication, so the third direction looks outside the EHR. A model can be anchored to trials, natural experiments and pharmacology, as HealthFormer was compared with published trials~\citep{lutsker2026healthformer}. Silent prospective use, in which forecasts are recorded before clinicians act, would show when such responses can support decisions.

We see this work as a step toward a digital twin of the ICU patient. \modelname{} forecasts ICU events about as well as the strongest supervised baseline, leads it on events that change care within a day and generates records close to real ones. It predicted about as well as that baseline or better on two public ICU databases it was never trained on. It also reproduces known side effects of drugs. Its predicted effects of treatments on their targets, however, do not yet separate the right treatment from the opposite one, so it is not yet a twin. Controls such as those of §\ref{sec:l4}, data from more health systems and prospective use will show when a model of this kind can inform decisions about care.

\section*{Ethics, data and code availability}
The MGB Institutional Review Board approved this retrospective study (protocol 2022P001232) and waived the requirement for informed consent. The MGB data contain protected health information and cannot be shared. MIMIC-IV and eICU are available from PhysioNet to credentialed researchers who sign its data use agreement. The code will be made publicly available, including the conversion and mapping of the external databases.

\section*{Author contributions}
Roles follow the CRediT taxonomy.\\
\textbf{Zhuoyang Lyu:} Conceptualization, Methodology, Software, Data curation, Formal analysis, Investigation, Visualization, Writing -- original draft, Writing -- review \& editing.\\
\textbf{Yiming Li:} Data curation, Supervision.\\
\textbf{Peter C. Hou:} Conceptualization, Supervision, Resources, Funding acquisition, Writing -- review \& editing.\\
\textbf{Pengyu Hong:} Conceptualization, Supervision, Resources, Funding acquisition, Writing -- review \& editing.\\
\textbf{Li Zhou:} Conceptualization, Supervision, Resources, Funding acquisition, Writing -- review \& editing.

\bibliographystyle{abbrvnat}
\bibliography{references}

\appendix

\section{Cohort}\label{app:data}
The data come from Epic Clarity through the MGB enterprise data warehouse. An ICU stay is one record in Epic's ICU location table (\texttt{F\_PAT\_ICU\_LOCATION}), so a transfer between two ICUs starts a new stay. The age and year criteria apply to the start of a stay. Because age is rounded to the year before the adult test, 293 stays (0.12\%) entered at an exact age between 17.45 and 18 years. Selection is on stays starting from 2017 to 2025, so 57 later unit admissions, all in encounters that entered the ICU in 2025, fall in 2026. For each encounter we took events from 24\,h before its first ICU admission to 24\,h after its last ICU discharge, so that deterioration before the ICU and the course after it are visible. Medications are orders, not administrations. After a name-based rescue of uncoded medications, 96.1\% of medication-order rows carry an ATC code~\citep{whoatc}. Conversion to MEDS removed 7,623,443 duplicate rows (0.79\%). The split is a deterministic hash of the patient identifier, so a re-cut cohort keeps every patient in the same split. Of the 236,906 ICU stays, 178,506 were still in the ICU 24\,h after admission and are scored. Of these, 8,947 are in validation (5,836 patients) and 8,995 in test (5,794).

Table~\ref{tab:cohort-full} gives the cohort by split. Race and ethnicity are shown as the record holds them. A patient with several race entries has one combined value. The most common such value, ``unavailable'' together with ``White'', is shown on its own, and the others are pooled. A category is pooled whenever a column would show fewer than 11 patients.

\begin{table}[ht]
  \caption{Cohort by split, full version of Table~\ref{tab:cohort}. $^a$One ICU unit admission in the record. $^b$Median [interquartile range]. $^c$Read from ventilator-setting markers (App.~\ref{app:tasks}). $^d$Age is rounded to the year before the adult test (§\ref{sec:data}). $^e$Encounters that passed our collaborators' screen for acute hypoxaemic respiratory failure. $^f$As recorded. Categories with fewer than 11 patients in any column are pooled.}
  \label{tab:cohort-full}
  \centering
  \scriptsize
  \begin{tabular}{@{}lrrrr@{}}
  \toprule
  & All & Train & Validation & Test \\
  \midrule
  Patients & 141,556 & 127,332 & 7,173 & 7,051 \\
  Hospital encounters & 185,244 & 166,745 & 9,195 & 9,304 \\
  ICU stays$^a$ & 236,906 & 213,265 & 11,789 & 11,852 \\
  Events (millions) & 959.0 & 863.5 & 48.0 & 47.5 \\
  Scored stays (in ICU at 24\,h) & 178,506 & 160,564 & 8,947 & 8,995 \\
  Patients with a scored stay & 115,295 & 103,665 & 5,836 & 5,794 \\
  Age at ICU admission, years$^b$ & 65.7 [53.8, 75.5] & 65.7 [53.8, 75.4] & 65.6 [53.3, 75.7] & 65.7 [53.3, 75.8] \\
  Stays at exact age $<$18 years$^d$ & 293 & 260 & 16 & 17 \\
  Female, \% & 43.7 & 43.7 & 43.7 & 43.8 \\
  ICU length of stay, days$^b$ & 2.0 [1.0, 4.2] & 2.0 [1.0, 4.2] & 2.0 [1.0, 4.3] & 2.1 [1.0, 4.2] \\
  Hospital length of stay, days$^b$ & 6.6 [3.4, 12.6] & 6.7 [3.4, 12.6] & 6.7 [3.4, 12.7] & 6.4 [3.3, 12.1] \\
  Invasive ventilation, \% of stays$^c$ & 43.4 & 43.5 & 43.5 & 42.3 \\
  In-hospital death, \% of encounters & 10.8 & 10.8 & 10.9 & 11.1 \\
  In-hospital death, \% of scored stays & 11.6 & 11.6 & 11.6 & 12.1 \\
  ICU death, \% of scored stays & 6.3 & 6.3 & 6.3 & 6.8 \\
  AHRF screen passed, \% of scored stays$^e$ & 51.6 & 51.7 & 51.5 & 50.1 \\
  \multicolumn{5}{@{}l}{Race, \% of patients$^f$} \\
  \quad White & 78.2 & 78.2 & 78.6 & 78.7 \\
  \quad Black or African American & 7.0 & 7.0 & 6.8 & 6.7 \\
  \quad Not listed & 5.2 & 5.2 & 5.1 & 5.0 \\
  \quad Asian & 3.2 & 3.2 & 3.1 & 3.3 \\
  \quad Unavailable & 3.1 & 3.1 & 3.1 & 3.2 \\
  \quad Unavailable and White (two entries) & 0.9 & 0.9 & 0.9 & 0.8 \\
  \quad Other, several or missing & 2.4 & 2.4 & 2.4 & 2.4 \\
  \multicolumn{5}{@{}l}{Ethnicity, \% of patients$^f$} \\
  \quad Not Hispanic & 85.8 & 85.8 & 85.9 & 85.7 \\
  \quad Hispanic & 7.9 & 7.9 & 7.9 & 7.8 \\
  \quad Unavailable & 5.5 & 5.5 & 5.4 & 5.6 \\
  \quad Declined & 0.6 & 0.6 & 0.5 & 0.6 \\
  \quad Other or missing & 0.2 & 0.2 & 0.3 & 0.3 \\
  \bottomrule
\end{tabular}

\end{table}

\section{Tasks}\label{app:tasks}
Labels are built once from the MEDS events by a registry of task definitions and shared by every method. A task combines a template (the first occurrence of an event, a measured value beyond a bound, a rise or fall against the patient's own last value, or the end of a support) with a horizon and an eligibility rule (Table~\ref{tab:tasks}). \emph{New}: no event earlier in the encounter. \emph{Not active}: no event, or no value beyond the bound, in the previous 24\,h (12\,h for imaging). \emph{Active}: at least one in the previous 24\,h. \emph{All}: every scored stay. The \pf{} and \sfr{} tasks use the last FiO$_2$ setting charted in the same ICU stay, carried forward, as the band tokens do (§\ref{sec:tokens}).

Two rules set the reported list. A task needs at least 50 positive rows in validation. This removes ECMO, tracheostomy and death among patients given corticosteroids. A task's label must also be computable from structured data for every stay. This removes ARDS onset. Its label rests on our collaborators' NLP-based adjudication of the Berlin criteria, which covers only encounters that passed a screen for acute hypoxaemic respiratory failure (App.~\ref{app:ards}). Both rules depend only on properties of the labels. Both were fixed before \modelname{} was scored on test.

A row is not scored for a threshold task when the series was not measured in the window. When the patient leaves the setting in which the event would be recorded before the window closes (the ICU for bedside measurements and ventilation, the hospital for orders and death), the row is labelled negative. On a 10,000-patient development subset, scoring such rows as undefined instead changed the bedside scores' AUROC by at most 0.029. Three definitions carry known biases. Ventilation has no intubation marker in the record. It is read from the first charting of ventilator settings, which lags intubation. ICU mortality closes at the end of the ICU stay, which is one unit's admission (App.~\ref{app:data}), so a patient who is transferred to another ICU and dies there is negative for the first stay. This lowers ICU mortality relative to in-hospital mortality (6.3\% and 11.6\% of scored stays). And the \sfr{} band $<$148 is the severe band of the global ARDS definition~\citep{matthay2024global}, which applies it to intubated patients with SpO$_2\leq$97\%. Our series applies neither restriction. Severe hypoxaemia by \pf{} has 49 positive rows on test, one fewer than the validation threshold, and stays in the suite because the list was fixed on validation. Counts in Table~\ref{tab:tasks} are scored rows (positives).

\begin{table}[h]
  \caption{The 19 reported tasks and the four excluded by rule.}
  \label{tab:tasks}
  \centering
  \scriptsize
  \begin{tabularx}{\textwidth}{@{}l>{\raggedright\arraybackslash}Xlrr@{}}
    \toprule
    Task & Definition & Eligible & Val $n$ (pos) & Test $n$ (pos) \\
    \midrule
    \multicolumn{5}{@{}l}{\emph{Respiratory}} \\
    Severe hypoxaemia & \pf{} $<100$ within 24\,h & not active & 1,022 (62) & 1,037 (49) \\
    Severe hypoxaemia, \sfr{} & \sfr{} $<148$ within 24\,h & not active & 1,591 (412) & 1,555 (404) \\
    Moderate hypoxaemia & \pf{} $<200$ within 24\,h & not active & 615 (135) & 631 (138) \\
    FiO$_2$ escalation & FiO$_2$ rises $\geq$0.20 over its last value in the previous 24\,h, within 12\,h & all & 2,817 (491) & 2,892 (456) \\
    FiO$_2$ weaning & FiO$_2$ falls $\geq$0.20 below its last value in the previous 24\,h, within 12\,h & all & 2,817 (230) & 2,892 (244) \\
    New invasive ventilation & first ventilator-setting marker (set tidal volume, set rate, peak or plateau pressure, mode) within 24\,h & new & 3,419 (226) & 3,549 (213) \\
    Ventilator liberation & ventilator markers stop for $\geq$48\,h, the stop starting within 24\,h & active & 3,825 (734) & 3,798 (677) \\
    \multicolumn{5}{@{}l}{\emph{Haemodynamic}} \\
    Hypotension & mean arterial pressure $<65$\,mmHg within 6\,h & all & 7,319 (1,690) & 7,250 (1,700) \\
    \multicolumn{5}{@{}l}{\emph{Metabolic and renal}} \\
    Acidaemia & pH $<7.30$ within 24\,h & not active & 1,751 (501) & 1,707 (475) \\
    Severe acidaemia & pH $<7.20$ within 24\,h & not active & 2,252 (511) & 2,260 (478) \\
    Hyperlactataemia & lactate $\geq$2\,mmol/L within 24\,h & not active & 922 (181) & 883 (155) \\
    Severe hyperlactataemia & lactate $\geq$4\,mmol/L within 24\,h & not active & 1,565 (80) & 1,548 (74) \\
    Acute kidney injury & creatinine rises $\geq$0.3\,mg/dL over its last value in the previous 48\,h, within 48\,h & all & 8,326 (1,079) & 8,344 (1,067) \\
    New CRRT & first CRRT marker within 48\,h & new & 8,567 (102) & 8,587 (94) \\
    \multicolumn{5}{@{}l}{\emph{Outcomes}} \\
    ICU mortality & death before ICU discharge & all & 8,944 (560) & 8,994 (611) \\
    In-hospital mortality & death before hospital discharge & all & 8,947 (1,041) & 8,993 (1,092) \\
    ICU discharge & ICU discharge within 24\,h & all & 8,947 (2,907) & 8,995 (2,902) \\
    Death on vasopressors & death within 24\,h, among patients with a vasopressor order in the previous 24\,h & active & 3,455 (100) & 3,440 (97) \\
    \multicolumn{5}{@{}l}{\emph{Care process}} \\
    Imaging order & imaging or echocardiography order within 6\,h & not active (12\,h) & 5,274 (833) & 5,342 (833) \\
    \midrule
    \multicolumn{5}{@{}l}{\emph{Excluded by rule}} \\
    ARDS onset & adjudicated onset within 48\,h; label provenance (App.~\ref{app:ards}) & not active & 928 (75) & -- \\
    Tracheostomy & tracheostomy order within 48\,h; $<$50 val positives & new & 8,810 (31) & -- \\
    ECMO & first ECMO marker within 48\,h; $<$50 val positives & new & 8,786 (18) & -- \\
    Death on corticosteroids & death within 24\,h, corticosteroid in previous 24\,h; $<$50 val positives & active & 1,780 (44) & -- \\
    \bottomrule
  \end{tabularx}
\end{table}

\section{Tokenisation}\label{app:tokens}
\paragraph{Vocabulary.} A code is kept as its own token if it is recorded for at least 0.5\% of training patients, counted over the whole training split. Otherwise it is mapped to the token of its broader parent code (a laboratory test to its LOINC analyte, a diagnosis to its ICD-10 category). A code without a parent, or whose parent is also rare, is mapped to the ``other'' token for its family, which carries presence only. Static fields, admission, ICU and discharge events, oxygenation, vital signs, ventilator settings and organ support are never pruned. Pruning acts on laboratory, medication, diagnosis, procedure, imaging, echocardiography and note codes. Encounter diagnoses, history diagnoses and medications are written hierarchically. An ICD-10 code is written as up to three tokens (characters 1--3, 4--5 and 6--7), and a medication as up to three tokens of its ATC code. The identity of the ICU unit is left out of the stream, and exact duplicate tokens at one instant are dropped. The vocabulary has 4,986 tokens.

\paragraph{Values and time.} A numeric value becomes the decile of its (code, unit) among training values, with the same edges as the tabular baseline's value-bin encoding (App.~\ref{app:baselines}). One set of ten value tokens serves every code. The record is stamped to the minute, so gaps under a minute emit no time token. Longer gaps use the time intervals of §\ref{sec:tokens}, whose open last interval is repeated at most eight times.

\paragraph{Order and bands.} Within one minute, events follow a fixed order by kind: openers (admission, ICU admission), observations, derived values, actions (medications, procedures), documentation, diagnoses, and closers (ICU discharge, death, discharge). Within a kind, codes are alphabetical. At every instant where an oxygenation ratio can be graded, a band token names its band: \pf{} $\geq$300, 200--300, 100--200 or $<$100, and \sfr{} $\geq$148 or $<$148. The ratio uses the last FiO$_2$ setting charted in the same ICU stay, as the labels do. The stream also carries a token at ARDS onset from our collaborators' adjudication of the Berlin criteria (App.~\ref{app:ards}). Only encounters that passed the screen for acute hypoxaemic respiratory failure were adjudicated, so the token's absence does not mean that ARDS did not occur.

\section{Model and training}\label{app:train}
Table~\ref{tab:train} gives \modelname{}'s architecture and training settings. A training window may run from the end of one encounter into the next, with no mask between them. A prompt holds the current encounter only. The smaller models of §\ref{sec:levers} have 6 layers of width 384 (11.7M parameters) and 8 layers of width 512 (26.2M). They were trained with the same recipe at context 1,024 with 128 windows per step, the same number of tokens per step, and an early-stopping patience of 10. The regularised 79.3M model of §\ref{sec:levers} differs from \modelname{} only in weight decay (0.1) and dropout (0.2), and trained for the full 500,000 steps.
\begin{table}[h]
  \caption{\modelname{} architecture and training.}
  \label{tab:train}
  \centering
  \small
  \begin{tabularx}{\textwidth}{@{}l>{\raggedright\arraybackslash}X@{}}
    \toprule
    Layers, width, feed-forward width & 12, 768, 2,048 (SwiGLU) \\
    Attention heads & 12 query, 4 key--value (grouped-query) \\
    Positions & rotary, $\theta=10{,}000$; context 2,048 tokens \\
    Normalisation & RMSNorm (pre-norm), $\epsilon=10^{-5}$ \\
    Embeddings & tied input and output; vocabulary 4,986 \\
    Parameters & 79.3M \\
    Objective & next-token cross-entropy on all tokens \\
    Windows & 2,048 tokens from a uniform token offset in the concatenated training sequences \\
    Optimiser & AdamW~\citep{loshchilov2019adamw}, $\beta=(0.9, 0.95)$, weight decay 0.01 \\
    Batch & 64 windows (131,072 tokens) per step \\
    Learning rate & $10^{-3}$ peak, 50,000 warm-up steps, cosine to $10^{-4}$ at 500,000 steps \\
    Dropout & 0.1 \\
    Precision & bfloat16 \\
    Checkpoint selection & held-out cross-entropy on 4,096 fixed windows from 5\% of training encounters, every 1,000 steps, patience 20 \\
    Stopped at & step 435,000 (57.0B tokens, 36.4 epochs), best at 415,000 (0.630 nats) \\
    Compute & about 32\,h on one NVIDIA H200 (three 12-hour segments, resumed exactly) \\
    \bottomrule
  \end{tabularx}
\end{table}

\section{Readout and its engineering}\label{app:readout}
Futures are sampled at temperature 1 from the full next-token distribution, without top-$k$ or nucleus truncation. The temperature was fixed on the fidelity of what the model generates, not on AUROC. With 50 futures, a probability is a multiple of $1/n$, where $n\leq50$ is the number of decided futures. For a threshold that falls inside a decile bin, the lower bound counts only the bins wholly beyond the threshold and the upper bound adds the bin that contains it. The midpoint is reported.

Sampling uses a ring-buffer key--value cache with CUDA graphs. Each new token costs one forward pass of one position, and the oldest token's keys and values are evicted once 2,048 are held. This is sliding-window attention over a model trained on full windows. Against an exact reference that re-encodes the last 2,048 tokens at every step, per-step token probabilities differ by about $10^{-3}$, and the 26.2M model's mean AUROC changed by $+0.001$ (20 tasks: $+0.0043$, $-0.0034$ to $+0.0128$; Table~\ref{tab:ablations}). Simulating the 8,947 scored validation stays at 50 futures of 8,192 tokens took 29.8\,h on one NVIDIA H200, and the 8,995 test stays 53.5\,h on one NVIDIA L40S. Runs resume exactly across 12-hour job segments.

\section{All baselines}\label{app:baselines}
Every baseline is scored on the same stays as \modelname{}, and every number in Tables~\ref{tab:base-tab}--\ref{tab:base-e0} is a mean AUROC over the 19 tasks. The supervised baselines are trained per task on the training split. Their inputs come from four nested variable sets: \texttt{essential} (static data, oxygenation and routine vital signs; 55 codes), \texttt{essentialTier2} (adds ventilator mechanics; 78), \texttt{extended} (adds organ support and every laboratory test; 1,292) and \texttt{full} (adds medication, procedure and imaging orders and diagnoses; 4,675).
\begin{itemize}
  \item \emph{Tabular} (Table~\ref{tab:base-tab}, 48 configurations). The events of the ICU stay in the 24\,h before $t$ become presence indicators; presence plus the count, last, minimum, maximum and mean of each code (adapted from \citealp{harutyunyan2019multitask}); or presence plus value bins (the deciles of §\ref{sec:tokens}). Each encoding also carries age, hours since ICU admission and the stay's order in the encounter, and these three alone form a fourth, time-only encoding. Each is fitted by LightGBM~\citep{ke2017lightgbm}, XGBoost~\citep{chen2016xgboost} or L2-regularised logistic regression. Tree counts come from early stopping and the L2 strength from a grid, both on validation. Every other hyperparameter is the library default.
  \item \emph{Sequence} (Table~\ref{tab:base-seq}, 96 configurations). A GRU~\citep{cho2014gru} or a transformer encoder~\citep{vaswani2017attention} reads the same 24\,h as 1\,h or 4\,h bins, each bin holding presence, z-scored values or value bins. It is trained as one model per task or as one shared trunk with a head per task: hidden size 128, one layer, dropout 0.3, AdamW at learning rate $10^{-3}$, stopping after 5 evaluations without improvement on validation.
  \item \emph{Bedside scores} (Table~\ref{tab:base-e0}). A partial SOFA (3 of its 6 components: respiratory, renal, and cardiovascular capped at 1 point, because the record holds vasopressor orders rather than doses; the neurological, coagulation and liver components were not computed), \pf{}, \sfr{}, creatinine and mean arterial pressure, each at $t$ or as the worst value in the previous 24\,h. A two-parameter logistic fit on the training split sets each score's direction and scale for each task. The best on validation, the partial SOFA as its worst value in the previous 24\,h, reaches 0.607 on validation and 0.601 on test.
\end{itemize}
The strongest supervised baseline is the best of the 144 supervised configurations on validation (bold in Table~\ref{tab:base-tab}). It is LightGBM on summaries of the \texttt{full} set, with 0.8329 on validation and 0.8345 on test. The best sequence configuration on validation, a transformer over z-scored 4\,h bins of the \texttt{extended} set, reaches 0.7823 on test, 0.052 below the strongest supervised baseline and 0.054 below \modelname{} (paired 95\% CI $0.044$ to $0.063$). Because early stopping and the choice of hyperparameters used the validation split, the supervised baselines' validation scores are slightly optimistic. No choice used the test split.

\begin{table}[ht]
  \caption{Tabular baselines: mean AUROC over the 19 tasks, by variable set. Bold: best on validation, the strongest supervised baseline.}
  \label{tab:base-tab}
  \centering
  \scriptsize
  \setlength{\tabcolsep}{4pt}
  \begin{tabular}{@{}lcccccccc@{}}
  \toprule
  & \multicolumn{2}{c}{\texttt{essential}} & \multicolumn{2}{c}{\texttt{essentialTier2}} & \multicolumn{2}{c}{\texttt{extended}} & \multicolumn{2}{c}{\texttt{full}} \\
  \cmidrule(lr){2-3}\cmidrule(lr){4-5}\cmidrule(lr){6-7}\cmidrule(lr){8-9}
  Configuration & val & test & val & test & val & test & val & test \\
  \midrule
  LightGBM, time in stay only & 0.5477 & 0.5503 & 0.5477 & 0.5503 & 0.5477 & 0.5503 & 0.5477 & 0.5503 \\
  XGBoost, time in stay only & 0.5465 & 0.5428 & 0.5468 & 0.5444 & 0.5461 & 0.5449 & 0.5480 & 0.5436 \\
  logistic regr., time in stay only & 0.5429 & 0.5449 & 0.5429 & 0.5448 & 0.5430 & 0.5448 & 0.5430 & 0.5450 \\
  LightGBM, presence & 0.6796 & 0.6690 & 0.6878 & 0.6766 & 0.7360 & 0.7301 & 0.7770 & 0.7749 \\
  XGBoost, presence & 0.6815 & 0.6666 & 0.6864 & 0.6765 & 0.7294 & 0.7261 & 0.7673 & 0.7632 \\
  logistic regr., presence & 0.6684 & 0.6636 & 0.6756 & 0.6701 & 0.7207 & 0.7218 & 0.7635 & 0.7679 \\
  LightGBM, summaries & 0.7708 & 0.7709 & 0.7776 & 0.7730 & 0.8158 & 0.8169 & \textbf{0.8329} & \textbf{0.8345} \\
  XGBoost, summaries & 0.7632 & 0.7656 & 0.7681 & 0.7654 & 0.8081 & 0.8040 & 0.8200 & 0.8236 \\
  logistic regr., summaries & 0.7524 & 0.7485 & 0.7555 & 0.7527 & 0.7691 & 0.7715 & 0.7847 & 0.7913 \\
  LightGBM, value bins & 0.7392 & 0.7375 & 0.7473 & 0.7434 & 0.7880 & 0.7849 & 0.8066 & 0.8090 \\
  XGBoost, value bins & 0.7372 & 0.7343 & 0.7434 & 0.7395 & 0.7784 & 0.7765 & 0.8007 & 0.7980 \\
  logistic regr., value bins & 0.7323 & 0.7320 & 0.7373 & 0.7343 & 0.7609 & 0.7648 & 0.7810 & 0.7879 \\
  \bottomrule
\end{tabular}

\end{table}

\begin{table}[ht]
  \caption{Sequence baselines: mean AUROC over the 19 tasks, by variable set. Bold: best on validation.}
  \label{tab:base-seq}
  \centering
  \scriptsize
  \setlength{\tabcolsep}{4pt}
  \begin{tabular}{@{}lcccccccc@{}}
  \toprule
  & \multicolumn{2}{c}{\texttt{essential}} & \multicolumn{2}{c}{\texttt{essentialTier2}} & \multicolumn{2}{c}{\texttt{extended}} & \multicolumn{2}{c}{\texttt{full}} \\
  \cmidrule(lr){2-3}\cmidrule(lr){4-5}\cmidrule(lr){6-7}\cmidrule(lr){8-9}
  Configuration & val & test & val & test & val & test & val & test \\
  \midrule
  GRU, presence, 4\,h, per task & 0.7152 & 0.7072 & 0.7197 & 0.7140 & 0.6854 & 0.6858 & 0.6445 & 0.6353 \\
  GRU, presence, 4\,h, multi-task & 0.7182 & 0.7105 & 0.7166 & 0.7107 & 0.6794 & 0.6739 & 0.6202 & 0.6131 \\
  GRU, presence, 1\,h, per task & 0.7149 & 0.7077 & 0.7150 & 0.7079 & 0.6477 & 0.6407 & 0.5808 & 0.5696 \\
  GRU, presence, 1\,h, multi-task & 0.7179 & 0.7142 & 0.7150 & 0.7122 & 0.6306 & 0.6215 & 0.5497 & 0.5457 \\
  GRU, z-score, 4\,h, per task & 0.7732 & 0.7685 & 0.7727 & 0.7728 & 0.7397 & 0.7365 & 0.6711 & 0.6714 \\
  GRU, z-score, 4\,h, multi-task & 0.7703 & 0.7701 & 0.7715 & 0.7724 & 0.7172 & 0.7123 & 0.6358 & 0.6270 \\
  GRU, z-score, 1\,h, per task & 0.7671 & 0.7599 & 0.7566 & 0.7504 & 0.6818 & 0.6787 & 0.5904 & 0.5877 \\
  GRU, z-score, 1\,h, multi-task & 0.7641 & 0.7666 & 0.7568 & 0.7607 & 0.6195 & 0.6087 & 0.5509 & 0.5477 \\
  GRU, value bins, 4\,h, per task & 0.7699 & 0.7677 & 0.7715 & 0.7666 & 0.7412 & 0.7387 & 0.6580 & 0.6543 \\
  GRU, value bins, 4\,h, multi-task & 0.7669 & 0.7661 & 0.7668 & 0.7649 & 0.7229 & 0.7169 & 0.6251 & 0.6150 \\
  GRU, value bins, 1\,h, per task & 0.7609 & 0.7533 & 0.7512 & 0.7450 & 0.6849 & 0.6837 & 0.6018 & 0.5952 \\
  GRU, value bins, 1\,h, multi-task & 0.7616 & 0.7637 & 0.7521 & 0.7518 & 0.6599 & 0.6540 & 0.6037 & 0.6000 \\
  Transformer, presence, 4\,h, per task & 0.7146 & 0.7095 & 0.7151 & 0.7124 & 0.7300 & 0.7321 & 0.7472 & 0.7447 \\
  Transformer, presence, 4\,h, multi-task & 0.7187 & 0.7136 & 0.7195 & 0.7144 & 0.7388 & 0.7357 & 0.7323 & 0.7294 \\
  Transformer, presence, 1\,h, per task & 0.7184 & 0.7086 & 0.7211 & 0.7154 & 0.7227 & 0.7203 & 0.7242 & 0.7201 \\
  Transformer, presence, 1\,h, multi-task & 0.7188 & 0.7151 & 0.7183 & 0.7172 & 0.7295 & 0.7231 & 0.7016 & 0.6960 \\
  Transformer, z-score, 4\,h, per task & 0.7703 & 0.7685 & 0.7703 & 0.7671 & \textbf{0.7804} & \textbf{0.7823} & 0.7706 & 0.7699 \\
  Transformer, z-score, 4\,h, multi-task & 0.7715 & 0.7683 & 0.7713 & 0.7727 & 0.7721 & 0.7712 & 0.7596 & 0.7578 \\
  Transformer, z-score, 1\,h, per task & 0.7658 & 0.7618 & 0.7655 & 0.7628 & 0.7505 & 0.7534 & 0.7374 & 0.7390 \\
  Transformer, z-score, 1\,h, multi-task & 0.7682 & 0.7696 & 0.7714 & 0.7729 & 0.7593 & 0.7584 & 0.7388 & 0.7377 \\
  Transformer, value bins, 4\,h, per task & 0.7647 & 0.7646 & 0.7670 & 0.7674 & 0.7700 & 0.7750 & 0.7669 & 0.7706 \\
  Transformer, value bins, 4\,h, multi-task & 0.7661 & 0.7680 & 0.7688 & 0.7674 & 0.7749 & 0.7756 & 0.7594 & 0.7644 \\
  Transformer, value bins, 1\,h, per task & 0.7623 & 0.7622 & 0.7610 & 0.7616 & 0.7509 & 0.7498 & 0.7345 & 0.7330 \\
  Transformer, value bins, 1\,h, multi-task & 0.7664 & 0.7726 & 0.7666 & 0.7704 & 0.7537 & 0.7503 & 0.7300 & 0.7312 \\
  \bottomrule
\end{tabular}

\end{table}

\begin{table}[ht]
  \caption{Bedside scores: mean AUROC over the 19 tasks, calibrated on training data, under the headline rule for carrying the last FiO$_2$ setting. Bold: best on validation. The other 30 configurations (raw scores, and a stricter rule that pairs each PaO$_2$ or SpO$_2$ with an FiO$_2$ charted within 5\,min) reach at most 0.604 on validation.}
  \label{tab:base-e0}
  \centering
  \small
  \begin{tabular}{@{}lcccc@{}}
  \toprule
  & \multicolumn{2}{c}{at $t$} & \multicolumn{2}{c}{worst in 24\,h} \\
  \cmidrule(lr){2-3}\cmidrule(l){4-5}
  Score & val & test & val & test \\
  \midrule
  Partial SOFA & 0.6048 & 0.6045 & \textbf{0.6073} & \textbf{0.6011} \\
  \pf{} & 0.5361 & 0.5333 & 0.5390 & 0.5414 \\
  \sfr{} & 0.5709 & 0.5708 & 0.5581 & 0.5574 \\
  Creatinine & 0.5779 & 0.5819 & 0.5790 & 0.5843 \\
  Mean arterial pressure & 0.5756 & 0.5612 & 0.5823 & 0.5819 \\
  \bottomrule
\end{tabular}

\end{table}

\section{Full per-task results, calibration and decision curves}\label{app:full}
\begin{figure}[ht]
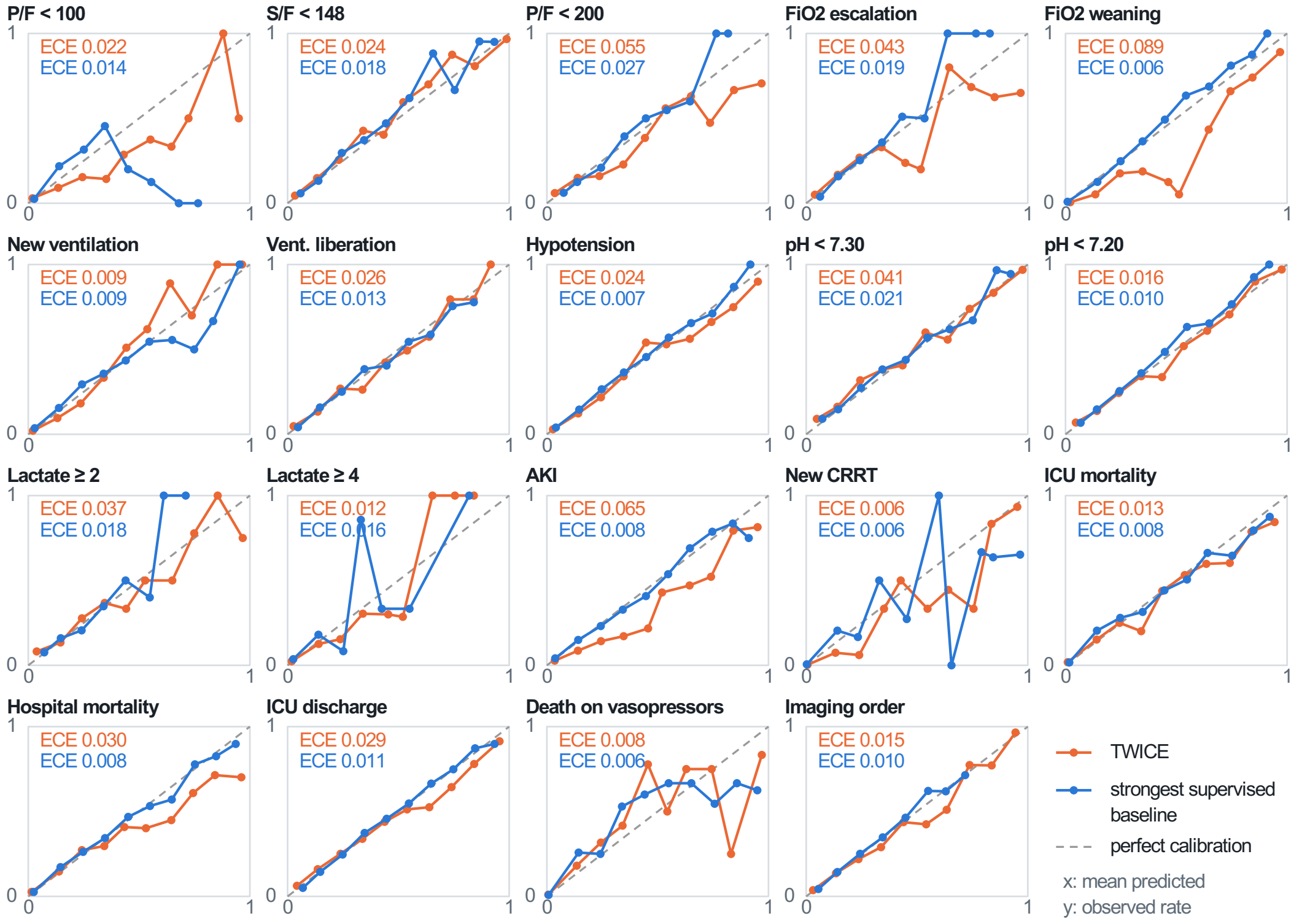

  \centering
  \figpage{7}{\textwidth}
  \caption{Reliability on test, per task. Each point is the observed rate against the mean predicted probability in one of the 10 equal-width bins used for the ECE, for \modelname{} (orange) and the strongest supervised baseline (blue). Every non-empty bin is shown.}
  \label{fig:reliability}
\end{figure}

\begin{figure}[ht]
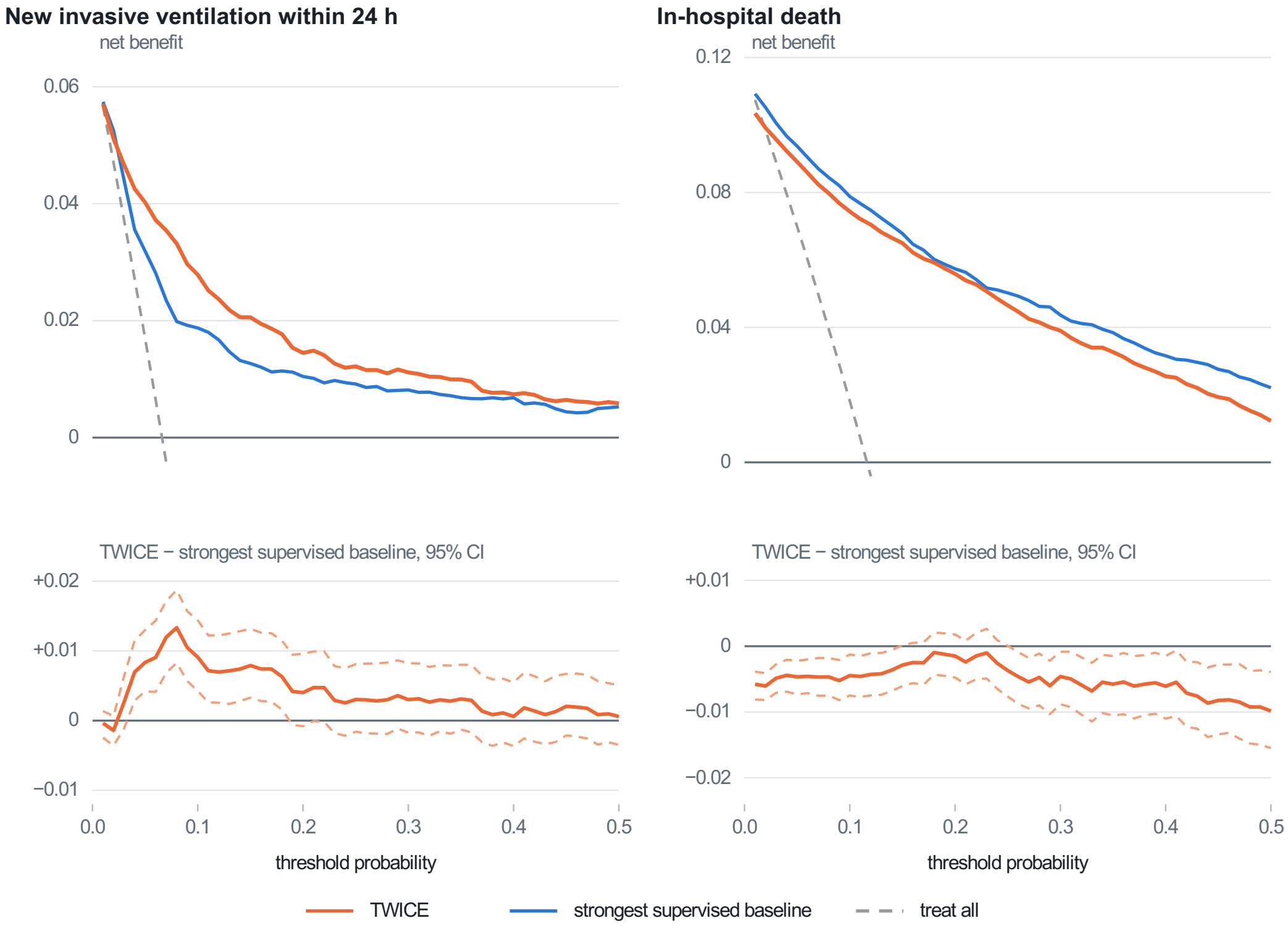

  \centering
  \figpage{8}{\textwidth}
  \caption{Decision curves on every scored validation stay~\citep{vickers2006dca}. The top plots show net benefit against the threshold probability for \modelname{}, the strongest supervised baseline and treating everyone. The bottom plots show \modelname{} minus the strongest supervised baseline, with a patient-clustered 95\% CI (dashed). Thresholds run from 0.01 to 0.50 in steps of 0.01.}
  \label{fig:decision}
\end{figure}

Table~\ref{tab:full-auroc} gives each task's test AUROC for both methods and their paired difference. The $p$-value is two-sided, from the bootstrap distribution of the difference. A tick marks a task that stays significant after Holm's correction across the 19 tasks at level 0.05. The last column is the validation difference. Table~\ref{tab:full-other} gives AUPRC, Brier score and ECE on test. Fig.~\ref{fig:reliability} shows each task's reliability on test. \modelname{}'s probabilities are close to the diagonal for most tasks. They over-predict across the middle of the range for FiO$_2$ weaning and acute kidney injury, the two tasks with the highest ECE, and at high risk for in-hospital death. No recalibration was applied. Fig.~\ref{fig:decision} gives decision curves on every scored validation stay for the two outcomes fixed for the individual trajectories (§\ref{sec:l3}).

\paragraph{Individual trajectories.} The eight stays of §\ref{sec:l3} were drawn at random, before any output was read, from validation stays of at most 14 days still in the ICU at the prediction time. A stay progressed to ventilation if the patient was not ventilated at the prediction time and ventilator charting began in the ICU at least 12\,h later. A stay ended in death if the patient died in hospital at least 12\,h after the prediction time and no later than ICU discharge. A stay had neither event if the patient was never ventilated, survived and stayed at least 12\,h past the prediction time. We drew three, three and two stays. For each stay and outcome we explain the largest rise between consecutive points from the prediction time on, 15 rises in all because one stay had no rise in predicted death. Each rise is picked as the largest of many noisy estimates, so its end is sampled again with a new set of 50 futures. The futures with events left out reuse the random numbers of this set, which keeps the noise in each difference small. In the first stay that progressed to ventilation, the new set gives a risk of 0.94 and leaving out every event of the rise gives 0.38. Of the 315 codes left out alone across these rises, seven have a 95\% CI that excludes zero. Two are the operative note and the procedure of one ventilation rise ($+0.32$ and $+0.20$). Four come from one rise in predicted death whose re-sampled risk at its end was zero. The seventh is a temperature-site entry ($+0.20$). At the prediction time, on the scored validation stays, \modelname{}'s mean predicted risk of new ventilation is 0.40 (95\% CI 0.32 to 0.48) for the 65 stays ventilated within 6\,h, 0.26 (0.20 to 0.33) for 6 to 12\,h and 0.19 (0.16 to 0.22) for 12 to 24\,h, against 0.050 for stays not ventilated within 24\,h (strongest supervised baseline 0.30, 0.13, 0.15 and 0.044). Its mean predicted risk of in-hospital death is 0.73 (0.69 to 0.77) for deaths within 24\,h and 0.37 (0.33 to 0.42) for deaths two weeks or more away, against 0.084 for survivors (strongest supervised baseline 0.63, 0.32 and 0.077). In the decision curves, \modelname{}'s net benefit~\citep{vickers2006dca} is above that of the strongest supervised baseline, with a 95\% CI above zero, at 15 of the 50 thresholds for new ventilation and below it at none. For in-hospital death it is below at 39 thresholds and above at none (Fig.~\ref{fig:decision}).

\paragraph{Spread of the futures.} We measured how often the real value of a series falls inside the central 50\% and 90\% of its simulated values at 6, 12 and 24\,h, for SpO$_2$, FiO$_2$, PEEP, mean arterial pressure, lactate, pH and creatinine. The value at each horizon is the last one charted between the prediction time and that horizon. Values are deciles, so coverage uses the randomised probability integral transform. Across the seven series and three windows, the central 90\% contains the real value in 83\% to 90\% of cases, and its 95\% CI includes 90\% in 16 of the 21. In the other five the futures are too narrow, for FiO$_2$, mean arterial pressure and pH at 6\,h, pH at 12\,h and lactate at 24\,h. The central 50\% contains the real value in 44\% to 54\% of cases, and its CI includes 50\% in 19 of the 21.

\paragraph{Value edits.} Setting a patient's last lactate, pH or creatinine to each decile in turn measures the weight \modelname{} puts on that value~\citep{stutz2026cat}. We expected predicted death to rise with lactate~\citep{casserly2015lactate} and creatinine~\citep{hoste2015epi} and to fall with pH. Between the top and bottom deciles, predicted death changes by $+0.036$ (95\% CI $+0.023$ to $+0.052$) for lactate, $+0.003$ ($-0.013$ to $+0.017$) for pH and $-0.006$ ($-0.015$ to $+0.004$) for creatinine (276, 224 and 286 patients). Only the lactate contrast has the expected sign with a CI that excludes zero. Each edit changes one value and leaves its correlates as charted, so it measures the model's reliance on an observation, not physiology.

\begin{table}[ht]
  \caption{Per-task AUROC on test, \modelname{} against the strongest supervised baseline, paired on identical rows.}
  \label{tab:full-auroc}
  \centering
  \scriptsize
  \setlength{\tabcolsep}{4pt}
  \begin{tabular}{@{}lrccccc c@{}}
  \toprule
  & & \multicolumn{5}{c}{Test} & Val. \\
  \cmidrule(lr){3-7}\cmidrule(l){8-8}
  Task & $n$ (pos.) & \modelname{} & Baseline & $\Delta$ [95\% CI] & $p$ & Holm & $\Delta$ \\
  \midrule
  Severe hypoxaemia & 1,037 (49) & 0.763 & 0.833 & $-0.070$ [$-0.145$, $+0.008$] & 0.070 &  & $-0.028$ \\
  Severe hypoxaemia, \sfr{} & 1,555 (404) & 0.885 & 0.878 & $+0.007$ [$-0.011$, $+0.023$] & 0.370 &  & $+0.015$ \\
  Moderate hypoxaemia & 631 (138) & 0.791 & 0.798 & $-0.006$ [$-0.039$, $+0.027$] & 0.746 &  & $-0.027$ \\
  FiO$_2$ escalation & 2,892 (456) & 0.738 & 0.749 & $-0.011$ [$-0.035$, $+0.013$] & 0.398 &  & $-0.006$ \\
  FiO$_2$ weaning & 2,892 (244) & 0.925 & 0.957 & $-0.032$ [$-0.047$, $-0.019$] & $<$0.001 & $\checkmark$ & $-0.024$ \\
  New invasive ventilation & 3,549 (213) & 0.861 & 0.826 & $+0.034$ [$+0.007$, $+0.061$] & 0.002 & $\checkmark$ & $+0.038$ \\
  Ventilator liberation & 3,798 (677) & 0.819 & 0.803 & $+0.016$ [$+0.003$, $+0.030$] & 0.030 &  & $+0.028$ \\
  Hypotension & 7,250 (1,700) & 0.872 & 0.852 & $+0.020$ [$+0.013$, $+0.027$] & $<$0.001 & $\checkmark$ & $+0.022$ \\
  Acidaemia & 1,707 (475) & 0.781 & 0.776 & $+0.005$ [$-0.019$, $+0.033$] & 0.708 &  & $-0.010$ \\
  Severe acidaemia & 2,260 (478) & 0.780 & 0.793 & $-0.013$ [$-0.040$, $+0.013$] & 0.330 &  & $-0.017$ \\
  Hyperlactataemia & 883 (155) & 0.722 & 0.686 & $+0.036$ [$-0.021$, $+0.090$] & 0.178 &  & $+0.025$ \\
  Severe hyperlactataemia & 1,548 (74) & 0.783 & 0.732 & $+0.051$ [$-0.009$, $+0.106$] & 0.118 &  & $+0.029$ \\
  Acute kidney injury & 8,344 (1,067) & 0.826 & 0.838 & $-0.012$ [$-0.021$, $-0.002$] & 0.018 &  & $-0.014$ \\
  New CRRT & 8,587 (94) & 0.953 & 0.964 & $-0.012$ [$-0.034$, $+0.008$] & 0.294 &  & $-0.008$ \\
  ICU mortality & 8,994 (611) & 0.900 & 0.924 & $-0.024$ [$-0.035$, $-0.013$] & $<$0.001 & $\checkmark$ & $-0.021$ \\
  In-hospital mortality & 8,993 (1,092) & 0.885 & 0.900 & $-0.015$ [$-0.023$, $-0.006$] & $<$0.001 & $\checkmark$ & $-0.034$ \\
  ICU discharge & 8,995 (2,902) & 0.826 & 0.793 & $+0.033$ [$+0.026$, $+0.041$] & $<$0.001 & $\checkmark$ & $+0.036$ \\
  Death on vasopressors & 3,440 (97) & 0.947 & 0.959 & $-0.011$ [$-0.035$, $+0.008$] & 0.280 &  & $+0.005$ \\
  Imaging order & 5,342 (833) & 0.832 & 0.797 & $+0.036$ [$+0.023$, $+0.049$] & $<$0.001 & $\checkmark$ & $+0.045$ \\
  \midrule
  Mean of 19 & & 0.836 & 0.834 & $+0.002$ [$-0.005$, $+0.009$] & 0.640 & & $+0.003$ \\
  \bottomrule
\end{tabular}

\end{table}

\begin{table}[ht]
  \caption{Per-task AUPRC, Brier score and ECE (10 equal-width bins) on test, \modelname{} against the strongest supervised baseline.}
  \label{tab:full-other}
  \centering
  \scriptsize
  \setlength{\tabcolsep}{4pt}
  \begin{tabular}{@{}lccccccc@{}}
  \toprule
  & & \multicolumn{2}{c}{AUPRC $\uparrow$} & \multicolumn{2}{c}{Brier $\downarrow$} & \multicolumn{2}{c}{ECE $\downarrow$} \\
  \cmidrule(lr){3-4}\cmidrule(lr){5-6}\cmidrule(l){7-8}
  Task & Prevalence & \modelname{} & Baseline & \modelname{} & Baseline & \modelname{} & Baseline \\
  \midrule
  Severe hypoxaemia & 0.047 & 0.236 & 0.209 & 0.042 & 0.042 & 0.022 & 0.014 \\
  Severe hypoxaemia, \sfr{} & 0.260 & 0.792 & 0.792 & 0.105 & 0.104 & 0.024 & 0.018 \\
  Moderate hypoxaemia & 0.219 & 0.523 & 0.579 & 0.139 & 0.132 & 0.055 & 0.027 \\
  FiO$_2$ escalation & 0.158 & 0.339 & 0.377 & 0.124 & 0.117 & 0.043 & 0.019 \\
  FiO$_2$ weaning & 0.084 & 0.673 & 0.687 & 0.068 & 0.042 & 0.089 & 0.006 \\
  New invasive ventilation & 0.060 & 0.483 & 0.329 & 0.041 & 0.048 & 0.009 & 0.009 \\
  Ventilator liberation & 0.178 & 0.529 & 0.474 & 0.113 & 0.119 & 0.026 & 0.013 \\
  Hypotension & 0.234 & 0.661 & 0.639 & 0.119 & 0.124 & 0.024 & 0.007 \\
  Acidaemia & 0.278 & 0.638 & 0.619 & 0.154 & 0.157 & 0.041 & 0.021 \\
  Severe acidaemia & 0.212 & 0.585 & 0.591 & 0.127 & 0.126 & 0.016 & 0.010 \\
  Hyperlactataemia & 0.176 & 0.416 & 0.351 & 0.128 & 0.133 & 0.037 & 0.018 \\
  Severe hyperlactataemia & 0.048 & 0.328 & 0.214 & 0.039 & 0.042 & 0.012 & 0.016 \\
  Acute kidney injury & 0.128 & 0.446 & 0.517 & 0.098 & 0.083 & 0.065 & 0.008 \\
  New CRRT & 0.011 & 0.484 & 0.382 & 0.008 & 0.009 & 0.006 & 0.006 \\
  ICU mortality & 0.068 & 0.499 & 0.506 & 0.045 & 0.045 & 0.013 & 0.008 \\
  In-hospital mortality & 0.121 & 0.526 & 0.585 & 0.077 & 0.072 & 0.030 & 0.008 \\
  ICU discharge & 0.323 & 0.714 & 0.655 & 0.153 & 0.167 & 0.029 & 0.011 \\
  Death on vasopressors & 0.028 & 0.509 & 0.473 & 0.019 & 0.019 & 0.008 & 0.006 \\
  Imaging order & 0.156 & 0.549 & 0.436 & 0.098 & 0.108 & 0.015 & 0.010 \\
  \midrule
  Mean of 19 & & 0.523 & 0.495 & 0.089 & 0.089 & 0.030 & 0.012 \\
  \bottomrule
\end{tabular}

\end{table}

\section{Ablations}\label{app:ablations}
Table~\ref{tab:ablations} lists the comparisons on validation behind §\ref{sec:levers}. Except where the table notes otherwise, each changes one setting against a matched control and is scored on the same stays, and $\Delta$ is the difference in mean AUROC over the 19 tasks. Training changes were screened on the 26.2M-parameter model, trained for 150,000 steps and read with 20 futures of 4,096 tokens, against a control trained the same way. Changes to \modelname{} itself were read at 50 futures of 8,192 tokens.
\begin{itemize}
  \item \emph{Demographic prefix.} The static block (sex, race, ethnicity, age band, admission year) is placed before every training window and kept in the prompt, so that a window starting mid-stay still sees it. The loss is not taken on the prefix.
  \item \emph{Windows drawn per stay.} Every encounter is equally likely to supply a training window, instead of in proportion to its length. The tie hides a trade. In-hospital and ICU mortality rose ($+0.042$ and $+0.049$ on the 20-task read, CIs clear of zero), while ventilator liberation, new CRRT and acute kidney injury fell. The model also overfit. At 150,000 steps, its held-out loss exceeded its training loss by 0.037 nats per token, against 0.002 for the control.
  \item \emph{Value bins.} Tail bins add edges at the 0.1st, 1st, 99th and 99.9th percentiles to the deciles. Log-spaced bins keep the 10th, 50th and 90th percentiles and the same tails. Thirty quantile bins triple the resolution, within the range that other work on event models has compared~\citep{lee2026values}. Each changes the vocabulary, so held-out losses do not compare across them.
  \item \emph{Value-regression head.} A linear head at each code token, used in training only, predicts the robust z-score of the value that follows, $(v-\mathrm{median})/(\mathrm{IQR}/1.35)$ clipped to $\pm5$, with weight 1 beside the token loss. Tasks are still read from the bins. Its pre-specified read, the 11 threshold tasks, also tied: $+0.0075$ ($-0.0049$ to $+0.0203$) at 26.2M and $-0.0048$ ($-0.0147$ to $+0.0057$) at 79.3M.
  \item \emph{Same-minute shuffling.} Events stamped with the same minute and of the same kind, whose order is otherwise alphabetical, are permuted in every training window. The mean tied, but the model then wrote ventilator settings far too often. The mean probability of new ventilation was 0.19 against a rate of 0.06.
  \item \emph{Auxiliary future heads.} A head used in training only reads the state after each instant and predicts which codes occur in the next 24\,h (binary cross-entropy), or the time bin in which each first occurs (a discrete-time hazard), with weight 0.3 beside the token loss. The first head learned its target (1.25 nats per event against 2.19 for per-code rates), yet its model's futures stopped for the same reasons, in the same shares, as the control's.
  \item \emph{Regularisation.} Weight decay 0.1 instead of 0.01 and dropout 0.2 instead of 0.1 on the 79.3M model at context 2,048, trained for the full 500,000 steps, since repeated data calls for more dropout~\citep{xue2023repeat} and weight decay~\citep{kim2025wd}. Its held-out loss matched \modelname{}'s (0.630).
  \item \emph{Context by warm start.} The converged 79.3M model at context 1,024 was trained for up to 100,000 more steps at 2,048. It stopped early at 29,000 steps. Its best checkpoint, used here, came at 5,000. It is compared with the same starting checkpoint at 1,024, both read at 20 futures of 4,096 tokens.
\end{itemize}

\begin{table}[ht]
  \caption{Paired comparisons on validation. $\Delta$: mean AUROC over the 19 tasks, $b - a$, on identical rows, with 95\% CIs from 1,000 patient-clustered bootstrap replicates. Readout: futures $\times$ token budget. $^\dagger$CI over 20 tasks, ARDS onset included, because these predictions were deleted before the 19-task re-read. $^\ddagger$The sampler also changed, from the exact sliding window to the ring buffer (ring-buffer row). $^\S$Context 1,024 $\rightarrow$ 2,048, weight decay 0.01 $\rightarrow$ 0.1, dropout 0.1 $\rightarrow$ 0.2 and early-stopping patience 10 $\rightarrow$ 20 change with size.}
  \label{tab:ablations}
  \centering
  \scriptsize
  \setlength{\tabcolsep}{4pt}
  \begin{tabular}{@{}>{\raggedright\arraybackslash}p{4.3cm}>{\raggedright\arraybackslash}p{3.2cm}lcc@{}}
  \toprule
  Change ($a \rightarrow b$) & Model and training & Readout & $\Delta$ AUROC & 95\% CI \\
  \midrule
  \multicolumn{5}{@{}l}{\emph{Readout (same checkpoint)}} \\
  \quad Futures 20 $\rightarrow$ 40$^\ddagger$ & 79.3M, context 1,024 & 4,096 tokens & $+0.0219$ & [$+0.0148$, $+0.0314$]$^\dagger$ \\
  \quad Futures 40 $\rightarrow$ 50 (unpaired) & 79.3M, context 1,024 & 4,096 tokens & $+0.0028$ & -- \\
  \quad Budget 4,096 $\rightarrow$ 8,192 tokens & 79.3M, context 1,024 & 40 futures & $+0.0067$ & [$+0.0002$, $+0.0146$]$^\dagger$ \\
  \quad Budget 4,096 $\rightarrow$ 8,192 tokens & \modelname{} & 50 futures & $+0.0086$ & [$+0.0020$, $+0.0146$] \\
  \quad Ring-buffer cache vs.\ exact sliding window & 26.2M, context 1,024 & 20 $\times$ 4,096 & $+0.0013$ & [$-0.0034$, $+0.0128$]$^\dagger$ \\
  \multicolumn{5}{@{}l}{\emph{Context length}} \\
  \quad Context 1,024 $\rightarrow$ 2,048 & 26.2M, 150k steps & 20 $\times$ 4,096 & $+0.0116$ & [$+0.0028$, $+0.0202$] \\
  \quad Context 1,024 $\rightarrow$ 2,048 & 79.3M & 50 $\times$ 4,096 & $+0.0088$ & [$+0.0010$, $+0.0165$] \\
  \quad Context 1,024 $\rightarrow$ 2,048 by warm start & 79.3M, +5k steps at 2,048 & 20 $\times$ 4,096 & $+0.0007$ & [$-0.0078$, $+0.0092$] \\
  \multicolumn{5}{@{}l}{\emph{Model size}} \\
  \quad 11.7M $\rightarrow$ 26.2M parameters & context 1,024 & 50 $\times$ 8,192 & $+0.0145$ & [$+0.0076$, $+0.0217$] \\
  \quad 26.2M $\rightarrow$ 79.3M parameters$^\S$ & context and regularisation step too & 50 $\times$ 8,192 & $+0.0136$ & [$+0.0069$, $+0.0206$] \\
  \multicolumn{5}{@{}l}{\emph{Changes that did not help}} \\
  \quad Regularisation (weight decay 0.1, dropout 0.2) & 79.3M, context 2,048 & 50 $\times$ 8,192 & $-0.0042$ & [$-0.0111$, $+0.0026$] \\
  \quad Demographic prefix on every window & 26.2M, 150k steps & 20 $\times$ 4,096 & $-0.0004$ & [$-0.0111$, $+0.0066$]$^\dagger$ \\
  \quad Windows drawn per stay, not per token & 26.2M, 150k steps & 20 $\times$ 4,096 & $-0.0016$ & [$-0.0111$, $+0.0077$] \\
  \quad Value bins: deciles plus tail bins & 26.2M, 150k steps & 20 $\times$ 4,096 & $-0.0005$ & [$-0.0092$, $+0.0084$] \\
  \quad Value bins: log-spaced (10th, 50th and 90th percentiles plus tails) & 26.2M, 150k steps & 20 $\times$ 4,096 & $-0.0079$ & [$-0.0166$, $+0.0009$] \\
  \quad Value bins: 30 quantiles & 26.2M, 150k steps & 20 $\times$ 4,096 & $+0.0024$ & [$-0.0060$, $+0.0109$] \\
  \quad Auxiliary value-regression head & 26.2M, 150k steps & 20 $\times$ 4,096 & $+0.0045$ & [$-0.0043$, $+0.0126$] \\
  \quad Auxiliary value-regression head & 79.3M, context 2,048 & 50 $\times$ 8,192 & $-0.0021$ & [$-0.0085$, $+0.0043$] \\
  \quad Same-minute events shuffled in training & 26.2M, 150k steps & 20 $\times$ 4,096 & $-0.0019$ & [$-0.0104$, $+0.0070$] \\
  \quad Auxiliary head: codes in the next 24\,h & 26.2M, 150k steps & 20 $\times$ 4,096 & $+0.0001$ & [$-0.0088$, $+0.0084$] \\
  \quad Auxiliary head: time to each code, 24\,h & 26.2M, 150k steps & 20 $\times$ 4,096 & $+0.0044$ & [$-0.0042$, $+0.0125$] \\
  \bottomrule
\end{tabular}

\end{table}

\section{Discharge diagnoses}\label{app:dx}
Each task asks, at the prediction time, whether the hospital stay will be recorded with a diagnosis in a given CCSR group~\citep{hcup2026ccsr}. Encounter diagnoses sit at the discharge time (§\ref{sec:data}), so each group is read from the futures that reach discharge by the rule of Eq.~\ref{eq:readout}. A \emph{new} variant counts only patients for whom the group is not yet documented at the prediction time, that is, not on the problem list and not recorded in an earlier admission. A group is scored on its own if its expected number of positives (training prevalence times scored rows) is at least 50 (\emph{common}: 116 groups on test). Groups with at least 10 training positives below that are summarised by a macro average (\emph{rare}: 333 groups). The 70 groups with fewer are not scored. The ARDS code J80 is removed from every group, since ARDS onset has its own task.

\paragraph{Reading a group.} A future's diagnosis tokens are read at two depths, the three-character ICD-10 category (\emph{cat.}) or the full hierarchical code (§\ref{sec:tokens}). A token counts for a group only if every vocabulary code sharing it belongs to that group. At full depth, the tokens can express a median 94\% of a common group's diagnosis rows, and on the patients' real futures the reader recovers the label with median recall 0.94. For 264 of the 333 rare groups, no row can be expressed, because their codes were pruned from the vocabulary and roll up to tokens shared with other groups. Such a group falls back to its training rate and scores an AUROC of exactly 0.5.

\paragraph{Comparator.} No supervised model was trained for this family. The \emph{documented} rule predicts each group's training rate given whether the group is already documented at the prediction time. On new diagnoses it gives every patient the same probability, so its AUROC is 0.5.

\paragraph{Results.} Table~\ref{tab:dx} gives macro averages over groups. \modelname{} reaches 0.717 on all diagnoses and 0.693 on new ones. The documented rule leads on all diagnoses by AUROC (0.742) and trails by AUPRC (0.141 vs.\ 0.174). The category-level reader is weaker (0.648 and 0.632), since many groups are defined below the three-character level. Validation is within 0.005 of test on every row.

\paragraph{Threats.} The new arm inherits the common set from all diagnoses, so 47 of its 115 scorable groups have fewer than 50 positives on test (macro AUROC 0.647, against 0.725 over the other 68). A future that stops at the token budget before discharge writes no diagnoses and is undecided. Of the test futures, 28.2\% stop this way. A future that dies before discharge counts as positive for a group it has already written and is undecided otherwise, which biases dying patients' probabilities upward. Encounter diagnoses record what was documented, which need not be what occurred.

\begin{table}[ht]
  \caption{Discharge diagnosis groups: macro averages over groups with both classes ($n$). Pos.: positive rows summed over groups. \modelname{} reads the full code (AUROC, AUPRC) or the three-character category only (cat.). \emph{Documented} is the rule that uses only whether the group is already documented at the prediction time.}
  \label{tab:dx}
  \centering
  \scriptsize
  \setlength{\tabcolsep}{4pt}
  \begin{tabular}{@{}llrrccccc c@{}}
  \toprule
  & & & & \multicolumn{3}{c}{\modelname{}, test} & \multicolumn{2}{c}{Documented, test} & Val. \\
  \cmidrule(lr){5-7}\cmidrule(lr){8-9}\cmidrule(l){10-10}
  Groups & Diagnoses & $n$ & Pos. & AUROC & AUPRC & AUROC (cat.) & AUROC & AUPRC & AUROC \\
  \midrule
  Common & all & 116 & 34,279 & 0.717 & 0.174 & 0.648 & 0.742 & 0.141 & 0.720 \\
  Common & new & 115 & 16,251 & 0.693 & 0.086 & 0.632 & 0.500 & 0.019 & 0.689 \\
  Rare & all & 293 & 4,101 & 0.515 & 0.007 & 0.505 & 0.776 & 0.091 & 0.513 \\
  Rare & new & 233 & 1,762 & 0.514 & 0.004 & 0.504 & 0.500 & 0.001 & 0.516 \\
  \bottomrule
\end{tabular}

\end{table}

\section{ARDS onset}\label{app:ards}
ARDS onset within 48\,h is excluded from the reported suite by the rule in §\ref{sec:data}. Its label rests on our collaborators' NLP-based adjudication of the Berlin criteria~\citep{ranieri2012berlin}, which exists only for encounters that passed a screen for acute hypoxaemic respiratory failure, while every other label is a rule over structured events. The rule was fixed before \modelname{} was scored on test. On test (868 rows, 68 positives), \modelname{} reaches an AUROC of 0.757 (95\% CI 0.687--0.817) and the strongest supervised baseline 0.816 (0.755--0.867). On validation (928 rows, 75 positives) they reach 0.780 and 0.830, a paired difference of $-0.050$ ($-0.094$ to $-0.009$). Including the task would move the mean difference over tasks from $+0.0017$ to $-0.0013$ on test (point estimates) and from $+0.0028$ to $+0.0002$ on validation. The two methods tie either way.

\section{What-if scenarios and trial emulation}\label{app:cf}
\paragraph{Design.} Each scenario of §\ref{sec:eval} is defined by its decision points, arms and outcomes (Table~\ref{tab:cf}).
\begin{itemize}
  \item \emph{Decision points.} A trigger is a charted event after which clinicians commonly choose a treatment. A decision point is the first trigger of a scenario in a validation ICU stay, and the history is cut there. Each stay therefore contributes at most one decision point to a scenario. In the two trial scenarios, the trigger is the moment a patient first meets the trial's enrolment criteria.
  \item \emph{Arms.} Each scenario compares treatment arms with a control arm (§\ref{sec:intervene}). For each arm at each decision point, \modelname{} samples 50 futures from the same history.
  \item \emph{Outcomes and effects.} An outcome is an event within a fixed window after the decision point. Its value in an arm is the fraction of the arm's futures in which the event occurs. An outcome on a laboratory value counts only the futures that measure the value in the window, as in §\ref{sec:readout}. The effect of an order on an outcome is its value in the order's arm minus its value in the control arm, averaged over decision points.
  \item \emph{Kinds of effect.} In each of the five scenarios that do not emulate a trial, a \emph{right treatment} is one that guidelines recommend for the trigger, and an \emph{opposite treatment} is expected to move the triggering value the other way. A \emph{target effect} is the effect of either on that value. For a right treatment, it is the intended effect~\citep{miettinen1983}. A \emph{side effect} is a known adverse effect of a drug on another outcome. An \emph{inert drug} is expected to change no outcome. Pantoprazole is the inert drug in every scenario.
  \item \emph{Confounding by indication.} A treatment is ordered because of its target, so target effects are the most open to confounding by indication (§\ref{sec:related}). A drug is rarely ordered for its side effects, so they are less open to it~\citep{miettinen1983}. Side effects therefore serve as positive controls, and the inert drug serves as a negative-control exposure~\citep{lipsitch2010negative}.
  \item \emph{Pass rule.} An effect passes when its 95\% CI excludes zero in the expected direction. It is reversed when its CI excludes zero in the other direction, and inconclusive otherwise. An expected-zero effect whose CI excludes zero counts as a shift. We pre-specified the outcomes, their windows, the expected directions and this rule.
\end{itemize}

\paragraph{Worked example.} The high-potassium scenario illustrates the design. Its trigger is a potassium of at least 6.0\,mmol/L in a patient who has received no potassium-lowering drug in the previous 24\,h. Validation contains 221 such decision points. The control arm forbids all of the scenario's drugs. The drug arms order insulin, salbutamol, IV calcium, potassium chloride and pantoprazole (Table~\ref{tab:cf}). Each order is written into the EHR 63\,min after the trigger, the median delay from trigger to order in the training data. One outcome is a first potassium below 5.5\,mmol/L within 4\,h. Its value in an arm is the fraction of futures with that event, among the futures that measure potassium in that window. The cutoff of 5.5\,mmol/L lies inside the top potassium decile, so this outcome is bracketed (§\ref{sec:readout}). Insulin lowers potassium~\citep{ukka2020hyperk}, so its target effect passes if this fraction is higher in the insulin arm than in the control arm, with a CI above zero. Potassium chloride is the opposite treatment, and its effect is expected to be negative. In the training data, potassium chloride ordered at a high potassium plausibly follows a sample judged haemolysed. A model that reads the order as a sign of the patient's state would then predict a lower potassium after potassium chloride, as after insulin. IV calcium protects the heart but does not lower potassium~\citep{ukka2020hyperk}, so it is a second inert drug for this outcome. Another outcome, any glucose below 70\,mg/dL within 6\,h, tests a side effect of insulin.

\begin{table}[ht]
  \caption{Clinical scenarios on validation, with the number of decision points of each. In the low-blood-pressure and start-of-ventilation scenarios, 1,000 decision points were drawn at random from 4,232 and 2,055 to limit compute. The control arm is listed first (§\ref{sec:intervene}). Pantoprazole, the inert drug, is also an arm of every scenario.}
  \label{tab:cf}
  \centering
  \scriptsize
  \begin{tabularx}{\textwidth}{@{}>{\raggedright\arraybackslash}p{2.1cm}>{\raggedright\arraybackslash}X>{\raggedright\arraybackslash}X>{\raggedright\arraybackslash}p{3.0cm}@{}}
    \toprule
    Scenario (points) & Decision point & Arms & Outcomes \\
    \midrule
    High potassium (221) & K$^+\geq$6.0\,mmol/L, no potassium-lowering drug in 24\,h & none; insulin; salbutamol; IV calcium (no effect on K$^+$ expected); KCl (opposite) & first K$^+<$6.0 in 2\,h; first K$^+<$5.5 in 4\,h; any glucose $<$70\,mg/dL in 6\,h \\
    Low glucose (945) & glucose $<$70\,mg/dL & none; dextrose; insulin (opposite) & first glucose $\geq$70 in 1\,h \\
    Low blood pressure (1,000) & mean arterial pressure $<$65\,mmHg, no vasopressor in 24\,h & none; norepinephrine; crystalloid; nicardipine (opposite) & first MAP $\geq$65 in 1\,h \\
    Low SpO$_2$ on the ventilator (198) & SpO$_2<$88\%, ventilated within 4\,h, FiO$_2$ and PEEP charted within 12\,h & usual care; FiO$_2$ $+$0.1; PEEP one ARDS Network step up; FiO$_2$ $-$0.1 (opposite) & first SpO$_2\geq$88\% in 1\,h \\
    Start of ventilation (1,000) & first ventilator marker, no sedative in 24\,h & none; propofol; dexmedetomidine; midazolam & any MAP $<$65 in 2\,h; any heart rate $<$50 in 6\,h \\
    ROSE~\citep{moss2019rose} (36) & \pf{} $<$150 on PEEP $\geq$8 within 48\,h of ARDS onset, ventilated, no neuromuscular blocker yet & none; cisatracurium & in-hospital death \\
    RECOVERY~\citep{recovery2021dexa}, ventilated (32) & COVID-19 diagnosis at the first ventilator marker, no glucocorticoid yet & none; dexamethasone & in-hospital death; any glucose $>$180 in 24\,h \\
    \bottomrule
  \end{tabularx}
\end{table}

\paragraph{Scenarios.} Thresholds follow guidelines: potassium from the UK Kidney Association~\citep{ukka2020hyperk}, glucose from the American Diabetes Association's level-1 hypoglycaemia~\citep{ada2024hypo}, blood pressure from the Surviving Sepsis Campaign~\citep{evans2021ssc}, SpO$_2$ from the ARDS Network's lower goal~\citep{ardsnet2000arma} and bradycardia from the 2018 ACC/AHA/HRS guideline~\citep{kusumoto2019brady}. The sedative arms follow the PADIS guideline~\citep{devlin2018padis}. The expected effects on the targets follow from the treatments these guidelines recommend. Insulin and salbutamol, for example, lower potassium by shifting it into cells~\citep{ukka2020hyperk}. The opposite arms are expected to move the target the other way. Potassium chloride adds potassium, insulin lowers glucose, nicardipine dilates blood vessels and a lower FiO$_2$ delivers less oxygen. The expected side effects are adverse effects of the drugs. These are hypoglycaemia after insulin~\citep{ukka2020hyperk}, hyperglycaemia after glucocorticoids~\citep{clore2009glucocorticoid}, hypotension after propofol and bradycardia after dexmedetomidine~\citep{hughes2012sedation}. The trial scenarios emulate the eligibility of ROSE~\citep{moss2019rose} and of RECOVERY's invasively ventilated stratum~\citep{recovery2021dexa}. ROSE's onset time comes from our collaborators' adjudication of the Berlin criteria (App.~\ref{app:ards}).

\paragraph{Sampling.} An order is inserted at the median delay from trigger to order among treated training points: 63\,min at a high potassium, 25 at a low glucose, 33 at a low blood pressure and 50 at the start of ventilation, at once for ventilator settings, and 156 and 465\,min for ROSE and RECOVERY. The order reached 96.4\% to 99.9\% of futures across scenarios. Futures stop at the end of the outcome window, after 2,048 tokens, or, in the trial scenarios, at discharge, death, 30 days or 32,768 tokens. A usual-care arm without insertion or ban is always run. Its futures give the orders after an abnormal value (App.~\ref{app:fidelity}).

\paragraph{Outcomes.} An outcome whose clinical cutoff falls inside a value decile is bracketed as in §\ref{sec:readout}. Such outcomes resolve the direction of an effect better than its size. The top potassium decile starts between 4.8 and 5.7\,mmol/L, depending on the laboratory code. ``Below 6.0\,mmol/L'' is therefore bracketed between ``below the top decile'' and ``any value''. The bottom glucose decile ends between 89 and 128\,mg/dL, depending on the code. At a high potassium, the potassium outcome pre-specified as primary was the first K$^+$ below 6.0\,mmol/L within 2\,h. §\ref{sec:l4} and Fig.~\ref{fig:interventions} report the second, below 5.5\,mmol/L within 4\,h, which lies in the same top decile and differs only in its window. On the primary outcome the effects point the same way and are smaller, $+0.011$ ($+0.0004$ to $+0.022$) for insulin and $+0.007$ ($-0.004$ to $+0.019$) for potassium chloride.

\paragraph{Trials.} Trial effects are compared on the difference scale for ROSE and on the log-ratio scale for RECOVERY, with RCT-DUPLICATE's agreement metrics~\citep{wang2023rctduplicate}. The endpoint is in-hospital death, not the trials' 90- or 28-day mortality, and eligibility is emulated from flowsheet and laboratory data in a different era and population. The control arms forbid the trial drug for the whole future, while 86 of ROSE's 505 usual-care patients (17.0\%) received a neuromuscular blocker in the first 48\,h~\citep{moss2019rose}. The longest budget is primary. It was fixed before any long run because it leaves the fewest futures undecided. At shorter budgets, read off the same futures, cisatracurium in ROSE raises death by 0.143 (0.004 to 0.289) at 4,096 tokens and by 0.082 (0.015 to 0.159) at 8,192, an excess of early deaths that is gone at the primary budget. \modelname{}'s absolute risk of death in the 26 untreated ROSE patients is 0.535, against an observed 0.308 (RECOVERY: 0.356 against 0.333 in 24). If most futures left undecided by the budget are survivors, as long stays tend to be, death among the decided futures is inflated. The trial contrasts assume both arms share this bias.

\paragraph{Observed outcomes and tabular estimators.} A decision point counts as treated when the treatment was ordered within 1 to 2\,h of it, or within 12 to 24\,h in the trial scenarios. In the four scenarios with an opposite treatment, the futures with the received treatment written in never predicted the observed outcome better than the futures with the opposite treatment. The tabular estimators use code counts and the deciles of the last values in the previous 24\,h. They are fitted on training decision points by scikit-learn's histogram-based gradient boosting at default settings. One is an S-learner and the other is inverse-probability weighting. Inverse-probability weighting reverses insulin on the 2\,h potassium outcome ($-0.415$, $-0.678$ to $-0.164$), and its CIs include zero for every other effect on a target. We also planned to calibrate the predictions under each treatment with inverse-probability weights. This could not be estimated reliably, since only 2 to 111 decision points per treatment fell where both choices were plausible.

\paragraph{Not reported.} PEEP changes read as natural experiments could not be scored. The inserted value token equalled the previous one at 55 of 211 points for the actual change and at 123 for its mirror image, and 57\% of futures stopped at the budget within 6\,h.

\section{Next-event and fidelity details}\label{app:fidelity}
\paragraph{Next token.} Positions are up to 64 random points in each of 8,938 validation encounters (6,920 patients; 571,002 positions), read with the true history as context, up to 2,048 tokens. Table~\ref{tab:ntp} gives top-$k$ accuracy by the type of the true next token, with patient-clustered CIs. The unigram and bigram models are counted on the training stays. Set recall SR@($i$,\,$m$)~\citep[Eq.~9]{pickard2026ehrmpc} is the share of the distinct tokens among the next $m$ that appear among the model's $i$ most likely next tokens (Table~\ref{tab:sr}). The event-code version counts codes only and skips values and time gaps. For long horizons and long lists, a frequency list recalls more than the next-token distribution, which concentrates on the immediate next token, so set recall is a weak test of a next-token model beyond short horizons.

\begin{table}[ht]
  \caption{Next-token accuracy (\%) on validation by the type of the true next token, with 95\% CIs.}
  \label{tab:ntp}
  \centering
  \scriptsize
  \setlength{\tabcolsep}{4pt}
  \begin{tabular}{@{}lrrccccc@{}}
  \toprule
  & & & \multicolumn{3}{c}{\modelname{}} & Unigram & Bigram \\
  \cmidrule(lr){4-6}\cmidrule(lr){7-7}\cmidrule(l){8-8}
  True next token & Positions & \% & Top-1 & Top-5 & Top-10 & Top-1 & Top-1 \\
  \midrule
  All tokens & 571,002 & 100.0 & 75.0 [74.7, 75.2] & 95.5 [95.4, 95.6] & 98.2 [98.2, 98.3] & 5.1 & 20.1 \\
  Value bin & 204,930 & 35.9 & 64.3 [63.9, 64.7] & 95.9 [95.8, 96.0] & 100.0 [100.0, 100.0] & 14.2 & 17.8 \\
  Bedside or ventilator code & 199,067 & 34.9 & 92.6 [92.5, 92.8] & 98.9 [98.9, 99.0] & 99.5 [99.5, 99.6] & 0.0 & 22.4 \\
  Time gap & 51,082 & 8.9 & 54.6 [54.1, 55.2] & 97.8 [97.6, 97.9] & 99.8 [99.7, 99.8] & 0.0 & 24.3 \\
  Laboratory test & 48,325 & 8.5 & 90.6 [90.3, 90.8] & 97.4 [97.2, 97.5] & 98.5 [98.3, 98.6] & 0.0 & 9.0 \\
  Procedure order & 29,344 & 5.1 & 54.9 [54.3, 55.4] & 76.6 [76.1, 77.1] & 83.7 [83.2, 84.1] & 0.0 & 23.0 \\
  Other & 22,732 & 4.0 & 88.9 [88.5, 89.4] & 99.6 [99.5, 99.7] & 100.0 [99.9, 100.0] & 0.0 & 44.1 \\
  Medication order & 8,141 & 1.4 & 19.8 [18.9, 20.7] & 50.5 [49.4, 51.6] & 68.3 [67.3, 69.3] & 0.0 & 0.1 \\
  Imaging order or note type & 5,662 & 1.0 & 24.6 [23.4, 25.7] & 71.3 [70.0, 72.4] & 84.9 [83.9, 85.8] & 0.0 & 1.9 \\
  Admission, ICU or discharge & 935 & 0.2 & 68.4 [65.6, 71.3] & 92.3 [90.6, 93.9] & 96.9 [95.7, 97.9] & 0.0 & 13.2 \\
  Diagnosis & 579 & 0.1 & 29.4 [25.6, 33.4] & 68.4 [64.4, 72.2] & 81.3 [78.1, 84.4] & 0.0 & 0.5 \\
  \bottomrule
\end{tabular}

\end{table}

\begin{table}[ht]
  \caption{Set recall SR@($i$,\,$m$) (\%) on validation: of the distinct tokens among the next $m$, the share in the top $i$.}
  \label{tab:sr}
  \centering
  \scriptsize
  \begin{tabular}{@{}lrcccccc@{}}
  \toprule
  & & \multicolumn{3}{c}{$m=5$} & \multicolumn{3}{c}{$m=10$} \\
  \cmidrule(lr){3-5}\cmidrule(l){6-8}
  Unit & $i$ & \modelname{} & Unigram & Bigram & \modelname{} & Unigram & Bigram \\
  \midrule
  Event codes & 10 & 54.2 & 24.1 & 29.6 & 34.1 & 23.8 & 23.0 \\
   & 20 & 64.8 & 43.5 & 46.8 & 45.2 & 43.2 & 38.4 \\
   & 50 & 74.4 & 66.6 & 66.0 & 58.1 & 66.4 & 57.8 \\
  Every token & 10 & 46.4 & 34.8 & 32.2 & 33.9 & 32.5 & 25.6 \\
   & 20 & 51.6 & 51.2 & 46.9 & 39.9 & 49.2 & 40.1 \\
   & 50 & 56.5 & 75.7 & 64.4 & 46.5 & 74.9 & 59.0 \\
  \bottomrule
\end{tabular}

\end{table}

\paragraph{Orders after an abnormal value.} The decision points are those of the high-potassium, low-glucose and low-blood-pressure scenarios (Table~\ref{tab:cf}). An order's predicted probability is Eq.~\eqref{eq:readout} over the usual-care futures, with the event being any order of its drug class in the window after the trigger (Table~\ref{tab:orders}). Classes are matched by ATC prefix. Insulin is A10A, IV calcium B05XA07 and B05XA19, $\beta_2$-agonists R03AC and R03CC, potassium binders V03AE, dextrose or glucagon V06DC, B05BA and H04AA, and vasopressors C01CA and H01BA. The comparator is the gradient-boosted model of App.~\ref{app:cf}, fitted on training decision points to whether the order was placed. The first drug ordered is read the same way. Norepinephrine is 56\% of the first vasopressors within 6\,h in the futures and 64\% of the 260 in the EHR. Propofol is 74\% of the first sedatives within 2\,h of the start of ventilation in the futures and 91\% of the 546 in the EHR.

\begin{table}[ht]
  \caption{Orders after an abnormal value on validation. Share ordered is, for clinicians, the fraction of decision points at which the order was placed and, for futures, the mean predicted probability. AUROCs have patient-clustered 95\% CIs.}
  \label{tab:orders}
  \centering
  \scriptsize
  \setlength{\tabcolsep}{4pt}
  \begin{tabular}{@{}lllrcccc@{}}
  \toprule
  & & & & \multicolumn{2}{c}{Share ordered} & \multicolumn{2}{c}{AUROC} \\
  \cmidrule(lr){5-6}\cmidrule(l){7-8}
  Decision point & Order & Window & Points & Clinicians & Futures & \modelname{} & Gradient boosting \\
  \midrule
  K$^+\geq$6.0\,mmol/L & Insulin & 2\,h & 221 & 0.26 & 0.12 & 0.811 [0.741, 0.871] & 0.785 [0.726, 0.839] \\
   & IV calcium & 2\,h & 221 & 0.24 & 0.12 & 0.839 [0.773, 0.900] & 0.826 [0.761, 0.883] \\
   & $\beta_2$-agonist & 2\,h & 221 & 0.03 & 0.02 & 0.652 [0.488, 0.832] & 0.717 [0.546, 0.833] \\
   & Potassium binder & 2\,h & 221 & 0.07 & 0.03 & 0.742 [0.608, 0.855] & 0.781 [0.692, 0.861] \\
  Glucose $<$70\,mg/dL & Dextrose or glucagon & 1\,h & 945 & 0.14 & 0.07 & 0.714 [0.662, 0.766] & 0.758 [0.709, 0.800] \\
  MAP $<$65\,mmHg & Vasopressor & 1\,h & 1,000 & 0.14 & 0.14 & 0.935 [0.910, 0.954] & 0.900 [0.864, 0.930] \\
  \bottomrule
\end{tabular}

\end{table}

\paragraph{Generated next 24\,h.} For every scored validation stay, one future is cut at 24\,h and its concepts are compared with the real window's (Table~\ref{tab:fid}). A concept is a laboratory test, a medication's ATC therapeutic subgroup, a procedure, a bedside code or a three-character diagnosis category. The all-rows columns give CoMET's RMSLE between real and generated rates over 8,947 rows, beside two references: two disjoint random halves of the real rows (each half the size, so noisier than a full-size comparison), and, for co-occurrence, the real rates of each concept under independence. The swap test uses the earliest row of each patient whose real and first two generated windows fit within the token budget (5,747 of 5,836). The test compares the observed RMSLE with 999 RMSLEs recomputed after swapping each patient's real and generated windows with probability one half. The ratio divides the observed RMSLE by the median of these 999. Concepts are tested one at a time with McNemar's exact test and the Benjamini--Hochberg procedure at 5\%. The personal-overlap control and its ceiling (§\ref{sec:l2}) use the same 5,747 patients and two generated futures each.

\begin{table}[ht]
  \caption{Generated against real next 24\,h on validation, per panel of concepts. Holm $p$: across the 20 panels.}
  \label{tab:fid}
  \centering
  \scriptsize
  \setlength{\tabcolsep}{3pt}
  \begin{tabular}{@{}lrcccccccc@{}}
  \toprule
  & & \multicolumn{3}{c}{All rows: RMSLE} & \multicolumn{4}{c}{Swap test} & Concepts \\
  \cmidrule(lr){3-5}\cmidrule(lr){6-9}\cmidrule(l){10-10}
  Panel & Cells & \modelname{} & Real halves & Independence & RMSLE & Swap median & Ratio & Holm $p$ & differ / tested \\
  \midrule
  Diagnoses & 144 & 0.317 & 0.362 & -- & 0.335 & 0.254 & 1.32 & 0.020 & 0 / 143 \\
  Laboratory & 1,173 & 0.138 & 0.218 & -- & 0.179 & 0.149 & 1.20 & 0.072 & 12 / 1,177 \\
  Medications & 258 & 0.165 & 0.206 & -- & 0.186 & 0.163 & 1.14 & 0.168 & 1 / 255 \\
  Procedures & 1,016 & 0.200 & 0.276 & -- & 0.235 & 0.203 & 1.16 & 0.020 & 3 / 1,023 \\
  Bedside & 100 & 0.074 & 0.106 & -- & 0.051 & 0.054 & 0.94 & 0.620 & 0 / 99 \\
  \midrule
  Diagnoses $\times$ diagnoses & 10,296 & 0.150 & 0.196 & 0.174 & 0.125 & 0.121 & 1.03 & 0.104 & \\
  Diagnoses $\times$ laboratory & 168,912 & 0.129 & 0.161 & 0.128 & 0.116 & 0.108 & 1.08 & 0.020 & \\
  Diagnoses $\times$ medications & 37,152 & 0.148 & 0.191 & 0.137 & 0.134 & 0.128 & 1.05 & 0.020 & \\
  Diagnoses $\times$ procedures & 146,304 & 0.089 & 0.113 & 0.103 & 0.077 & 0.074 & 1.05 & 0.020 & \\
  Diagnoses $\times$ bedside & 14,400 & 0.210 & 0.250 & 0.184 & 0.191 & 0.169 & 1.13 & 0.020 & \\
  Laboratory $\times$ laboratory & 687,378 & 0.180 & 0.232 & 0.338 & 0.171 & 0.164 & 1.04 & 0.140 & \\
  Laboratory $\times$ medications & 302,634 & 0.192 & 0.261 & 0.168 & 0.187 & 0.182 & 1.03 & 0.040 & \\
  Laboratory $\times$ procedures & 1,191,768 & 0.176 & 0.231 & 0.182 & 0.165 & 0.159 & 1.03 & 0.020 & \\
  Laboratory $\times$ bedside & 117,300 & 0.195 & 0.296 & 0.246 & 0.199 & 0.193 & 1.03 & 0.200 & \\
  Medications $\times$ medications & 33,153 & 0.200 & 0.265 & 0.182 & 0.190 & 0.186 & 1.02 & 0.200 & \\
  Medications $\times$ procedures & 262,128 & 0.187 & 0.246 & 0.160 & 0.177 & 0.172 & 1.03 & 0.020 & \\
  Medications $\times$ bedside & 25,800 & 0.205 & 0.278 & 0.232 & 0.198 & 0.191 & 1.04 & 0.168 & \\
  Procedures $\times$ procedures & 515,620 & 0.160 & 0.204 & 0.163 & 0.145 & 0.141 & 1.03 & 0.020 & \\
  Procedures $\times$ bedside & 101,600 & 0.219 & 0.295 & 0.243 & 0.215 & 0.201 & 1.07 & 0.020 & \\
  Bedside $\times$ bedside & 4,950 & 0.161 & 0.250 & 0.454 & 0.163 & 0.156 & 1.04 & 0.598 & \\
  \bottomrule
\end{tabular}

\end{table}

\section{Data-integrity checks}\label{app:integrity}
\paragraph{Leakage by timing.} Encounter diagnoses are placed at the discharge time, so no input before discharge contains them (§\ref{sec:data}). Note-type tokens carry no text but are placed at the note's service time, which for 10.15\% of notes precedes the note's creation. Moving every note token to the time its first version was filed changed the mean validation AUROC of the tabular baseline over the 44 tasks with at least 50 positives from 0.8444 to 0.8434 ($-0.0010$), the size of the difference between two fits. Of all 47 tasks, 24 fell, 21 rose and 2 were unchanged. Every result is therefore reported on the original timing.

\paragraph{Test exposure.} Before 10 October 2026, the development runs of the generative model drew a random sample of 2,000 prediction rows from validation and test together for their simulation diagnostics. The real next 24\,h of the sampled test patients entered the fidelity diagnostics of those runs (no test label did). No number in this paper comes from those draws, and the exposed test patients were not excluded. From that date, simulation drew from validation only.

\paragraph{Record checks.} Conversion to MEDS kept every patient, encounter and ICU stay of the extract and removed 0.79\% of rows as exact duplicates. The name-based rescue of uncoded medications placed medications in ATC groups with a leave-one-out precision of 99.0\%. Across the cohort, ICU admission and discharge events differ in number by 124, 38 encounters have no complete ICU stay, and one has no admission or discharge time.

\section{Performance by subgroup}\label{app:subgroups}
Table~\ref{tab:subgroups} gives \modelname{} minus the strongest supervised baseline on the 19 test tasks within groups of age ($<$65 and $\geq$65 years, the cohort's median), sex, race (White, Black or African American, Asian, other or unknown), ethnicity (Hispanic, not Hispanic), whether the encounter passed the screen for acute hypoxaemic respiratory failure, and era (ICU admission before 2020, and from 2020), as TRIPOD+AI recommends for sociodemographic groups~\citep{collins2024tripodai}. The groups use the values as recorded (Table~\ref{tab:cohort-full}), and the comparisons are descriptive, without correction for multiplicity. A task enters a group's mean when both outcomes occur in the group, so some means rest on few positive stays.

Over all scored stays the difference is $+0.0017$, as in Table~\ref{tab:l1}. Its 95\% CI excludes zero in one of the 14 groups, the stays that passed the screen for acute hypoxaemic respiratory failure ($+0.012$, 95\% CI $+0.004$ to $+0.020$). With 14 groups, about one such interval is expected by chance. In no group does \modelname{} trail the strongest supervised baseline with a CI that excludes zero.

\begin{table}[ht]
  \caption{Performance by subgroup on test. Mean AUROC over the tasks with both outcomes in the group, \modelname{} against the strongest supervised baseline (Baseline), paired on identical rows. $\Delta$: \modelname{} minus the baseline, with 95\% CIs from 1,000 patient-clustered bootstrap replicates. Min.\ pos.: the fewest positive stays of any task in the group. AHRF: acute hypoxaemic respiratory failure. The ethnicity groups leave out 512 stays with ethnicity unavailable, declined or unknown.}
  \label{tab:subgroups}
  \centering
  \scriptsize
  \setlength{\tabcolsep}{5pt}
  \begin{tabular}{@{}lrrrrcccc@{}}
  \toprule
  Group & Stays & Patients & Tasks & Min.\ pos. & \modelname{} & Baseline & $\Delta$ & 95\% CI \\
  \midrule
  All & 8,995 & 5,794 & 19 & 49 & 0.8362 & 0.8345 & $+0.0017$ & [$-0.0053$, $+0.0090$] \\
  \multicolumn{9}{@{}l}{\emph{Age}} \\
  \quad $<$65 years & 4,292 & 2,722 & 19 & 30 & 0.8414 & 0.8397 & $+0.0017$ & [$-0.0081$, $+0.0113$] \\
  \quad $\geq$65 years & 4,703 & 3,112 & 19 & 19 & 0.8304 & 0.8296 & $+0.0008$ & [$-0.0105$, $+0.0125$] \\
  \multicolumn{9}{@{}l}{\emph{Sex}} \\
  \quad Female & 3,903 & 2,514 & 19 & 16 & 0.8415 & 0.8359 & $+0.0056$ & [$-0.0052$, $+0.0163$] \\
  \quad Male & 5,092 & 3,280 & 19 & 33 & 0.8330 & 0.8337 & $-0.0007$ & [$-0.0109$, $+0.0092$] \\
  \multicolumn{9}{@{}l}{\emph{Race}} \\
  \quad White & 7,058 & 4,541 & 19 & 35 & 0.8338 & 0.8330 & $+0.0007$ & [$-0.0080$, $+0.0090$] \\
  \quad Black or African American & 705 & 402 & 19 & 4 & 0.8464 & 0.8380 & $+0.0084$ & [$-0.0190$, $+0.0327$] \\
  \quad Asian & 276 & 182 & 18 & 1 & 0.8486 & 0.8567 & $-0.0081$ & [$-0.0453$, $+0.0260$] \\
  \quad Other or unknown & 956 & 669 & 19 & 8 & 0.8461 & 0.8383 & $+0.0078$ & [$-0.0088$, $+0.0238$] \\
  \multicolumn{9}{@{}l}{\emph{Ethnicity}} \\
  \quad Hispanic & 732 & 458 & 19 & 3 & 0.8496 & 0.8312 & $+0.0184$ & [$-0.0010$, $+0.0378$] \\
  \quad Not Hispanic & 7,751 & 4,970 & 19 & 39 & 0.8335 & 0.8346 & $-0.0011$ & [$-0.0109$, $+0.0071$] \\
  \multicolumn{9}{@{}l}{\emph{AHRF screen}} \\
  \quad Passed & 4,508 & 2,859 & 19 & 49 & 0.8180 & 0.8058 & $+0.0121$ & [$+0.0042$, $+0.0203$] \\
  \quad Not passed & 4,487 & 3,389 & 18 & 1 & 0.8565 & 0.8617 & $-0.0052$ & [$-0.0252$, $+0.0144$] \\
  \multicolumn{9}{@{}l}{\emph{ICU admission}} \\
  \quad Before 2020 & 2,613 & 1,786 & 19 & 14 & 0.8297 & 0.8215 & $+0.0082$ & [$-0.0071$, $+0.0231$] \\
  \quad From 2020 & 6,382 & 4,169 & 19 & 35 & 0.8381 & 0.8386 & $-0.0005$ & [$-0.0090$, $+0.0079$] \\
  \bottomrule
\end{tabular}

\end{table}

\section{External validation}\label{app:ext}
\paragraph{Conversion.} MIMIC-IV version 3.1~\citep{johnson2024mimiciv31} and eICU version 2.0~\citep{pollard2019eicu20} were converted to MEDS with MIMIC-IV-MEDS and eICU-MEDS, both at version 0.3.0~\citep{mcdermott2026mimicivmeds,mcdermott2026eicumeds}. We then wrote each database in the layout of the MGB data, so that the same code builds the inputs, labels and prompts at all three sites. As at MGB, we kept the events from 24\,h before the first ICU admission to 24\,h after the last ICU discharge of each encounter. In MIMIC-IV, a transfer between ICUs starts a new stay, as at MGB. MIMIC-IV shifts its dates and gives each patient's real year only as a three-year range. We took the middle year of a stay's range as its admission year in the static block (§\ref{sec:tokens}) and moved the patient's clock by whole weeks into that year. The admission years of the sampled MIMIC-IV stays fall between 2018 and 2022. The eICU stays, from 2014 and 2015, take the admission year 2017, the first year of the MGB cohort. eICU-MEDS writes one subject per hospital stay, so we grouped the stays of each patient by eICU's patient identifier.

\paragraph{Mapping.} Bedside charting was mapped item by item to MGB codes. Values were converted to MGB's units and cleaned by the rules used at MGB. In eICU we used the vital signs charted by nurses. Laboratory results were mapped by LOINC code and unit to the laboratory codes used at MGB. In MIMIC-IV, blood-gas results were assigned to arterial or venous codes by the specimen recorded with them. eICU names its blood-gas results as arterial. Medication orders were placed in ATC groups~\citep{whoatc} by drug name and route, with the name-based rescue used for uncoded medications at MGB (App.~\ref{app:data}). Each order is represented by the most common MGB code of its group. In MIMIC-IV and eICU, 86.8\% and 98.4\% of the laboratory rows within the window were placed on an MGB code, as were 89.3\% and 86.8\% of medication orders. Sex, race and ethnicity were mapped to the values recorded at MGB.

\paragraph{Cohorts.} Table~\ref{tab:ext-cohort} compares the three cohorts. In MIMIC-IV, 74,557 ICU stays of adults (54,358 patients) were still in the ICU 24\,h after admission. We kept the 68,402 stays whose first care unit was one of eight ICUs, which leaves out step-down, intermediate and recovery units. We kept the 24,374 stays of 18,940 patients whose range of real years lies wholly in 2017 or later. In eICU, 132,590 ICU stays of adults (104,304 patients) were still in the ICU at 24\,h. We kept the 133 of 208 hospitals in which at least half of these stays had a heart rate charted by a nurse in the first 24\,h and at least half had a named medication order. This left 103,650 stays of 80,866 patients. In each database we then drew patients at random and took all their stays until 5,000 stays were reached. These are 3,897 patients in MIMIC-IV and 3,863 patients from 125 hospitals in eICU. Splitting MIMIC-IV stays at transfers between ICUs gives 5,245 scored stays from 3,884 patients.

\paragraph{Models and statistics.} Both models were applied as trained on MGB. \modelname{} sampled 50 futures for each stay with a budget of 8,192 tokens (§\ref{sec:readout}). When no future was decided, it predicted the fraction of MGB training stays in which the event occurred. Simulation took 49.1\,h for MIMIC-IV and 51.5\,h for eICU, each on one NVIDIA L40S GPU. Fewer futures stopped at the budget than on MGB validation (13\% and 11\%, against 28\%). We refitted the strongest supervised baseline on MGB to store its fitted models. The refit reproduced its validation AUROC on every task to four decimals. Its inputs were computed from each external database with the value bins fitted at MGB. Its 24\,h summaries held on average 84 non-missing values per prediction time in MIMIC-IV and 55 in eICU, against 154 at MGB. The task list was held fixed, so some tasks have fewer than 50 positive stays at an external site and wide intervals. CIs come from 1,000 bootstrap replicates that resample patients. Holm's correction is applied across the 17 tasks within each database. We pre-specified a sensitivity analysis without the two acidaemia tasks, because the laboratory codes that their label reads differ between the databases.

\paragraph{Results.} Table~\ref{tab:ext-auroc} gives each task's AUROC, and Table~\ref{tab:ext-other} gives AUPRC, Brier score and ECE. In both databases, the point estimates favour the strongest supervised baseline on severe and moderate hypoxaemia by \pf{} and \modelname{} on hyperlactataemia, in the same direction as at MGB. In eICU, \modelname{} leads on both mortality tasks with CIs that exclude zero, although it trails on them at MGB. Only FiO$_2$ escalation in MIMIC-IV and ICU discharge and in-hospital mortality in eICU stay significant after Holm's correction. On the two acidaemia tasks, both models reach AUROCs of 0.53 to 0.64 in both databases, against about 0.78 at MGB. Without these tasks, the mean difference is $-0.0001$ (95\% CI $-0.0126$ to $+0.0128$) in MIMIC-IV and $+0.0257$ ($+0.0087$ to $+0.0429$) in eICU. Neither model was recalibrated. \modelname{}'s mean ECE over the 17 tasks is 0.051 in MIMIC-IV and 0.090 in eICU, against 0.057 and 0.080 for the strongest supervised baseline. At MGB the two are 0.032 and 0.013.

\begin{table}[ht]
  \caption{Cohorts at the three sites. The MGB column is the test split (Table~\ref{tab:cohort-full}), and the external columns are the sampled stays. $^a$One ICU unit admission, so a transfer between ICUs starts a new stay. In eICU, 52 unit stays of zero or negative length are not counted. $^b$Median [interquartile range]. $^c$Read from ventilator-setting markers (App.~\ref{app:tasks}). $^d$At MGB, age is rounded to the year before the adult test. $^e$As recorded. Categories with fewer than 11 patients are pooled into the last row of their group, and -- marks a category pooled in that column. In eICU, patients are counted by eICU's patient identifier, and the shares of sex, race and ethnicity are over hospital stays.}
  \label{tab:ext-cohort}
  \centering
  \scriptsize
  \begin{tabular}{@{}lrrr@{}}
  \toprule
  & MGB (test) & MIMIC-IV & eICU \\
  \midrule
  Patients & 7,051 & 3,897 & 3,863 \\
  Hospital encounters & 9,304 & 4,525 & 4,556 \\
  ICU stays$^a$ & 11,852 & 5,770 & 5,668 \\
  Events (millions) & 47.5 & 7.2 & 5.3 \\
  Scored stays (in ICU at 24\,h) & 8,995 & 5,245 & 5,000 \\
  Patients with a scored stay & 5,794 & 3,884 & 3,863 \\
  Age at ICU admission, years$^b$ & 65.7 [53.3, 75.8] & 67.9 [56.3, 76.9] & 66.5 [54.5, 77.5] \\
  Stays at exact age $<$18 years$^d$ & 17 & 0 & 0 \\
  Female, \% & 43.8 & 40.4 & 45.2 \\
  ICU length of stay, days$^b$ & 2.1 [1.0, 4.2] & 2.2 [1.4, 4.3] & 2.1 [1.3, 3.8] \\
  Hospital length of stay, days$^b$ & 6.4 [3.3, 12.1] & 8.8 [5.6, 16.0] & 6.2 [3.7, 10.9] \\
  Invasive ventilation, \% of stays$^c$ & 42.3 & 50.9 & 27.4 \\
  In-hospital death, \% of encounters & 11.1 & 13.2 & 8.4 \\
  In-hospital death, \% of scored stays & 12.1 & 14.6 & 8.6 \\
  ICU death, \% of scored stays & 6.8 & 8.8 & 3.4 \\
  \multicolumn{4}{@{}l}{Race, \% of patients$^e$} \\
  \quad White & 78.7 & 62.4 & 76.1 \\
  \quad Black or African American & 6.7 & 9.9 & 12.2 \\
  \quad Not listed & 5.0 & 7.0 & 3.3 \\
  \quad Asian & 3.3 & 3.0 & 1.1 \\
  \quad Unavailable & 3.2 & 16.6 & 6.6 \\
  \quad Unavailable and White (two entries) & 0.8 & -- & -- \\
  \quad Declined & -- & 0.7 & -- \\
  \quad American Indian or Alaska Native & -- & -- & 0.7 \\
  \quad Other, several or missing & 2.4 & 0.4 & 0.0 \\
  \multicolumn{4}{@{}l}{Ethnicity, \% of patients$^e$} \\
  \quad Not Hispanic & 85.7 & 75.8 & 90.1 \\
  \quad Hispanic & 7.8 & 3.6 & 3.3 \\
  \quad Unavailable & 5.6 & 20.0 & 6.6 \\
  \quad Declined & 0.6 & 0.7 & -- \\
  \quad Other or missing & 0.3 & 0.0 & 0.0 \\
  \bottomrule
\end{tabular}

\end{table}

\begin{table}[ht]
  \caption{Per-task AUROC in the external databases, \modelname{} against the strongest supervised baseline, paired on identical rows. The $p$-value is two-sided, from the bootstrap distribution of the difference. A tick marks a task that stays significant after Holm's correction across the 17 tasks at level 0.05. The last column is the difference on the MGB test split (Table~\ref{tab:full-auroc}). The mean without acidaemia leaves out both acidaemia tasks.}
  \label{tab:ext-auroc}
  \centering
  \scriptsize
  \setlength{\tabcolsep}{4pt}
  \begin{tabular}{@{}lrccccc c@{}}
  \toprule
  & & \multicolumn{5}{c}{External site} & MGB \\
  \cmidrule(lr){3-7}\cmidrule(l){8-8}
  Task & $n$ (pos.) & \modelname{} & Baseline & $\Delta$ [95\% CI] & $p$ & Holm & $\Delta$ \\
  \midrule
  \multicolumn{8}{@{}l}{\emph{MIMIC-IV}} \\
  Severe hypoxaemia & 875 (58) & 0.712 & 0.784 & $-0.072$ [$-0.142$, $-0.009$] & 0.016 &  & $-0.070$ \\
  Severe hypoxaemia, \sfr{} & 1,275 (139) & 0.762 & 0.759 & $+0.004$ [$-0.045$, $+0.054$] & 0.832 &  & $+0.007$ \\
  Moderate hypoxaemia & 543 (141) & 0.736 & 0.779 & $-0.043$ [$-0.086$, $+0.001$] & 0.056 &  & $-0.006$ \\
  FiO$_2$ escalation & 1,839 (175) & 0.673 & 0.742 & $-0.070$ [$-0.107$, $-0.033$] & $<$0.001 & $\checkmark$ & $-0.011$ \\
  FiO$_2$ weaning & 1,839 (123) & 0.959 & 0.963 & $-0.005$ [$-0.013$, $+0.003$] & 0.234 &  & $-0.032$ \\
  New invasive ventilation & 2,445 (88) & 0.730 & 0.655 & $+0.075$ [$+0.008$, $+0.144$] & 0.026 &  & $+0.034$ \\
  Ventilator liberation & 2,613 (401) & 0.677 & 0.642 & $+0.035$ [$+0.005$, $+0.065$] & 0.028 &  & $+0.016$ \\
  Hypotension & 4,920 (1,661) & 0.833 & 0.826 & $+0.007$ [$-0.001$, $+0.015$] & 0.100 &  & $+0.020$ \\
  Acidaemia & 1,139 (347) & 0.585 & 0.620 & $-0.035$ [$-0.081$, $+0.011$] & 0.144 &  & $+0.005$ \\
  Severe acidaemia & 1,606 (361) & 0.566 & 0.603 & $-0.037$ [$-0.082$, $+0.006$] & 0.092 &  & $-0.013$ \\
  Hyperlactataemia & 621 (106) & 0.617 & 0.533 & $+0.084$ [$+0.011$, $+0.158$] & 0.016 &  & $+0.036$ \\
  Severe hyperlactataemia & 1,182 (49) & 0.650 & 0.643 & $+0.006$ [$-0.098$, $+0.104$] & 0.884 &  & $+0.051$ \\
  Acute kidney injury & 5,060 (1,070) & 0.781 & 0.787 & $-0.006$ [$-0.020$, $+0.006$] & 0.324 &  & $-0.012$ \\
  ICU mortality & 5,245 (462) & 0.848 & 0.851 & $-0.003$ [$-0.017$, $+0.012$] & 0.682 &  & $-0.024$ \\
  In-hospital mortality & 5,232 (763) & 0.818 & 0.833 & $-0.014$ [$-0.028$, $-0.001$] & 0.032 &  & $-0.015$ \\
  ICU discharge & 5,245 (1,960) & 0.718 & 0.709 & $+0.009$ [$-0.004$, $+0.023$] & 0.148 &  & $+0.033$ \\
  Death on vasopressors & 2,484 (71) & 0.861 & 0.870 & $-0.009$ [$-0.060$, $+0.034$] & 0.698 &  & $-0.011$ \\
  Mean of 17 & & 0.737 & 0.741 & $-0.004$ [$-0.018$, $+0.008$] & 0.518 & & $+0.001$ \\
  Mean without acidaemia & & 0.758 & 0.758 & $0.000$ [$-0.013$, $+0.013$] & 0.960 & & $+0.001$ \\
  \midrule
  \multicolumn{8}{@{}l}{\emph{eICU}} \\
  Severe hypoxaemia & 495 (40) & 0.751 & 0.807 & $-0.057$ [$-0.137$, $+0.020$] & 0.136 &  & $-0.070$ \\
  Severe hypoxaemia, \sfr{} & 771 (82) & 0.797 & 0.816 & $-0.020$ [$-0.087$, $+0.038$] & 0.520 &  & $+0.007$ \\
  Moderate hypoxaemia & 320 (102) & 0.739 & 0.786 & $-0.047$ [$-0.104$, $+0.010$] & 0.122 &  & $-0.006$ \\
  FiO$_2$ escalation & 1,515 (69) & 0.692 & 0.676 & $+0.016$ [$-0.042$, $+0.074$] & 0.562 &  & $-0.011$ \\
  FiO$_2$ weaning & 1,515 (98) & 0.921 & 0.945 & $-0.024$ [$-0.042$, $-0.006$] & 0.004 &  & $-0.032$ \\
  New invasive ventilation & 3,489 (72) & 0.667 & 0.635 & $+0.032$ [$-0.035$, $+0.100$] & 0.336 &  & $+0.034$ \\
  Ventilator liberation & 1,358 (244) & 0.655 & 0.626 & $+0.029$ [$-0.017$, $+0.073$] & 0.206 &  & $+0.016$ \\
  Hypotension & 4,057 (1,206) & 0.851 & 0.845 & $+0.007$ [$-0.002$, $+0.015$] & 0.130 &  & $+0.020$ \\
  Acidaemia & 623 (72) & 0.592 & 0.590 & $+0.002$ [$-0.075$, $+0.085$] & 0.920 &  & $+0.005$ \\
  Severe acidaemia & 825 (32) & 0.639 & 0.532 & $+0.107$ [$-0.026$, $+0.231$] & 0.120 &  & $-0.013$ \\
  Hyperlactataemia & 209 (46) & 0.699 & 0.553 & $+0.146$ [$+0.037$, $+0.254$] & 0.014 &  & $+0.036$ \\
  Severe hyperlactataemia & 320 (22) & 0.858 & 0.714 & $+0.144$ [$+0.020$, $+0.282$] & 0.026 &  & $+0.051$ \\
  Acute kidney injury & 4,107 (556) & 0.791 & 0.802 & $-0.011$ [$-0.029$, $+0.006$] & 0.198 &  & $-0.012$ \\
  ICU mortality & 4,998 (171) & 0.879 & 0.849 & $+0.030$ [$+0.010$, $+0.052$] & 0.004 &  & $-0.024$ \\
  In-hospital mortality & 4,994 (429) & 0.843 & 0.805 & $+0.038$ [$+0.021$, $+0.055$] & $<$0.001 & $\checkmark$ & $-0.015$ \\
  ICU discharge & 5,000 (2,051) & 0.694 & 0.657 & $+0.038$ [$+0.021$, $+0.055$] & $<$0.001 & $\checkmark$ & $+0.033$ \\
  Death on vasopressors & 880 (41) & 0.898 & 0.834 & $+0.064$ [$-0.014$, $+0.140$] & 0.088 &  & $-0.011$ \\
  Mean of 17 & & 0.763 & 0.734 & $+0.029$ [$+0.011$, $+0.048$] & $<$0.001 & & $+0.001$ \\
  Mean without acidaemia & & 0.782 & 0.757 & $+0.026$ [$+0.009$, $+0.043$] & $<$0.001 & & $+0.001$ \\
  \bottomrule
\end{tabular}

\end{table}

\begin{table}[ht]
  \caption{Per-task AUPRC, Brier score and ECE (10 equal-width bins) in the external databases, \modelname{} against the strongest supervised baseline.}
  \label{tab:ext-other}
  \centering
  \scriptsize
  \setlength{\tabcolsep}{4pt}
  \begin{tabular}{@{}lccccccc@{}}
  \toprule
  & & \multicolumn{2}{c}{AUPRC $\uparrow$} & \multicolumn{2}{c}{Brier $\downarrow$} & \multicolumn{2}{c}{ECE $\downarrow$} \\
  \cmidrule(lr){3-4}\cmidrule(lr){5-6}\cmidrule(l){7-8}
  Task & Prevalence & \modelname{} & Baseline & \modelname{} & Baseline & \modelname{} & Baseline \\
  \midrule
  \multicolumn{8}{@{}l}{\emph{MIMIC-IV}} \\
  Severe hypoxaemia & 0.066 & 0.245 & 0.230 & 0.058 & 0.058 & 0.028 & 0.020 \\
  Severe hypoxaemia, \sfr{} & 0.109 & 0.405 & 0.454 & 0.081 & 0.083 & 0.028 & 0.053 \\
  Moderate hypoxaemia & 0.260 & 0.529 & 0.587 & 0.170 & 0.151 & 0.079 & 0.041 \\
  FiO$_2$ escalation & 0.095 & 0.168 & 0.222 & 0.090 & 0.082 & 0.030 & 0.039 \\
  FiO$_2$ weaning & 0.067 & 0.646 & 0.643 & 0.044 & 0.038 & 0.061 & 0.029 \\
  New invasive ventilation & 0.036 & 0.156 & 0.074 & 0.035 & 0.034 & 0.026 & 0.006 \\
  Ventilator liberation & 0.153 & 0.268 & 0.228 & 0.126 & 0.127 & 0.057 & 0.026 \\
  Hypotension & 0.338 & 0.689 & 0.712 & 0.164 & 0.164 & 0.070 & 0.081 \\
  Acidaemia & 0.305 & 0.375 & 0.399 & 0.225 & 0.241 & 0.105 & 0.162 \\
  Severe acidaemia & 0.225 & 0.278 & 0.322 & 0.185 & 0.209 & 0.074 & 0.166 \\
  Hyperlactataemia & 0.171 & 0.277 & 0.212 & 0.145 & 0.153 & 0.073 & 0.090 \\
  Severe hyperlactataemia & 0.041 & 0.074 & 0.083 & 0.045 & 0.040 & 0.034 & 0.009 \\
  Acute kidney injury & 0.211 & 0.490 & 0.499 & 0.140 & 0.139 & 0.048 & 0.058 \\
  ICU mortality & 0.088 & 0.437 & 0.463 & 0.063 & 0.066 & 0.010 & 0.038 \\
  In-hospital mortality & 0.146 & 0.468 & 0.492 & 0.100 & 0.100 & 0.023 & 0.041 \\
  ICU discharge & 0.374 & 0.564 & 0.560 & 0.217 & 0.218 & 0.098 & 0.099 \\
  Death on vasopressors & 0.029 & 0.413 & 0.277 & 0.025 & 0.025 & 0.030 & 0.017 \\
  Mean of 17 & & 0.381 & 0.380 & 0.112 & 0.113 & 0.051 & 0.057 \\
  \midrule
  \multicolumn{8}{@{}l}{\emph{eICU}} \\
  Severe hypoxaemia & 0.081 & 0.327 & 0.294 & 0.066 & 0.069 & 0.034 & 0.034 \\
  Severe hypoxaemia, \sfr{} & 0.106 & 0.410 & 0.439 & 0.083 & 0.083 & 0.052 & 0.103 \\
  Moderate hypoxaemia & 0.319 & 0.567 & 0.628 & 0.189 & 0.172 & 0.078 & 0.056 \\
  FiO$_2$ escalation & 0.046 & 0.088 & 0.080 & 0.059 & 0.052 & 0.079 & 0.084 \\
  FiO$_2$ weaning & 0.065 & 0.562 & 0.470 & 0.082 & 0.043 & 0.131 & 0.031 \\
  New invasive ventilation & 0.021 & 0.059 & 0.032 & 0.034 & 0.023 & 0.064 & 0.019 \\
  Ventilator liberation & 0.180 & 0.287 & 0.276 & 0.148 & 0.145 & 0.082 & 0.043 \\
  Hypotension & 0.297 & 0.680 & 0.713 & 0.149 & 0.145 & 0.066 & 0.060 \\
  Acidaemia & 0.116 & 0.152 & 0.158 & 0.171 & 0.237 & 0.221 & 0.323 \\
  Severe acidaemia & 0.039 & 0.065 & 0.073 & 0.129 & 0.181 & 0.267 & 0.332 \\
  Hyperlactataemia & 0.220 & 0.393 & 0.248 & 0.164 & 0.180 & 0.076 & 0.066 \\
  Severe hyperlactataemia & 0.069 & 0.494 & 0.209 & 0.051 & 0.063 & 0.040 & 0.030 \\
  Acute kidney injury & 0.135 & 0.341 & 0.432 & 0.116 & 0.098 & 0.092 & 0.022 \\
  ICU mortality & 0.034 & 0.307 & 0.232 & 0.035 & 0.030 & 0.043 & 0.004 \\
  In-hospital mortality & 0.086 & 0.444 & 0.351 & 0.069 & 0.067 & 0.052 & 0.006 \\
  ICU discharge & 0.410 & 0.583 & 0.548 & 0.224 & 0.239 & 0.084 & 0.113 \\
  Death on vasopressors & 0.047 & 0.509 & 0.424 & 0.041 & 0.038 & 0.067 & 0.030 \\
  Mean of 17 & & 0.369 & 0.330 & 0.106 & 0.110 & 0.090 & 0.080 \\
  \bottomrule
\end{tabular}

\end{table}

\end{document}